\documentclass[10pt, journal]{IEEEtran}

\usepackage{graphicx}
\usepackage{amsmath} % for 'multline*' env.
\usepackage{mathtools}
\usepackage{nccmath}
\usepackage{amssymb,amsthm}
\usepackage{booktabs}
\usepackage{gensymb}
\usepackage{algorithm}
\usepackage{algorithmic}
\usepackage[dvipsnames]{xcolor}
\usepackage{multirow}

\newcommand{\matP}[1]{#1^{(v)}}
\newcommand{\matQ}[1]{#1^{(u)}}
\newcommand{\matPQ}[1]{#1^{(vu)}}
\newcommand{\norm}[1]{\left\lVert#1\right\rVert}
\newcommand{\normSquare}[1]{{\norm{#1}}^2}
\newcommand{\tr}{\ensuremath{\text{tr}}} 
\newcommand{\R}{\ensuremath{\mathbb{R}}}
\newcommand{\dnew}{\ensuremath{d}}
\newcommand{\M}{\ensuremath{M}}
\newcommand{\N}{\ensuremath{N}}
\newcommand{\Nm}{\ensuremath{N_m}}
\newcommand{\n}{\ensuremath{n}}
\newcommand{\m}{\ensuremath{m}}
\newcommand{\V}{\ensuremath{V}}
\newcommand{\p}{\ensuremath{v}}
\newcommand{\q}{\ensuremath{u}}
\newcommand{\x}{\ensuremath{x}}
\newcommand{\xii}{\ensuremath{x_i}}
\newcommand{\xj}{\ensuremath{x_j}}
\newcommand{\xp}{\ensuremath{x^{(v)}}}
\newcommand{\xq}{\ensuremath{x^{(u)}}}
\newcommand{\xip}{\ensuremath{x_i^{(v)}}}
\newcommand{\xiq}{\ensuremath{x_i^{(u)}}}
\newcommand{\xjp}{\ensuremath{x_j^{(v)}}}
\newcommand{\xjq}{\ensuremath{x_j^{(u)}}}
\newcommand{\xlq}{\ensuremath{x_l^{(u)}}}
\newcommand{\xlp}{\ensuremath{x_l^{(v)}}}
\newcommand{\xkr}{\ensuremath{x_k^{(r)}}}
\newcommand{\yip}{\ensuremath{y_i^{(v)}}}
\newcommand{\yiq}{\ensuremath{y_i^{(u)}}}
\newcommand{\yjp}{\ensuremath{y_j^{(v)}}}
\newcommand{\yjq}{\ensuremath{y_j^{(u)}}}

\newcommand{\dy}{\ensuremath{d}}
\newcommand{\fp}{\ensuremath{f^{(v)}}}
\newcommand{\fq}{\ensuremath{f^{(u)}}}
\newcommand{\fr}{\ensuremath{f^{(r)}}}
\newcommand{\B}{\ensuremath{\eta}}
\newcommand{\Adel}{\ensuremath{R_\delta}}
\newcommand{\nump}{\ensuremath{\nu_m^{(v)}}}
\newcommand{\numq}{\ensuremath{\nu_m^{(u)}}}
\newcommand{\Aq}{\ensuremath{A^{(u)}}}
\newcommand{\K}{\ensuremath{K}}
\newcommand{\ux}{\ensuremath{z}}

\newtheorem{theorem}{Theorem}
\newtheorem{definition}{Definition}

\newtheorem{lemma}{Lemma}

\begin{document}

\title{Learning Multi-Modal Nonlinear Embeddings: Performance Bounds and an Algorithm}
\author{Semih Kaya and Elif Vural
\IEEEcompsocitemizethanks{
\IEEEcompsocthanksitem S. Kaya and E. Vural are with the Department of Electrical and Electronics Engineering, METU, Ankara.\protect\\
E-mail: kaya.semih@metu.edu.tr, velif@metu.edu.tr
}
}
\maketitle

\begin{abstract}
While many approaches exist in the literature to learn low-dimensional representations for data collections in multiple modalities, the generalizability of multi-modal nonlinear embeddings to previously unseen data is a rather overlooked subject. In this work, we first present a theoretical analysis of learning multi-modal nonlinear embeddings in a supervised setting. Our performance bounds indicate that for successful generalization in multi-modal classification and retrieval problems, the regularity of the interpolation functions extending the embedding to the whole data space is as important as the between-class separation and cross-modal alignment criteria. We then propose a multi-modal nonlinear representation learning algorithm that is motivated by these theoretical findings, where the embeddings of the training samples are optimized jointly with the Lipschitz regularity of the interpolators. Experimental comparison to  recent multi-modal and single-modal learning algorithms suggests that the proposed method yields promising performance in multi-modal image classification and cross-modal image-text retrieval applications.  
\end{abstract}
\section{INTRODUCTION}
\label{sectionIntroduction}

%\IEEEPARstart{W}{ith} the increasing accessibility of data acquisition and storing technologies, the need for successfully analyzing and interpreting multi-modal data collections has become more important. 

\IEEEPARstart{M}{any} data analysis applications involve the acquirement or analysis of data collections in multiple modalities. In some problems, the purpose is to fuse the information in different modalities to improve the detection or classification accuracy, while some other applications require the retrieval of data samples in a certain modality that are relevant to a query sample provided in another modality. For instance, in an image-text cross-modal retrieval problem,  one might be interested in retrieving image samples from the same category as a query text sample. In this paper, we study the problem of learning supervised nonlinear representations for multi-modal classification and cross-modal retrieval applications. 

Multi-modal learning algorithms often aim to compute joint representations in a common domain, where the main challenge is to efficiently align different modalities without damaging their inherent geometry. Subspace learning methods such as CCA \cite{surveyOnMl} align different modalities via linear projections or transformations. Supervised linear embedding methods such as GMLDA \cite{gma} and its various extensions aim to enhance the separation between different data classes in addition to the alignment of different modalities. However, when different modalities have significantly dissimilar geometric structures, linear methods may fall short of learning effective joint representations since they are mostly restricted by the original geometry of the individual modalities. Kernel extensions of linear methods such as Kernel CCA \cite{surveyOnMl}, Kernel GMLDA  \cite{gma} and its variants provide nonlinear representations that may improve some of these shortcomings; however, the resulting algorithms might still lack in flexibility in certain problems. In particular, the suitability of the selected kernel type might vary largely depending on the data set and the embedding may generalize poorly to test data. In the recent years, impressive performance has been attained in retrieval and classification problems with deep learning algorithms based on cross-modal CNNs and autoencoders \cite{multimodalDeepLearning},  \cite{WeiZLWLZY17}, \cite{FengWL14}. While these methods compute powerful nonlinear representations, they typically require much larger data sets and their training complexity is significantly higher. 

%inear embeddings have limitations in challenging data sets where different modalities are weakly linked. In particular, 

While different multi-modal learning approaches might be preferable to each other depending on the setting, their capacity to generalize to novel test samples is a questionable issue in general. A multi-modal learning method may yield promising performance figures on training data, while its performance may be much lower on previously unseen test data, especially if it involves complex and rich models. In fact, the theoretical characterization of the generalization capability of multi-modal embedding algorithms is a somewhat overlooked problem in the literature. Some previous studies have focused on the analysis of co-training  \cite{BlumM98}, \cite{DasguptaLM01} or co-regularized RKHS problems \cite{RosenbergB07}, \cite{Sun11}, which however do not tell what geometric properties a nonlinear multi-modal embedding should have for successful generalization.  

%The previous study \cite{supervisedManifold}  proposes generalization bounds on the performance of supervised nonlinear embedding algorithms; however, it treats the problem in a single modality. %To the best of our knowledge, a mathematically rigorous study of the performance of supervised multi-modal embedding algorithms has not been proposed so far. 

 %In particular,  due to the limited capacity of the model they learn, linear subspace methods may be expected to have relatively close accuracy on training and test data. On the other hand, algorithms involving complex and rich models such as deep learning methods may suffer from overfitting in case of insufficient training data. Although the learnt model fits to the characteristics of the training data very well, it might fail to generalize to previously unseen test data.

In this paper, we consider the problem of learning supervised nonlinear embeddings for multi-modal classification and cross-modal retrieval applications that can generalize well to new test data. Our main purpose in preferring nonlinear embeddings as opposed to subspace methods is to achieve a relatively high model capacity that can adapt to challenging data geometries. On the other hand, we adhere to a shallow data representation model with a single-stage embedding as opposed to deep methods, in order to achieve applicability to settings with restricted availability of training data or limited computation budget. Our study has two main contributions. We first propose a theoretical analysis of learning supervised multi-modal embeddings. We consider a nonlinear embedding model where the training samples from different modalities are jointly mapped to a common lower-dimensional domain, and the training embeddings are extended to the whole data space via Lipschitz-continuous interpolation functions. Our theoretical bounds suggest that for good generalization performance, the multi-modal embedding of training samples should satisfy three conditions: (1) Different modalities should be aligned sufficiently well; (2) Different classes should be sufficiently well-separated from each other; (3) The geometric structure of each modality (captured through nearest neighborhoods) should be preserved. Then, under these conditions, we show that the embedding generalizes well to test data, provided that the Lipschitz constants of the interpolation functions are sufficiently low. This points to an important trade-off: Multi-modal methods may fail to generalize to test data when the nonlinear interpolation functions are too irregular, even if the  embeddings of training samples exhibit good cross-modal alignment and between-class separation properties.

%Our effort hence seeks a balance between the ease of training and the  flexibility of nonlinear models, while ensuring good generalizability to new test data.

Our next contribution is to propose a new supervised nonlinear multi-modal learning algorithm. Motivated by the above theoretical findings, we formulate an optimization problem where a cross-modal alignment term and a between-class separation term are jointly optimized with the Lipschitz constants of the interpolation functions generalizing the embeddings. The resulting objective function is minimized iteratively, by jointly learning the nonlinear embedding coordinates with the  interpolator parameters. Our method has the advantage of providing more flexible representations than subspace methods thanks to the employed nonlinear models, while it entails a relatively lightweight training phase compared to elaborate approaches such as deep learning methods. The proposed method is suitable for multi-modal problems with significantly different data types in different modalities, as well as multi-view problems with closely related or same data types across different views. We test the proposed algorithm in multi-view image classification and image-text cross-modal retrieval applications. Experimental results show that the proposed method yields quite satisfactory performance in comparison with recent multi-modal learning approaches.

The rest of the paper is organized as follows. In Section \ref{sec:RelatedWork}, we overview the related literature. In Section \ref{sec:theo_anly}, we present a theoretical analysis of the multi-modal representation learning problem. In Section \ref{sec:prop_method}, we describe our  supervised nonlinear multi-modal representation learning algorithm. In Section \ref{sec:exp_res}, we experimentally evaluate the  performance of the proposed method, and in Section \ref{sec:concl}, we conclude.

\section{RELATED WORK}
\label{sec:RelatedWork}

The multi-modal learning approaches in the literature can be mainly grouped as co-training methods, subspace learning approaches, kernel methods and deep learning methods. Co-training methods learn separate models in different modalities by encouraging their predictions to be similar \cite{BlumM98}.  A probabilistic model for Support Vector Machine (SVM) is constructed in \cite{coEM} based on the Co-EM approach. There also exist co-regression algorithms employing the co-training idea \cite{coRegularization}. The co-training technique is also used in graph-based methods such as \cite{bayesianCoTraining}, where a Gaussian process model is used on an undirected Bayesian graph. Co-training algorithms have been used in various data analysis applications \cite{fusionCotraining}, \cite{DuanTTLH17}.

Subspace learning methods are based on computing linear projections or transformations that suitably align samples from different modalities. The well-known unsupervised subspace learning algorithm CCA (Canonical Correlation Analysis) maximizes the correlation between different modalities \cite{surveyOnMl}. Alternative versions of CCA such as cluster CCA \cite{clusterCCA}, multi-label CCA \cite{multilabelCca} and three-view CCA \cite{threeViewCca} have been proposed to improve the performance of CCA in various supervised tasks, all of which employ linear projections. In the recent years, many supervised subspace methods have been proposed, which aim to enhance the between-class separation and cross-modal alignment when learning linear projections of data. The GMLDA (Generalized Multiview Analysis) method proposes a multi-modal extension of the LDA algorithm within this framework \cite{gma}. Projection directions for different modalities are learnt by optimizing a quadratic objective function that contains within-class scatter, between-class scatter, and cross-modal correlation terms \cite{gma}. A kernel extension of the GMLDA algorithm for learning nonlinear mappings is also presented in \cite{gma}. Following the approach in  \cite{gma}, many extensions of this work have been proposed in succeeding studies. In  \cite{KanSZLC16}, a view consistency term is added to the objective function of GMLDA so as to impose the similarity of the linear projection functions of different views. The cross-media retrieval method in \cite{WeiZZWXFY16} addresses a supervised linear projection learning problem as in \cite{gma}; however, applies regularization on projection matrices. Several works have focused on kernel extensions of the problem \cite{JingHZWLY14}, \cite{CaoICG18}. The KMvMDA (Kernel Multi-View Modular Discriminant Analysis) method proposed in \cite{CaoICG18} learns kernel representations by imposing within-class and between-class correlation constraints across different modalities. The kernel method in \cite{SunXY16} uncorrelates the feature vectors of the individual modalities as an additional consideration. Another body of methods model data by constructing label-aware data graphs and 
include a graph-based regularization term in the objective in order to preserve the geometry of the data set \cite{YinWW17},  \cite{ZhangMLHT18}. The JFSSL (Joint Feature Selection and Subspace Learning)  method uses a joint graphical model for calculating projections with relevant and irrelevant features \cite{jfssl}.  Various methods are based on combining kernels in different modalities, such as convex combinations of multiple Laplacian kernels \cite{ArgyriouHP05}, or the power mean of multilayer graph kernels \cite{MercadoTH19}. Among all these methods, our method bears similarities to especially supervised kernel methods in that it learns  nonlinear data representations in view of class separation and cross-modality alignment objectives. On the other hand, it has two major differences from these approaches: (1) Our nonlinear representation model is particularly flexible and effective as the pointwise embeddings of training samples are optimized individually by respecting the data geometry. (2) We explicitly incorporate the generalization performance of the algorithm in the objective function, which is a unique and distinctive feature of our method. While there are multi-view algorithms learning pointwise nonlinear mappings as in our method, these often address unsupervised problems such as spectral embedding \cite{spectralEmbeddingMv} and multi-modal clustering \cite{multimodalJointClustering}.

%
%Ext 9,10: Subspace methods (Skip)
%Ext2: Skip for now (similarities expressed in a different way)
%Ext3: (Fei Wu et al, Slightly related, may skip ) Adapts the idea to the cross-media ranking problem, where the projection directions are optimized so as to adjust the cross-modality correlation term by weighing the sample pair alignment terms with respect to their  relative ranking positions. 
%Ext7: Skip, ranking method like Ext3
% Some subspace learning methods propose solutions based on the metric learning \cite{metricLearning} and matrix factorization \cite{matrixFactorization} ideas.

In the recent years, deep learning methods have provided quite effective solutions for analyzing large multi-modal data sets. Deep multi-view autoencoders can learn shared representations  \cite{multimodalDeepLearning}, \cite{FengWL14}, or cross-weights  \cite{deepMultimodalCrossWeights} across different data modalities. Convolutional neural networks are widely used in multi-modal problems as well, where CNN structures for visual modalities can be combined with other modalities at the feature level \cite{WeiZLWLZY17}, \cite{multimodalDeepImageAnnotation}, or the classifier level \cite{multiViewImageClassification}. GAN-type architectures adversarially train feature generators and domain discriminators across different modalities \cite{deepMultimodalRepresentation}. The method in \cite{ZhaoDF17} learns a common latent representation for different modalities via a deep matrix factorization scheme. 
%Deep multi-view autoencoders can learn shared representations  \cite{multimodalDeepLearning}, \cite{FengWL14}, \cite{mmRetrievalDeepLearning}, or cross-weights  \cite{deepMultimodalCrossWeights} across different data modalities. ...CNN structures for visual modalities can be combined with other modalities at the feature level \cite{WeiZLWLZY17}, \cite{CastrejonAVPT16}, \cite{multimodalDeepImageAnnotation}, or the classifier level \cite{multiViewImageClassification}. 

Some previous studies proposing a theoretical analysis of multi-view learning are the following. The study in \cite{BlumM98} analyses the learnability of joint models in the two-view co-training problem, assuming the conditional independence of two views. Several other PAC-style bounds are proposed in \cite{DasguptaLM01}, \cite{SunSM17}, mainly stating that the agreement of the classifiers of the two views on training data guarantees a good estimate of the expected test error. Several studies have proposed generalization bounds for co-regularized RKHS methods \cite{RosenbergB07}, \cite{Sun11}, \cite{XuTX15} in terms of the Rademacher complexities of the involved function classes. These previous analyses differ from ours in that they all aim to bound the difference between the training loss and the expected loss in multi-view classification, while our analysis addresses the particular problem of nonlinear dimensionality reduction in multi-modal learning. Our distinctive contribution is that we explicitly characterize the geometric properties and the regularity conditions of the nonlinear embedding to achieve successful generalization.

%The performance of the SVM-2K algorithm combining the KCCA and SVM approaches is analyzed in \cite{FarquharHMSS05}. 

Finally, some other previous works related to our study are the following. The theoretical analysis in  \cite{supervisedManifold} provides performance bounds for supervised nonlinear embeddings in a single modality. The idea in \cite{supervisedManifold} is developed in this paper to perform a theoretical analysis for multi-modal embeddings. The previous work \cite{nsse} proposes a supervised nonlinear dimensionality reduction algorithm via smooth representations like in our work; however, it treats the embedding problem in a single modality. Lastly, a preliminary version of our work was presented in \cite{KayaV19}. The current paper builds on \cite{KayaV19} by including a theoretical analysis of the multi-modal learning problem and significantly extending the experimental results.

\section{Performance Bounds for Multi-Modal Learning with Supervised Embeddings}
\label{sec:theo_anly}

In this section, we first describe the multi-modal representation learning setting considered in this study and then present a theoretical analysis of multi-modal classification and retrieval with supervised embeddings. 

\subsection{Notation and Setting}

We consider a setting with $\M$ data classes and $\V$ modalities (also called \textit{views)} such that a data sample $\x$ has an observation $\xp$ in each modality (or view) $\p=1, \dots, \V$. Let the data samples from each class $\m=1, \dots, \M$ in each modality $\p=1, \dots, \V$ be drawn from a probability measure $ \matP \nu_\m$ on a Hilbert space $\matP H$. We assume that the probability measure $ \matP \nu_\m$  has a bounded support $\matP{\mathcal{M}_\m} \subset \matP H$ for each $\p$, and that the probability measures $\{ \matP \nu_\m \}$ in different modalities $\p$ are independent for each class $\m$. 

Let $\mathcal{X}=\{\x_i\}$ be a set of training samples such that each $i$-th training sample $x_i$ belongs to one of the classes  $\m=1, \dots, \M$. In each modality $\p$, the observations of the training samples $\{\xip\}$ from each class $\m$ are independent and identically distributed, drawn from the probability measure  $ \matP \nu_\m$. In this paper, we study a setting where the training samples from all modalities are embedded as $\mathcal{Y}=\{ \yip \}$ into a  common Euclidean domain ${\R}^{\dy}$, such that each training sample $\xip \in \matP H$ from modality $\p$ is mapped to a vector $\yip \in {\R}^{\dy}$. Although we do not impose any conditions on the dimension $\dy$ of the embedding, $\dy$ is typically small in many methods. %Notice that in the considered embedding scenario, the observations $\xip$ of a training sample $\x_i$ in different modalities are embedded individually as separate vectors $\yip$ into the common domain $\R^\dy$. This is for two reasons: First, the observations $\xip$ of a training sample $\x_i$ may be missing in some modalities $\p$ in practice. Next, such an additional degree of freedom when computing the embeddings provides a relaxation of the learnt representations, which makes it possible to compute interpolators with better regularity.

%Our main purpose in this study to find an embedding that can be successfully generalized to initially unavailable test samples with unknown class labels in classification or retrieval applications. We propose to generalize the embedding of the training samples to the whole data space through interpolation functions $\matP{f} : \matP H \rightarrow \R^ \dy $, for $\p=1, \dots, \V$, such that each training sample in a modality $\p$ is mapped to its embedding as $\matP{f} ( \matP{x_i} ) = \matP{y_i}$.

Focusing mainly on a scenario where the embedding is nonlinear in this work, we assume that the embedding of the training samples is extended to the whole data space through interpolation functions $\matP{f} : \matP H \rightarrow \R^ \dy $, for $\p=1, \dots, \V$, such that each training sample in a modality $\p$ is mapped to its embedding as $\matP{f} ( \matP{x_i} ) = \matP{y_i}$. We characterize the regularity of the interpolation functions $\fp $ with their Lipschitz continuity, which is defined as follows.

\begin{definition}
A function $f: H \rightarrow \R^\dy$ defined on a Hilbert space $H$ is Lipschitz continuous with constant $L>0$ if for any $x_1,  x_2 \in H$, the function satisfies 
$
\| f(x_1) - f(x_2) \| \leqslant L \, \| x_1 - x_2 \|
$.
\end{definition}
The notation $\|  \cdot \|$ will denote the usual  norm in the space of interest (e.g. $L^2$-norm, or $\ell^2$-norm), unless stated otherwise. Now, for each modality $\p$, let $B_\delta(\xp) \subset \matP{H}$ be an open ball of radius $\delta$ around the point $\xp$
\[
B_\delta(\xp)  = \{ \matP{\ux} \in \matP{H}:   \| \xp -  \matP{\ux}  \| < \delta  \}.
\]
Then, for each class $\m$, we define a parameter $\eta_{\m, \delta}$, which is a lower bound on the measure of the open ball $B_\delta(\xp)$ around any point from class $\m$ in any modality
\[
\eta_{\m, \delta} := \min_{\p=1, \dots, V} \inf_{\xp \in \matP{\mathcal{M}_\m}}    \matP \nu_\m \left( B_\delta(\xp) \right).
\]

In the following, $C(\cdot)$ denotes the class label of a sample, $| \cdot |$ refers to the cardinality of a set, the notation $z \sim \nu$ means that the sample $z$ is drawn from the distribution $\nu$,  $P(\cdot)$ denotes the probability of an event, and $\|  \cdot \|_F$ denotes the Frobenius norm. The notation $\tr(\cdot)$ stands for the trace of a matrix, and $(\cdot)_{ij}$ indicates the entry of a matrix in the $i$-th row and the $j$-th column. 

\subsection{Theoretical Analysis of Classification and Retrieval Performance}

We now present performance bounds for the multi-modal classification problem and the cross-modal retrieval problem.\\

\subsubsection{Multi-Modal Classification Performance}

Let $\x$ be a test sample with an observation $\xp$ available in a specific modality $\p$. Denoting the true class of $\x$ by $m$, we assume that the observation $\xp$ of the test sample is drawn from the probability measure $ \matP \nu_\m$ independently of the training samples. 

We consider a classification setting where the class label of $\xp$ is estimated by first embedding  $\xp$ into $\R^\dy$ as $\matP{f} (\xp)$ through the interpolator $\matP{f} $ learnt using the training samples. Then the estimate $\hat C (\x)$ of the class label $\hat C (\x)$ of $\x$ is found via nearest-neighbor classification in $\R^\dy$ over the embeddings $\yiq$ of the training samples $\xiq$ from all modalities $\q=1, \dots, \V$. Hence, the class label of the test sample $\x$ is estimated as $\hat C (\x) = C (\x_{i^*})$,  where \footnote{We adopt the notation $C(\x)$ instead of $C(\xp)$ for class labels as the observation $\xp$ of a sample $\x$ in any modality $\p$ has the same class label.}  

\begin{equation}
\label{eq:nn_class_rule}
i^* = \arg \min_{i} \min_{\q=1, \dots, \V} \| \yiq  - \matP{f} (\xp) \|.
\end{equation}

In the following theorem, we present our main result for multi-modal classification with supervised embeddings.

\begin{theorem}
\label{thm:mm_class_bound}
Let the training sample set $\mathcal{X}$ contain at least $\Nm$ training samples $\{\xii \}_{i=1}^{\Nm}$ from class $\m$, whose observations $\{ \xiq \}$ with $\xiq \sim  \matQ \nu_\m$ are available in all modalities $\q=1, \dots, \V$. Let $\mathcal{Y}$ be an embedding of $\mathcal{X}$ in $\R^\dy$ with the following properties
\begin{equation*}
\begin{split}
(P1) \quad \| \yip &- \yiq \| \leq \B \text{ for all training samples } \x_i \text{ and} \\ &\quad \text{ for  all } \p, \q \in \{1, \dots, \V \}  \\
(P2) \quad \| \yiq &- \yjq \| \leq \Adel \ \text{ for all } \q \in \{1, \dots, \V \}, \\ 
&\quad  \text{ if  }  \| \xiq - \xjq \| \leq 2\delta 
\text{ and } C(\xii) = C(\xj) \\
(P3) \quad \| \yip &- \yjq \| > \gamma \ \text{ for all } \p, \q, \in \{1, \dots, \V \} \\
&\quad  \text{ if } C(\xii) \neq  C(\xj)
\end{split}
\end{equation*}
where $\B $ and $\gamma$ are some constants and  $\Adel $ is a $\delta$-dependent constant. Assume that the interpolation function $\fq: \matQ H \rightarrow \R^ \dy $ in each modality $\q$ is a Lipschitz continuous function with constant $L$ such that for some parameters $\epsilon >0$ and $\delta >0$, the following inequality is satisfied
\begin{equation}
\label{eq:cond_L_gamma}
6 L \delta + 2 \sqrt{\dy} \epsilon + 2 \Adel + 2\B \leq \gamma.
\end{equation}
Then for some $Q \geq 1$, if the number of training samples is such that 
\begin{equation}
\label{eq:cond_Nm_Q}
\Nm > \frac{Q}{\eta_{\m, \delta}},
\end{equation}
the probability of correctly classifying a test sample $\x$ from class $\m$ observed as $\xp$ in modality $\p$ via the nearest neighbor classification rule in \eqref{eq:nn_class_rule} is lower bounded as
\begin{equation}
\label{eq:bnd_PCx_m}
\begin{split}
P \left( \hat C(\x)=m \right) &\geq 1 - \bigg[  \exp \left(- \frac{2 (\Nm \eta_{\m, \delta} - Q)^2 }{\Nm} \right) \\
& +  2\dy \exp \left(-\frac{ Q  \epsilon^2}{2 L^2 \delta^2} \right) 
+  (1-\eta_{\m, \delta} )^Q \bigg]^V.
\end{split}
\end{equation}

\end{theorem}

\begin{figure}[t]
\centering
  \includegraphics[width=9cm]{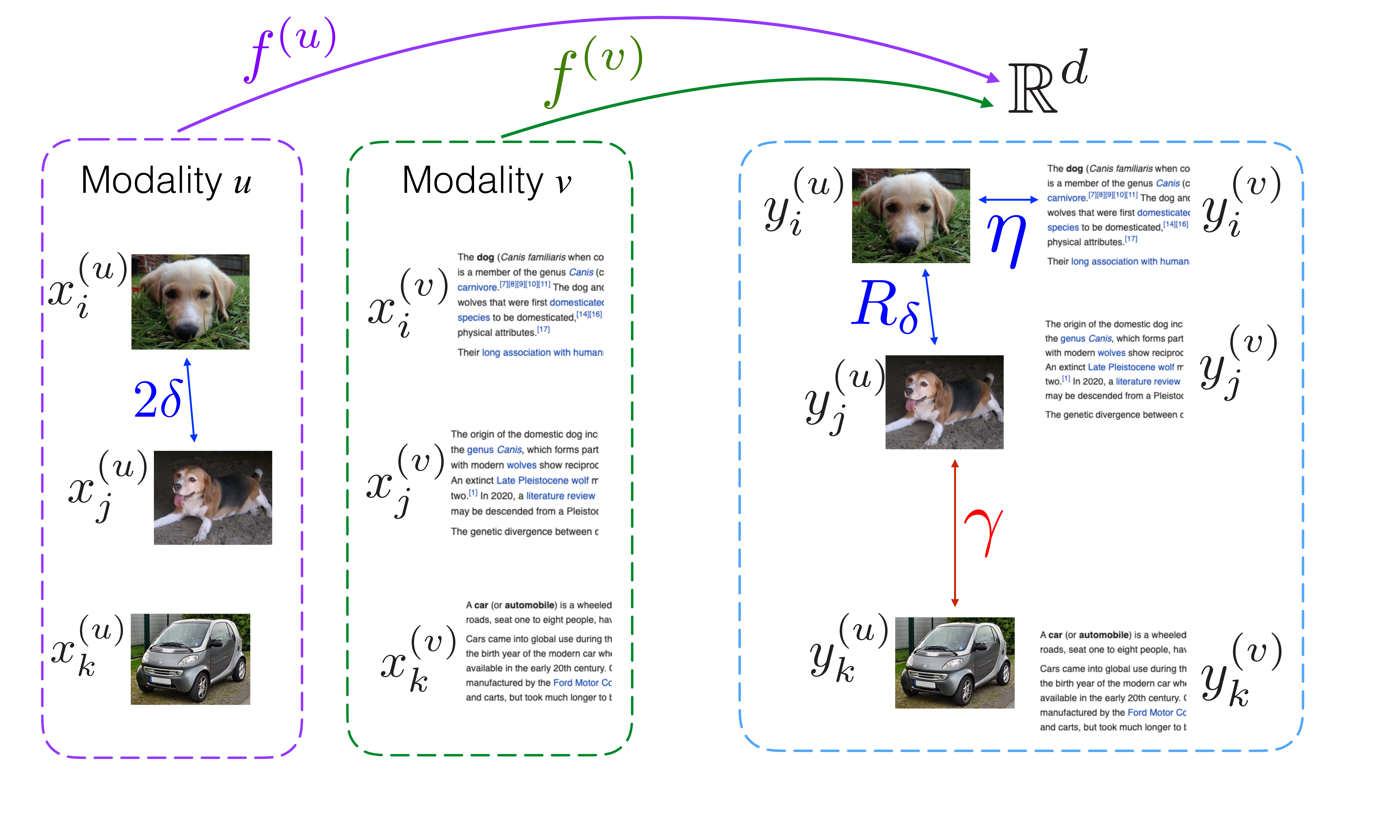}
    \caption{Illustration of the studied multi-modal embedding setting. Modalities $\q$ (image) and $\p$ (text) are mapped to the common domain $\R^\dnew$ via interpolators $\fq$ and $\fp$. The parameters $\B $, $\Adel$, and $\gamma$ respectively measure the alignment between different modalities, the within-class compactness, and the separation between different classes. (Images:  wikipedia.org)}
  \label{fig: illus_model}
\end{figure}

The proof of Theorem \ref{thm:mm_class_bound} is given in the Appendix. The theorem intuitively states the following: First, (P1), (P2), and (P3) define the properties that the embedding should have, which are illustrated in Figure \ref{fig: illus_model}. (P1) requires the observations $\xip$, $\xiq$ of the same training sample $\xii$ in two different modalities to be mapped to nearby points in the common domain $\R^\dy$ of embedding, so that the distance between their embeddings does not exceed some threshold $\B>0$. This property imposes that different modalities be well aligned through the learnt embedding. The property (P2) indicates that two nearby samples from the same modality and the same class should be mapped to nearby points, so that a distance of $2 \delta$ in the original domain is mapped to a distance of at most $\Adel$ in the domain of embedding, where $\Adel$ is a constant depending on $\delta$. This can be seen as a condition for the preservation of the local geometry of each modality within the same class. Lastly, the property (P3) imposes samples from different classes to be separated by a distance of at least $\gamma$ in the domain of embedding, regardless of their modality. Here, the parameter $\gamma>0$ can be seen as a separation margin between different classes in the learnt embedding.

If the embedding of the training samples has these properties, supposing that the condition in \eqref{eq:cond_L_gamma} is satisfied, Theorem \ref{thm:mm_class_bound} guarantees that the probability of correctly classifying a test sample from some class $\m$ approaches $1$ at an exponential rate as the number of training samples $\Nm$ from that class increases. This can be verified by observing that $\Nm$ should be chosen proportionally to the parameter $Q$ as seen in \eqref{eq:cond_Nm_Q}, in which case the correct classification probability  in \eqref{eq:bnd_PCx_m} improves at rate $1-  e^{-O(\V \Nm)} $. Here, an important observation is that as the number of modalities $V$ increases, the correct classification probability improves at an exponential rate. This confirms that the multi-modal learning algorithm can successfully fuse the information obtained from different modalities for improving the classification performance. 

Finally, a crucial implication of Theorem \ref{thm:mm_class_bound} is that the condition in  \eqref{eq:cond_L_gamma} must be satisfied in order to achieve high classification accuracy. The condition  \eqref{eq:cond_L_gamma}  is quite central to our study and it will be of importance when proposing an algorithm in Section \ref{sec:prop_method}.  It states that a certain compromise must be sought between the Lipschitz regularity of the interpolator and the separation between different classes: When learning nonlinear embeddings, the separation $\gamma$ between training samples from different classes should be adjusted in a way to allow the existence of a sufficiently regular interpolator, so that $L$ remains sufficiently small. While an embedding with a too small $\gamma$ value would fail to satisfy the condition  \eqref{eq:cond_L_gamma}, increasing $\gamma$ too much would result in a highly irregular warping of the training samples, which typically leads to an increase in the magnitude of the interpolator parameters. This results in an interpolator with poor Lipschitz regularity with a large $L$ value where the condition \eqref{eq:cond_L_gamma} would fail again. Hence, the condition \eqref{eq:cond_L_gamma} points to how the separation margin and the interpolator regularity should be jointly taken into account when learning an embedding with good generalization properties.\\

\subsubsection{Cross-Modal Retrieval Performance}

Next, we analyze the performance of cross-modal retrieval via supervised embeddings. Given the multi-modal data set $\mathcal{X}=\{\xii\}$, where each data sample $\xii$ belongs to one of the classes $\m=1, \dots, \M$, we formally define the retrieval problem as follows. Let $\xp$ be a query test sample observed in modality $\p$. We study a cross-modal retrieval setting where the purpose is to retrieve samples from a certain modality $\q$ that are ``relevant'' to the query sample $\xp$ from modality $\p$. We consider two samples to be relevant if they belong to the same class.

Denoting the modality of the query sample by $\p$ and the modality of the retrieved samples by $\q$, we consider a retrieval strategy that returns the most relevant $\K$ samples to the query sample, based on the distance of the samples in the domain of embedding. Hence, given the query sample $\xp$, it is first embedded into $\R^\dy$ as $\fp(\xp)$ via the interpolator $\fp$; and then the $\K$ training samples $\{\xiq \}$ from modality $\q$ whose embeddings $ \{  \fq (\xiq ) \}$ have the smallest distance to  $\fp(\xp)$ are retrieved as the most relevant samples, thus returning the set $\{ \xq_{i_k} \}_{k=1}^\K$, where
\begin{equation}
\label{eq:retriev_rule}
\begin{split}
i_1 &= \arg \min_i  \|  \fq (\xiq )  - \fp(\xp)  \|  \\
i_k &= \arg \min_{i \notin \{ i_1, \dots i_{k-1} \} }  \|  \fq (\xiq )  - \fp(\xp)  \|, 
 k=2, \dots, \K.
\end{split}
\end{equation}

The precision rate $P$ and the recall rate $R$ of the retrieval algorithm are then given by
\begin{equation}
\label{eq:prec_recall_defn}
P=\frac{TP}{TP+FP}, \quad \quad
R=\frac{TP}{TP+FN}
\end{equation}
where $TP$, $FP$, and $FN$  respectively denote the number of true positive, false positive, and false negative samples depending on whether the retrieved and unretrieved samples are relevant or not.

We present the following main result regarding the performance of cross-modal retrieval with supervised embeddings.

\begin{theorem}
\label{thm:cm_retrieval_bnd}
Let the training sample set $\mathcal{X}$ contain $\Nm$ training samples $\{\xii \}_{i=1}^{\Nm}$ from class $\m$, with observations $\{ \xip \}$ and $\{ \xiq \}$  available in the modalities $\p$ and $\q$. Let $\mathcal{Y}$ be an embedding of $\mathcal{X}$ in $\R^\dy$ with the following properties: 
\begin{equation*}
\begin{split}
 (P1) \quad & \| \yip - \yiq \| \leq \B \text{ for all training samples } \xii \\ 
 (P2) \quad & \text{For two samples }  \xii \text{ and } \xj \text{ with } C(\xii) = C(\xj)\\
 \quad \quad & \| \yip - \yjp \|  \leq \Adel \ \text{ if  } \  \| \xip - \xjp \| \leq 2\delta; \\
\quad \quad  & \| \yiq - \yjq \|  \leq \Adel \ \text{ if  } \  \| \xiq - \xjq \| \leq 2\delta \\
 (P3) \quad & \| \yip - \yjq \| > \gamma   \text{ if } C(\xii) \neq  C(\xj),
\end{split}
\end{equation*}
where $\B $ and $\gamma$ are some constants and  $\Adel $ is a $\delta$-dependent constant. Assume that the interpolation functions $\fp: \matP H \rightarrow \R^ \dy $ and $\fq: \matQ H \rightarrow \R^ \dy $   in modalities $\p$ and $\q$ are Lipschitz continuous with constant $L$ such that for some parameters $\epsilon >0$ and $\delta >0$, the following inequality holds
\begin{equation}
\label{eq:cond_L_gamma_retr}
6 L \delta + 2 \sqrt{\dy} \epsilon + 2 \Adel + 2\B \leq \gamma.
\end{equation}
For some $Q \geq 1$, let the number of training samples from class $\m$ be such that 
\[
\Nm > \frac{Q}{\eta_{\m, \delta}}.
\]
Let $\xp \sim \nump$ be a query sample from class $\m$ observed in modality $\p$,  the relevant samples to which are sought in modality $\q$. Then,
with probability at least
\begin{equation*}
\begin{split}
1 &- \exp \left(- \frac{2 (\Nm \eta_{\m, \delta} - Q)^2 }{\Nm} \right) 
 -  2\dy \exp \left(-\frac{ Q  \epsilon^2}{2 L^2 \delta^2} \right) \\
&-  (1-\eta_{\m, \delta} )^Q
\end{split}
\end{equation*}
the precision rate $P$ of the retrieval algorithm in \eqref{eq:retriev_rule} satisfies
\begin{equation}
\label{eq:prec_rate}
\begin{split}
P&=1,  \quad \text{ if } \K \leq Q \\
P &\geq \frac{Q}{\K},  \quad \text{  if } \K>Q
\end{split}
\end{equation}
and the recall rate $R$ of the retrieval algorithm satisfies
\begin{equation}
\label{eq:recall_rate}
\begin{split}
R &= \frac{K}{\Nm}.  \quad \text{ if } \K \leq Q \\
R &\geq \frac{Q}{\Nm},  \quad \text{  if } \K>Q.
\end{split}
\end{equation}

\end{theorem}

The proof of Theorem \ref{thm:cm_retrieval_bnd} is given in the Appendix. Theorem \ref{thm:cm_retrieval_bnd} can be interpreted similarly to Theorem \ref{thm:mm_class_bound}. The properties (P1), (P2) and (P3) ensure that the learnt embedding aligns modalities $\p$ and $\q$ sufficiently well, while mapping nearby samples from the same classes to nearby points, and increasing the distance between samples from different classes. Assuming that the condition \eqref{eq:cond_L_gamma_retr} is satisfied, the precision and recall rates given in \eqref{eq:prec_rate} and \eqref{eq:recall_rate} are attained with probability approaching $1$ at an exponential rate as the number of training samples increases. In the proof of the theorem, the precision and recall rates in \eqref{eq:prec_rate} and \eqref{eq:recall_rate} are obtained by identifying the conditions under which at least $Q$ samples out of the $K$ samples returned by the retrieval algorithm are relevant to the query sample. 

The condition \eqref{eq:cond_L_gamma_retr} required for successful cross-modal retrieval is the same as the condition \eqref{eq:cond_L_gamma} for accurate multi-modal classification. Hence, similarly to the findings of our multi-modal classification analysis, the results of our retrieval analysis also suggest that it is necessary to find a good compromise between the Lipschitz continuity of the interpolators and the separation between different classes when learning nonlinear embeddings for cross-modal retrieval applications.
\section{PROPOSED MULTI-MODAL SUPERVISED EMBEDDING  METHOD}
\label{sec:prop_method}

% --------------------------------------------------------------
% SUBSECTION 2.1
% --------------------------------------------------------------

In this section, we propose a multi-modal nonlinear dimensionality reduction algorithm that relies on the theoretical findings of Section \ref{sec:theo_anly}.  We formulate the nonlinear embedding problem in Section \ref{sec:prob_form} and then discuss its solution in Section \ref{ssec:soln_prob}.

\subsection{Problem Formulation}
\label{sec:prob_form}

Let $\matP{X} \in \R^{\matP{N} \times \matP{\n}} $ denote the training data matrix of modality $\p$, each row of which is the observation $\matP{x_i}$ of some training sample $\xii$ in the $\p$-th modality. Here $\matP{N}$ is the total number of observations\footnote{Although the observations of all training samples were assumed to be available in all modalities for the simplicity of the  theoretical analysis in Section \ref{sec:theo_anly}, here we remove this assumption and allow some observations to be missing in some modalities. Hence  $\matP{N}$ may be different for different $\p$.} from all classes in modality $\p$, and $\matP{\n}$ is the dimension of the Hilbert space  $\matP H$ of modality $\p$, assumed to be finite in a practical setting. Given the training samples $\matP{X} $ from modalities $\p=1, \dots, V$, we would like to compute embeddings  $\matP{Y} \in \R^{\matP{N} \times \dnew} $ of the training samples into the common domain $\R^ \dnew $, such that each $\matP{x_i} \in \R^{\matP{\n}}$ is mapped to a vector $ \matP{y_i} \in \R^ \dnew $. The embedding is extended to the whole data space through interpolation functions $\matP{f} : \R^{\matP{\n}} \rightarrow \R^ \dnew $ such that each training sample is mapped to its embedding as $\matP{f} ( \matP{x_i} ) = \matP{y_i}$.

Our main purpose is to find an embedding that can be successfully generalized to initially unavailable test samples. We recall from our theoretical analysis that for successful generalization in multi-modal classification and retrieval, the embedding must have the properties (P1), (P2) and (P3) given in Theorems \ref{thm:mm_class_bound} and \ref{thm:cm_retrieval_bnd}, while the  Lipschitz constant of the interpolators must be kept sufficiently small as imposed by the conditions \eqref{eq:cond_L_gamma} and \eqref{eq:cond_L_gamma_retr}. We now formulate our multi-modal learning problem in the light of these results.

%Furthermore, optimization of samples in different class can be demonstrated as $\underset{\matP{Y}}{max} \ tr \Big( {\matP{Y}}^T \matP{L_b} \matP{Y} \Big)$. Corresponding Laplacian matrices $\matP{L_w}$ and $\matP{L_b}$ can be calculated as $\matP{L_w} = \matP{D_w} - \matP{W_w}$ and $\matP{L_b} = \matP{D_b} - \matP{W_b}$ respectively. Weight and degree matrices are obtained in equation (\ref{eqnWeights1}), (\ref{eqnWeights2}) and equation (\ref{eqnWeights3}) :
% --------------------------------------------------------------
% Weight Matrices Equations
%% --------------------------------------------------------------
%\begin{align} 
%    \label{eqnWeights1}
%    \matP{W_w}(i,j) = exp \Bigg(\frac{-\norm{\matP{x_i} - \matP{x_j}}^2}{2 {\matP{\theta}}^2} \Bigg) \\
%% --------------------------------------------------------------
%    \label{eqnWeights2}
%    \matP{W_b}(i,j) = 
%    \begin{cases}
%        1, & \text{if $C(\matP{x_i}) \neq C(\matP{x_j})$}\\
%        0, & \text{otherwise}
%    \end{cases} \\ 
%% --------------------------------------------------------------
%    \label{eqnWeights3}
%    ve \
%    \matP{D_w} = \matP{I_w} - \matP{W_w}, \ \matP{D_b} = \matP{I_b} - \matP{W_b} 
%\end{align}

\textbf{Lipschitz regularity of the interpolators.} For the extension of the embedding, we choose to use RBF interpolation functions, which are analytical functions with well-studied properties. Hence, the interpolator of each modality $\p=1, \dots, \V$ has the form
 $ \matP{f} (\xp) =[f_1^{(\p)}(\xp) \ \dots \ f_\dnew^{(\p)}(\xp) ] $,  where 
 \begin{equation}
 \label{eq:form_rbf_interp}
 f_k^{(\p)}(\xp) = \sum_{i=1}^{\matP{N}} C^{(\p)}_{ik} \, \phi^{(\p)} \left( \|  \xp -\matP{x_i} \| \right)
 \end{equation}
 is the $k$-th component of $ f^{(\p)}(\xp)$. Here
 \[
 \phi^{(\p)}(r)=e^{-r^2/(\sigma^{(\p)})^2}
 \] is a Gaussian RBF kernel with scale parameter $\sigma^{(\p)}$ and $C^{(\p)}_{ik}$ are the interpolator coefficients. 
 
The Lipschitz continuity of Gaussian RBF interpolators has been studied in \cite{nsse}, from which it follows that $f^{(\p)}(\xp)$ is Lipschitz-continuous with constant
\begin{equation}
\label{eq:expr_lips_const}
 L^{(\p)} = \sqrt{2}e^{-\frac{1}{2}} \sqrt{\matP{N}} ( \sigma^{(\p)})^{-1} {\norm{C^{(\p)}}}_F.
\end{equation}
 Here $C^{(\p)}$ is the coefficient matrix with entries $C^{(\p)}_{ik}$. The interpolator coefficients can be easily obtained as
 \[
 C^{(\p)}= (\Psi^{(\p)})^{-1}\matP{Y}
 \]
 by fitting the embedding coordinates $\matP{Y}$ to the training data $\matP{X}$, where  $\Psi^{(\p)} \in \R^{ \matP{\N} \times  \matP{\N}}$ is the RBF kernel matrix with entries $\Psi^{(\p)}_{ij}=\phi^{(\p)}(\| \matP{x_i}  -\matP{x_j} \|)$. 
 
The conditions \eqref{eq:cond_L_gamma} and \eqref{eq:cond_L_gamma_retr} suggest that the Lipschitz constants of the interpolators should be sufficiently small for successful generalization of the embedding to test data. In view of these results, when learning a nonlinear embedding, we propose to minimize the kernel scale of each modality $\p$ through the term
\[
    \sum_{\p=1}^\V ( \matP{\sigma} )^{-2} 
\]
as well as the interpolator coefficients of all modalities through
\begin{equation*}
\begin{split}
  \sum_{\p=1}^\V {\norm{C^{(\p)}}}_F^2 
=      \sum_{\p=1}^\V   \|  (\Psi^{(\p)})^{-1}\matP{Y} \|_F^2   
&=  \tr( \tilde{Y}^T  \tilde{\Psi} ^{-2}  \tilde{Y} )
\end{split}
\end{equation*}
so that the Lipschitz constant  $L^{(\p)} $ in \eqref{eq:expr_lips_const} is minimized for each modality $\p$. Here 
\[
    \tilde{Y} = [ (Y^{(1)})^T  \  (Y^{(2)})^T \  \hdots \ (Y^{(\V)})^T  ]^T 
 \in \R^{\N  \times \dnew }
\]
denotes the matrix containing the embeddings from all modalities (with $\N=\sum_{\p} \matP{\N}$) and $ \tilde{\Psi} \in \R^{\N \times \N}$ is a block-diagonal matrix containing the kernel matrix $\matP{\Psi}$ in its $\p$-th block.
%

% in order to ensure that the interpolators of all modalities have small Lipschitz constants, for each modality $\p=1, \dots, \V$, we propose to minimize 
 
%
%in addition to the minimization of the kernel scale term $ ( \matP{\sigma} )^{-1}  $. 

\textbf{Within-class compactness.} Theorems  \ref{thm:mm_class_bound} and \ref{thm:cm_retrieval_bnd} suggest that the constant $\Adel$ in (P2) should be kept small, so that the conditions \eqref{eq:cond_L_gamma} and \eqref{eq:cond_L_gamma_retr} are more likely to be met. Although it is not easy to analytically formulate the minimization of  $\Adel$, in practice if nearby samples from the same modality and same class are embedded into nearby points, $\Adel$ will be small. This problem is well-studied in the manifold learning literature. The total weighted distance between the embeddings of same-class samples can be formulated as
\begin{equation}
\label{eq:within_class_modal}
  \sum_{\p=1}^\V \sum_{i,j=1}^{\matP{N}} (\matP{W_w})_{ij} \,  \|  \matP{y_i} - \matP{y_j} \|^2
%=   \sum_{\p=1}^\V \tr \Big({\matP{Y}}^T \matP{L_w} \matP{Y} \Big)
= \tr( \tilde{Y}^T  \tilde{L}_w \tilde{Y} ).
\end{equation}
Here $\matP{W_w} \in \R^{\matP{N} \times \matP{N}} $ is chosen as a weight matrix whose entries $ ( \matP{W_w})_{ij} = \exp ( - \| \matP{x_i} - \matP{x_j} \| ^2 / (\matP{\theta})^2 ) $ represent the affinity between the data samples when $ \matP{x_i} $ and $ \matP{x_j}$ are from the same class (for a scale parameter $ \matP{\theta}$), and $ (\matP{W_w})_{ij} =0$ otherwise. 
In the equality, the block-diagonal matrix $ \tilde{L}_w \in \R^{\N \times \N}$ contains the within-class Laplacian  $\matP{L_w}= \matP{D_w} - \matP{W_w}$ in its $\p$-th block, where $\matP{D_w}$ is the diagonal degree matrix with $i$-th diagonal entry given by $ \sum_j (\matP{W_w})_{ij} $. The term in \eqref{eq:within_class_modal} hence imposes nearby samples $\matP{x_i}$, $ \matP{x_j} $ from the same class and the same modality to be mapped to nearby coordinates.
%
%\par For each modal $p$, optimization of samples in the same class can be indicated as $\underset{\matP{Y}}{min} \ tr \Big({\matP{Y}}^T \matP{L_w} \matP{Y} \Big)$. 

\textbf{Between-class separation.} In Theorems \ref{thm:mm_class_bound} and \ref{thm:cm_retrieval_bnd}, the between-class margin $\gamma$ in (P3) must be sufficiently large for conditions \eqref{eq:cond_L_gamma} and \eqref{eq:cond_L_gamma_retr} to be satisfied. Since it is difficult to formulate the maximization of the exact value of $\gamma$, we relax this problem to the maximization of 
 \[
  \sum_{\p=1}^\V \sum_{i,j=1}^{\matP{N}} (\matP{W_b})_{ij} \,  \|  \matP{y_i} - \matP{y_j} \|^2 
  = \tr( \tilde{Y}^T  \tilde{L}_b \tilde{Y} )
 % =   \sum_{\p=1}^\V \tr \Big( {\matP{Y}}^T \matP{L_b} \matP{Y} \Big)
\]
which aims to increase the separation between the samples from different classes within each modality $\p$. Here the matrix $ \matP{W_b} \in \R^{\matP{N} \times \matP{N}}$ has entries $( \matP{W_b})_{ij} =1 $ when $\matP{x_i}$ and $\matP{x_j}$ are from different classes; and $( \matP{W_b})_{ij} =0$, otherwise.  The block-diagonal matrix $\tilde{L}_b \in \R^{\N \times \N}$ contains the between-class Laplacian $\matP{L_b} = \matP{D_b} - \matP{W_b}$ in its $\p$-th block, where $ \matP{D_b} $ is the diagonal between-class degree matrix with $i$-th diagonal entry given by $ \sum_j (\matP{W_b})_{ij} $.

\textbf{Cross-modal alignment.} Finally, the constant $ \B $ in property (P1) in Theorems \ref{thm:mm_class_bound} and \ref{thm:cm_retrieval_bnd} should be sufficiently small for conditions  \eqref{eq:cond_L_gamma} and \eqref{eq:cond_L_gamma_retr} to be met. The parameter $ \B $ represents the distance between the embeddings of the observations of the same sample in different modalities. We relax the minimization of $\B$ to the minimization of the following term, which aims to embed samples  of high affinity from different modalities $\p$, $\q$ into nearby points 
%
%compute nonlinear mappings of data samples from different modalities into a common space so that samples from the same classes across a pair of different modalities 
%
\[
   \sum_{\p=1}^\V \sum_{\q\neq \p}  \sum_{i=1}^{\matP{N}} \sum_{j=1}^{\matQ{N}} 
 \  (\matPQ{W_w})_{ij} \ \normSquare{\matP{y_i} -\matQ{y_j}}   
 = \tr( \tilde{Y}^T  \tilde {L}_{cw}  \tilde{Y} ).
\]
Here, the matrix $\matPQ{W_w} \in \R^{\matP{\N} \times \matQ{\N}}$ encodes the affinities between sample pairs from different modalities. $(\matPQ{W_w})_{ij} $ is nonzero only if $\matP{x_i}$ and $\matQ{x_j}$ are from the same class, in which case it is computed with the Gaussian kernel based on the distance between $\matP{x_i}$ and $\matQ{x_j}$  when transferred to a common modality (i.e., using  $\| \matP{x_i} - \matP{x_j} \|$ or $\| \matQ{x_i} - \matQ{x_j} \|$, otherwise  $\| x_i^{(r)} - x_j^{(r)} \|$ in some other modality $r$ if the former ones are not possible). Denoting by $\tilde {W}_{cw} \in \R^{\N \times \N}$ the cross-modal within-class weight matrix containing $\matPQ{W_w}$ in its $(\p,\q)$-th block, the corresponding Laplacian matrix $\tilde {L}_{cw} \in \R^{\N \times \N}$ is computed as $\tilde {L}_{cw} =\tilde {D}_{cw} - \tilde {W}_{cw}  $, where  $\tilde {D}_{cw} $ is the diagonal degree matrix with $i$-th diagonal entry  given by $ \sum_j (\tilde {W}_{cw} )_{ij} $.

Meanwhile,  the property (P3) in Theorems \ref{thm:mm_class_bound} and \ref{thm:cm_retrieval_bnd} suggests that two samples from modalities  $\p$, $\q$ should be separated if they are from different classes. We thus propose to maximize 
\[
\sum_{\p=1}^\V \sum_{\q\neq \p} \sum_{i=1}^{\matP{N}} \sum_{j=1}^{\matQ{N}} 
 \ (\matPQ{W_b})_{ij} \ \normSquare{\matP{y_i} -\matQ{y_j}}    
= \tr(   \tilde{Y}^T  \tilde {L}_{cb}  \tilde{Y} )
\]
where the matrix $\matPQ{W_b} \in \R^{\matP{\N} \times \matQ{\N}}$ is formed by setting $(\matPQ{W_b})_{ij}=1$ if $\matP{x_i}$ and $\matQ{x_j}$ are from different classes, and 0 otherwise. The cross-modal between-class weight matrix $\tilde {W}_{cb} \in \R^{\N \times \N}$ contains the matrix  $\matPQ{W_b}$ in its $(\p,\q)$-th block, while $ \tilde {L}_{cb} \in \R^{\N \times \N}$ is the corresponding Laplacian matrix given by $\tilde {L}_{cb} = \tilde {D}_{cb} - \tilde {W}_{cb} $, with $\tilde {D}_{cb} $ denoting the diagonal degree matrix with $i$-th diagonal entry  given by $ \sum_j (\tilde {W}_{cb} )_{ij} $.

\textbf{Overall problem.} We now combine all these objectives in the following overall optimization problem
\begin{multline} \label{eqnOverallOptimization}
     \underset {  \tilde{Y} , \ \{ \matP{\sigma} \} }  {\text{minimize }} 
       \tr( \tilde{Y}^T  \tilde{L}_w \tilde{Y} )
	- 
      \mu_{1} \, \tr( \tilde{Y}^T  \tilde{L}_b \tilde{Y} ) 
      +
      \mu_{2} \, \tr( \tilde{Y}^T  \tilde{\Psi} ^{-2}  \tilde{Y} )  \\
      +
      \mu_{3} \,       \sum_{\p=1}^{\V} ( {\matP{\sigma})}^{-2}   
     + 
      \mu_{4} 
     \tr( \tilde{Y}^T  \tilde {L}_{cw}  \tilde{Y} )
     - 
      \mu_{5} 
      \tr(   \tilde{Y}^T  \tilde {L}_{cb}  \tilde{Y} )
%      \text{subject to }  {\matP{Y}}^T\matP{Y} = I 
\end{multline}
%
%%
%\begin{multline} \label{eqnOverallOptimization}
%     \underset { \{ \matP{Y} \}, \ \{ \matP{\sigma} \} }  {\text{minimize}} 
%      \sum_{p=1}^{V} 
%      \bigg\{ \tr\Big( {\matP{Y}}^T \matP{L_w} \matP{Y} \Big)  - 
%      \mu_{1} \, \tr \Big( {\matP{Y}}^T \matP{L_b} \matP{Y} \Big) \\
%      +
%      \mu_{2} \, \tr \Big( {\matP{Y}}^T (\matP{\Psi})^{-2} \ \matP{Y} \Big) +
%      \mu_{3} \, ( {\matP{\sigma})}^{-2}   
%      \bigg\} \\
%     + 
%   \sum_{p=1}^V \sum_{q\neq p}
%      \ \bigg\{
%      \mu_{4}
% \sum_{i=1}^{\matP{N}} \sum_{j=1}^{\matQ{N}} 
% \  \normSquare{\matP{y_i} -\matQ{y_j}}  (\matPQ{W_w})_{ij}      
%      \\
%     - 
%      \mu_{5} 
%      \sum_{i=1}^{\matP{N}} \sum_{j=1}^{\matQ{N}} 
%\normSquare{\matP{y_i} -\matQ{y_j}}     (\matPQ{W_b})_{ij}     
%      \bigg\}  
%%      \text{subject to }  {\matP{Y}}^T\matP{Y} = I 
%\end{multline}
%
subject to $  \tilde{Y}^T  \tilde{Y} = I $, where $\mu_1, \dots, \mu_5$ are positive weight parameters, $I \in \R^{\dnew \times \dnew}$ is the identity matrix, and the optimization constraint $  \tilde{Y}^T  \tilde{Y} = I $ is for the normalization of the learnt coordinates.  

\subsection{Solution of the Optimization Problem}
\label{ssec:soln_prob}

%We first rewrite the problem in  \eqref{eqnOverallOptimization} in a more compact form. Let 
%%
%%\begin{multline} \label{eqnCompact}
%%    \tilde{Y} = 
%%        \begin{bmatrix} 
%%            y_1^{(1)} y_2^{(1)} \hdots y_{N_p}^{(\p)} \hdots y_{N_V}^{(V)} 
%%        \end{bmatrix}^T, 
%%\end{multline}
%%
%$
%    \tilde{Y} = [  y_1^{(1)} y_2^{(1)} \hdots y_{N_p}^{(\p)} \hdots y_{N_V}^{(V)}  ]^T
%$
%%
%denote the matrix containing the embeddings from all modalities. Let us also define $\tilde{\Psi}, \tilde{L}_w$, and  $\tilde{L}_b $ as the block diagonal matrices containing, respectively, the kernel matrix $\Psi^{(\p)}$, the within-class Laplacian $ \matP{L_w}$, and the between-class Laplacian  $ \matP{L_b}$ in their $p$-th block \cite{mnse}. Next, let $\tilde {W}_{cw}$ and $\tilde {W}_{cb}$ denote the cross-modal within-class and between-class weight matrices obtained, respectively, by tiling the matrices $\matPQ{W_w}$ and $\matPQ{W_b}$  in their $(p,q)$-th block \cite{mnse}.
%%
%We can then define the corresponding Laplacian matrices $\tilde {L}_{cw} =\tilde {D}_{cw} - \tilde {W}_{cw}  $ and $\tilde {L}_{cb} = \tilde {D}_{cb} - \tilde {W}_{cb} $, where $\tilde {D}_{cw} $ and $\tilde {D}_{cb}$ are the corresponding diagonal degree matrices obtained by summing up the entries of $\tilde {W}_{cw}$ and $\tilde {W}_{cb}$  in each row. 

Defining
\begin{equation}
\label{eq:defn_A_matrix}
A = \tilde{L}_w - \mu_{1}\tilde{L}_b + \mu_{2} \tilde{\Psi} ^{-2}   + \mu_{4} \tilde {L}_{cw}  - \mu_{5} \tilde {L}_{cb} 
\end{equation}
the problem in \eqref{eqnOverallOptimization} can be rewritten as
\begin{equation}
\label{eq:opt_prob_final}
 \underset {  \tilde{Y} , \ \{ \matP{\sigma} \} }  {\text{minimize }}  
  \tr (  \tilde{Y}^T A    \tilde{Y} ) 
  + \mu_3 \sum_{\p=1}^V  ( {\matP{\sigma})}^{-2} , 
  \text{ subject to }  {  \tilde{Y}^T  \tilde{Y} = I }.
\end{equation}

The above problem is not jointly convex in $ \tilde{Y} $ and $\{ \matP{\sigma} \}$, hence it is not easy to find its global optimum. We minimize the objective function with an iterative alternating optimization scheme, where we first optimize $ \tilde{Y} $ by fixing $\{ \matP{\sigma} \}$, and then optimize $\{ \matP{\sigma} \}$ by fixing $ \tilde{Y} $ in each iteration as follows.

%Terms in the equation (\ref{eqnOverallOptimization}) can be expressed as in equation (\ref{eqnCompact}):  
% --------------------------------------------------------------
% Compact Versions
% --------------------------------------------------------------

%
%Thus, $\tilde{Y}$ and $\matP{\sigma}$ can be optimized separately: \\

\textbf{Optimization of $\tilde{Y}$:} When $\{ \matP{\sigma} \}$ are fixed, the optimization problem in \eqref{eq:opt_prob_final} becomes
% When Individual optimization can be described as in equation (\ref{eqnYOptimization}):
% --------------------------------------------------------------
% Y Optimization Equation
% --------------------------------------------------------------
\begin{equation}
\label{eqnYOptimization}
 \underset {  \tilde{Y}  }  {\text{minimize }}  
  \tr (  \tilde{Y}^T A    \tilde{Y} ) 
  \text{ subject to }  {  \tilde{Y}^T  \tilde{Y} = I }.
\end{equation}
The solution to this problem is given by the $\dnew $ eigenvectors of the matrix $A$ corresponding to its smallest $\dnew$ eigenvalues.
%

%
%\begin{multline} \label{eqnYOptimization}
%    J(\tilde{Y})= \underset {\tilde{Y}} {arg min}  
%      \ tr \Big( {\tilde{Y}}^T A {\tilde{Y}} \Big) \ ve \ 
%      \tilde{Y}^T\tilde{Y} = I, \\
%    A = \tilde{L_w} - \mu_{1}\tilde{L_b} + \mu_{2}\tilde{\Psi} + \mu_{4}\matPQ{\tilde{L}}_w - \mu_{5}\matPQ{\tilde{L}}_b 
%\end{multline}
%%
%\par Solutions of the equation (\ref{eqnYOptimization}) are eigenvectors of matrix $A$ that correspond the smallest $k$ eigenvalues. \\

\textbf{Optimization of $\{ \matP{\sigma} \}$:} Fixing $\tilde{Y}$, the problem \eqref{eq:opt_prob_final} becomes
\begin{equation}
\label{eq:sigma_opt}
% \label{eqnSigmaOptimization}
       \underset { \{ \matP{\sigma} \} } { \text{minimize }  }
      \ \mu_{2} \, \tr \big( {\tilde{Y}}^T \tilde{\Psi} ^{-2}  \tilde{Y} \big) +
      \mu_{3}  \sum_{\p=1}^{V}{ \left( {\matP{\sigma}} \right )}^{-2} .
\end{equation}

Note that the first term in the objective depends on the kernel scale parameters  $\{ \matP{\sigma} \}$  through the entries of the kernel matrix $\tilde{\Psi}$. Due to the block diagonal structure of $ \tilde{\Psi} $ and the separability of the second term, the objective \eqref{eq:sigma_opt} can be decomposed into $V$ individual objectives, each one of which is a function of only one scale parameter $ \matP{\sigma}$. We minimize these objective functions one by one, by optimizing one scale parameter  $ \matP{\sigma}$ at a time through exhaustive search. 

If $\mu_1$ and $\mu_5$ are sufficiently small, the matrix $A$ becomes positive semi-definite. In this case, the objective function is guaranteed to converge since it is nonnegative, and both updates on $\tilde{Y}$ and $ \{ \matP{\sigma} \}$  reduce it. We continue the iterations until the convergence of the objective.  We call the proposed algorithm  Multi-modal Nonlinear Supervised Embedding (MNSE), which is summarized in Algorithm \ref{alg:MNSE}.
\begin{algorithm}[h]

\scriptsize
\caption{ \footnotesize{\textbf{M}ulti-modal \textbf{N}onlinear \textbf{S}upervised \textbf{E}mbedding (MNSE)} }

\begin{algorithmic}

\STATE
\label{algInput}
\textbf{Input: } Training data matrices $\{ \matP{X} \}$ and training data labels \\

\STATE
\label{algInit}
\textbf{Initialization: } \\
Obtain the graph Laplacian matrices $\tilde{L}_w, \tilde{L}_b, \tilde {L}_{cw}$, and $\tilde {L}_{cb}$,
\\ Assign weight parameters $\{\mu_1, \mu_2, \cdots, \mu_5\}$,
and initial kernel scales $\matP{\sigma}$

\REPEAT

\STATE
\label{algObtainY}
Compute the nonlinear embeddings $\tilde{Y}$ through \eqref{eqnYOptimization} by fixing $\{ \matP{\sigma} \}$ 

\STATE
\label{algObtainSigma}
Compute the kernel scale parameters $\{ \matP{\sigma} \}$ through \eqref{eq:sigma_opt} by fixing $\tilde{Y}$

\UNTIL the maximum number of iterations or the convergence of the objective

\STATE
\label{algOutput}
\textbf{Output: } \\
Kernel scale parameters  $\matP{\sigma}$ and
projected training data $\matP{Y}$\\
Kernel coefficients $\matP{C} = \big(\matP{\Psi}\big)^{-1} \matP{Y}$ 

\end{algorithmic}
\label{alg:MNSE}
\end{algorithm}

%

%\vspace{-0.5cm}
%

\subsection{Complexity Analysis}
The complexity of the proposed MNSE method is mainly determined by those of the problems \eqref{eqnYOptimization} and \eqref{eq:sigma_opt} repeated in the main loop of the algorithm. When computing the matrix $A$ in \eqref{eq:defn_A_matrix}, the matrices $\tilde{L}_w$,  $\tilde{L}_b$,  $\tilde{\Psi}$, $ \tilde {L}_{cw}$, and $ \tilde {L}_{cb}$ can be constructed with complexity not exceeding $O(\N^2)$, where $\N=\sum_{\p} \matP{\N}$ is the total number of observations from all modalities. The eigenvalue decomposition step in \eqref{eqnYOptimization} is of complexity $O(\N^3)$. In the optimization problem \eqref{eq:sigma_opt}, the evaluation of the objective for each $\matP{\sigma}  $ value requires $O( (\matP{\N})^3)$ operations in modality $\p$; hence, the total complexity of finding all  $\{ \matP{\sigma}  \}$ is smaller than  $O(\N^3)$. Therefore, the overall complexity of the algorithm is determined as $O(\N^3)$.

\section{EXPERIMENTAL RESULTS}
\label{sec:exp_res}

% --------------------------------------------------------------
% Classification
% --------------------------------------------------------------

\subsection{Data sets}
\label{ssec:data_sets}
The following data sets are used in the experiments.

\textit{The MIT-CBCL multi-view face data set} \cite{MITCBCL} contains face images of 10 participants captured under 36 illumination conditions and 9 different pose angles. Images with frontal and profile poses are used as Modality 1 and Modality 2, respectively. Images are converted to greyscale and downsampled to a resolution of $30 \times 30$ pixels. The experiments are repeated 10 times by randomly dividing the data set into training and test images.

\textit{The Multi-PIE multi-view face data set} \cite{GrossMCKB10} consists of face images of many participants under varying camera angles and facial expressions. We conduct our experiments on a cropped and reduced version of this data set \cite{TianPZZM18}, where the images of 120 participants captured under 6 camera angles, 20 lighting conditions, and 2 facial expressions (neutral and smiling) are used. Greyscale images of resolution $32\times 32$ pixels are used. Modality 1 and Modality 2  are respectively chosen as the frontal camera angle ($0$\degree) and the five other camera angles ($15$\degree, $30$\degree, $45$\degree, $60$\degree, $75$\degree). Four experimentation settings are prepared. In each setting, the images of the participants under either the first 10 or the last 10 lighting conditions, and either the neutral or the smiling facial expression are used for training; and the rest of the images are used for test. The results are averaged over these four experimentation settings.

In the classification experiments with the MIT-CBCL and the Multi-PIE data sets, embedding parameters are learnt using the training images from both modalities. Test images are assumed to be available in only one modality and mapped to the common domain with the learnt embeddings. The class labels of test images are estimated via NN classification on the embeddings of the training images of their own modality.

\textit{The Wikipedia image-text data set} \cite{wiki} contains 2866 image-text pairs describing the contents of Wikipedia articles, which are categorized into 10 classes. The image-text pairs are randomly divided into 1300 training and 1566 test pairs in each trial of the experiments and the results are averaged over 10 trials. 128-dimensional SIFT histogram features are used in the image modality, and 10-dimensional text features obtained with a latent Dirichlet allocation model are used in the text modality \cite{BleiNJ03}, \cite{WangY0W016}.

\textit{The Pascal VOC2007 image-text data set} \cite{pascalVoc2007} contains image-text pairs from 20 different object classes, where the experiments are done on 2808 training and 2841 test pairs whose images contain only one object. GIST feature vectors in the image modality and word count feature vectors in the text modality are used in the experiments. 

In the retrieval experiments with the Wikipedia and the Pascal VOC2007 data sets, embedding functions are learnt using the training set, and then the relevant matches of an image (text) query are searched in the embedding of the text (image) database based on cosine similarity. The precision and recall rates are computed as in \eqref{eq:prec_recall_defn} by considering a retrieved item relevant if it is from the same class as the query sample. The Mean Average Precision (MAP) scores of the methods are computed by averaging all average precision values over all query samples. When computing the MAP scores, for each query sample the number of retrieved items is set so as to retrieve all training samples relevant to it.

\subsection{Stabilization and Sensitivity Analysis of MNSE}
\label{ssec:sensit_analy}

We first study the stabilization of the proposed MNSE algorithm and its sensitivity to the algorithm parameters. The image classification performance of MNSE is analyzed on the MIT-CBCL and the Multi-PIE data sets. 100 training and 260 test images are used for the MIT-CBCL data set. The average of the misclassification errors of  Modalities 1 and 2 is reported. The retrieval performance of the algorithm is studied on the Wikipedia and the Pascal VOC2007 data sets. The MAP scores are averaged over the image and the text queries.

%
%\begin{figure}[b]
%%
%\begin{minipage}[b]{0.49\linewidth}
%  \centering
%  \centerline{\includegraphics[height=0.53cm]{figures/mit_modal1.pdf}}
%  \centerline{(a) Modality 1}\medskip
%\end{minipage}
%\hfill
%\begin{minipage}[b]{0.49\linewidth}
%  \centering
%  \centerline{\includegraphics[height=0.53cm]{figures/mit_modal2.pdf}}
%  \centerline{(b) Modality 2}\medskip
%\end{minipage}
%\vspace{-0.5cm}
%%
%\caption{Sample images from the MIT-CBCL face data set}
%\label{figFaceData}
%%
%%\vspace{-0.5cm}
%\end{figure}
%%
%
%
%

\begin{figure}[t]
\begin{minipage}[b]{0.49\linewidth}
  \centering
  \centerline{\includegraphics[width=4.3cm]{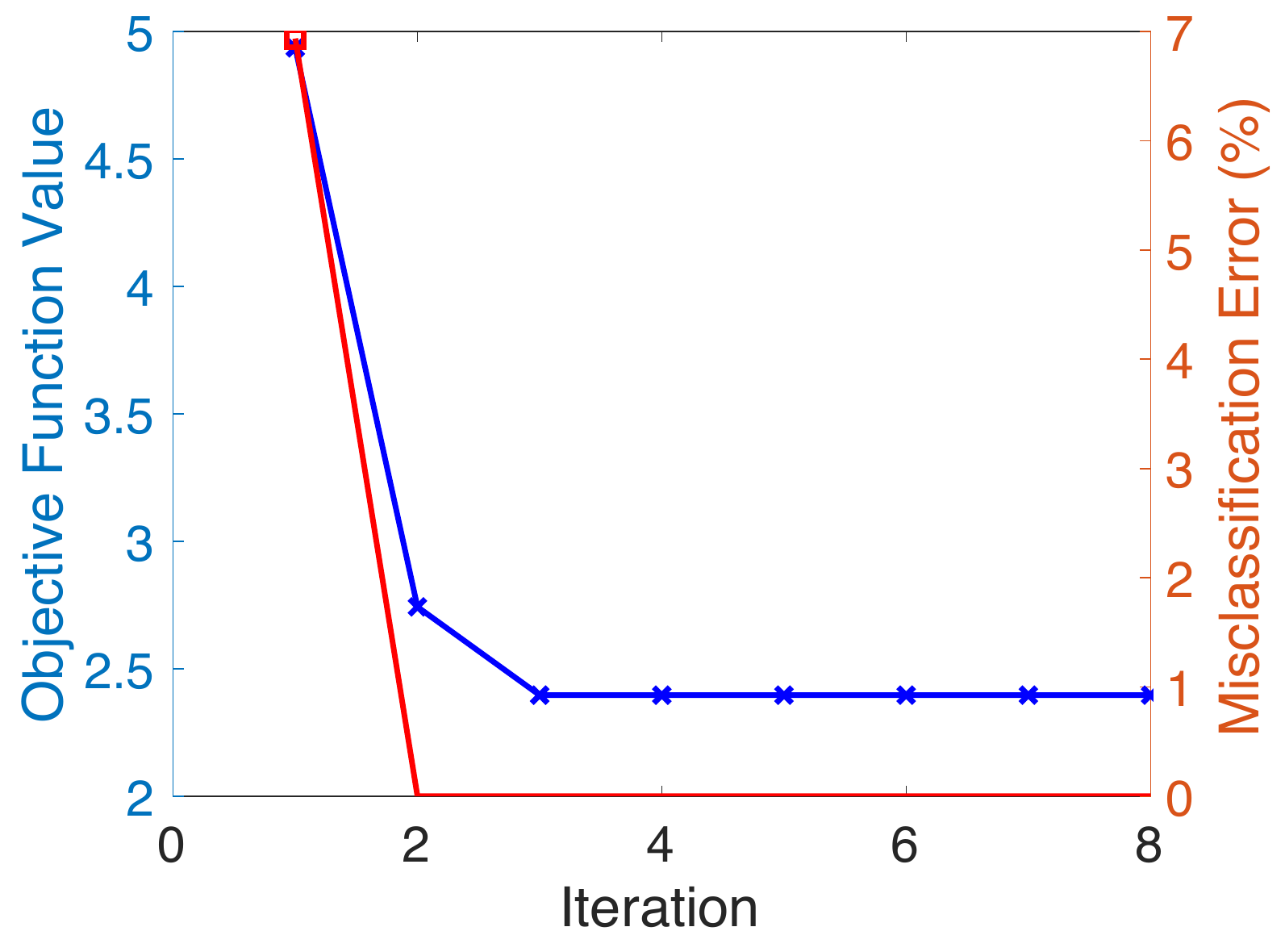}}
  \centerline{(a) MIT-CBCL}\medskip
\end{minipage}
\hfill
\begin{minipage}[b]{0.49\linewidth}
  \centering
  \centerline{\includegraphics[width=4.3cm]{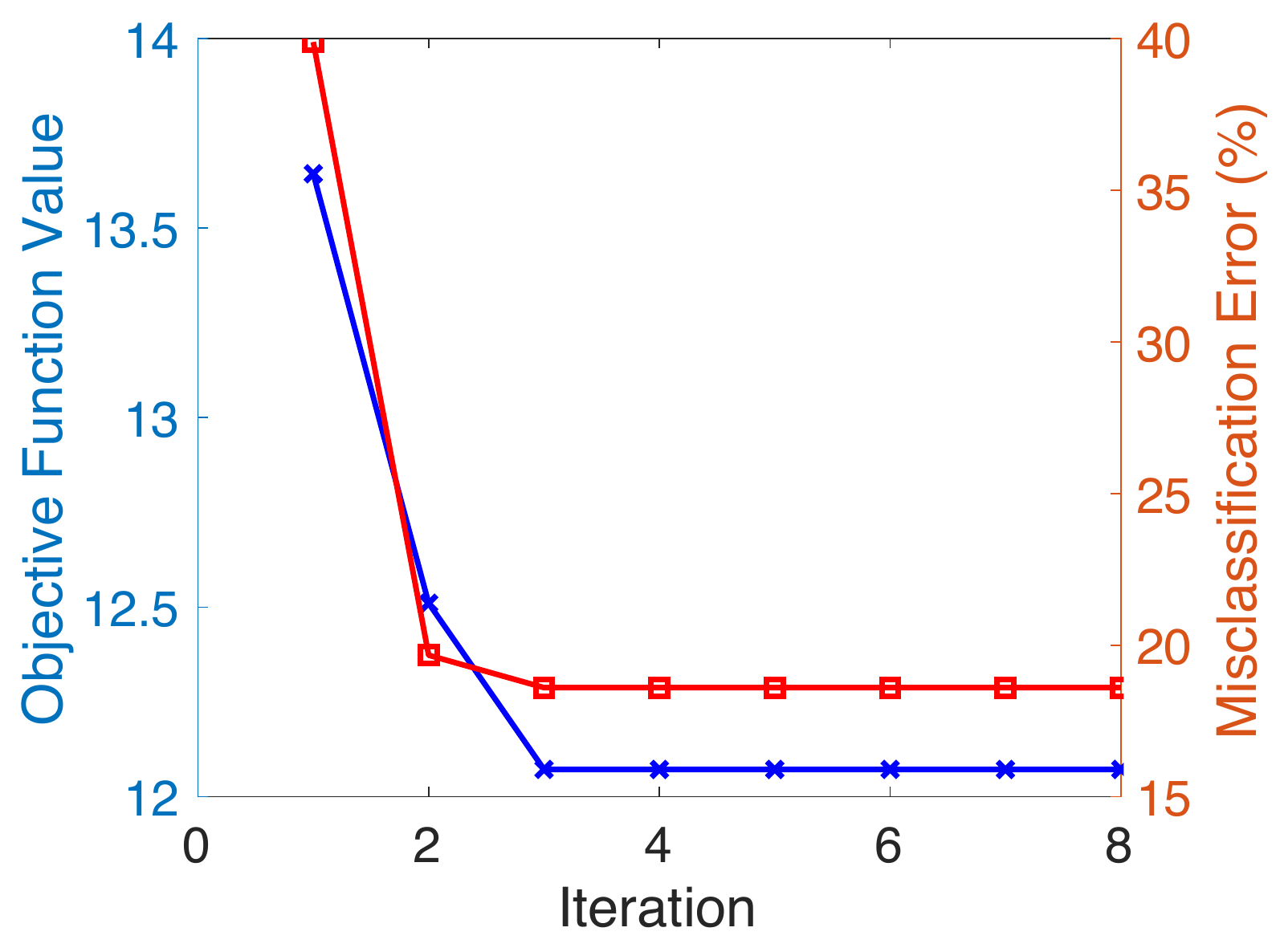}}
  \centerline{(b) Multi-PIE}\medskip
\end{minipage}
\hfill
\begin{minipage}[b]{0.49\linewidth}
  \centering
  \centerline{\includegraphics[width=4.3cm]{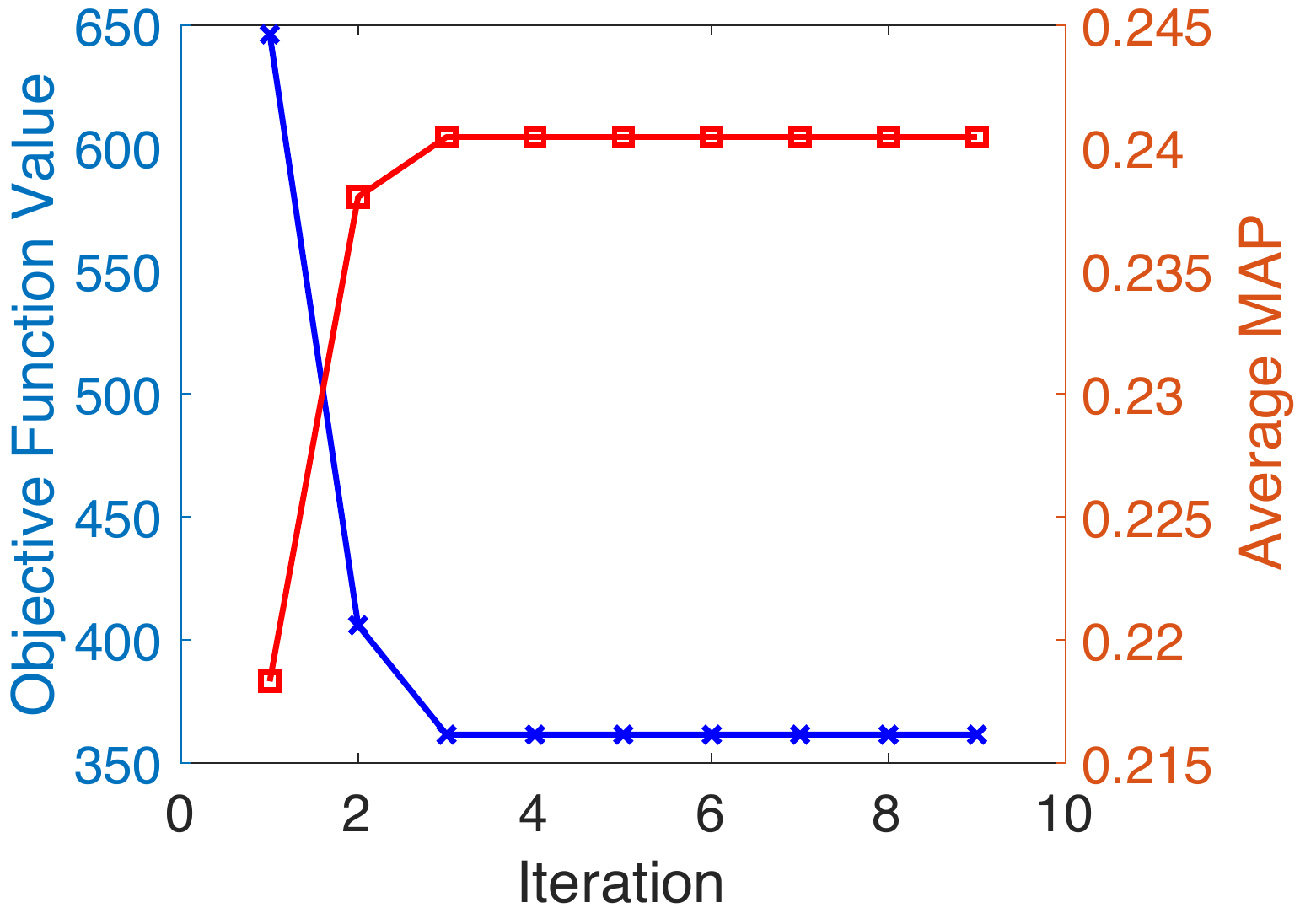}}
  \centerline{(c) Wikipedia}\medskip
\end{minipage}
\hfill
\begin{minipage}[b]{0.49\linewidth}
  \centering
  \centerline{\includegraphics[width=4.5cm]{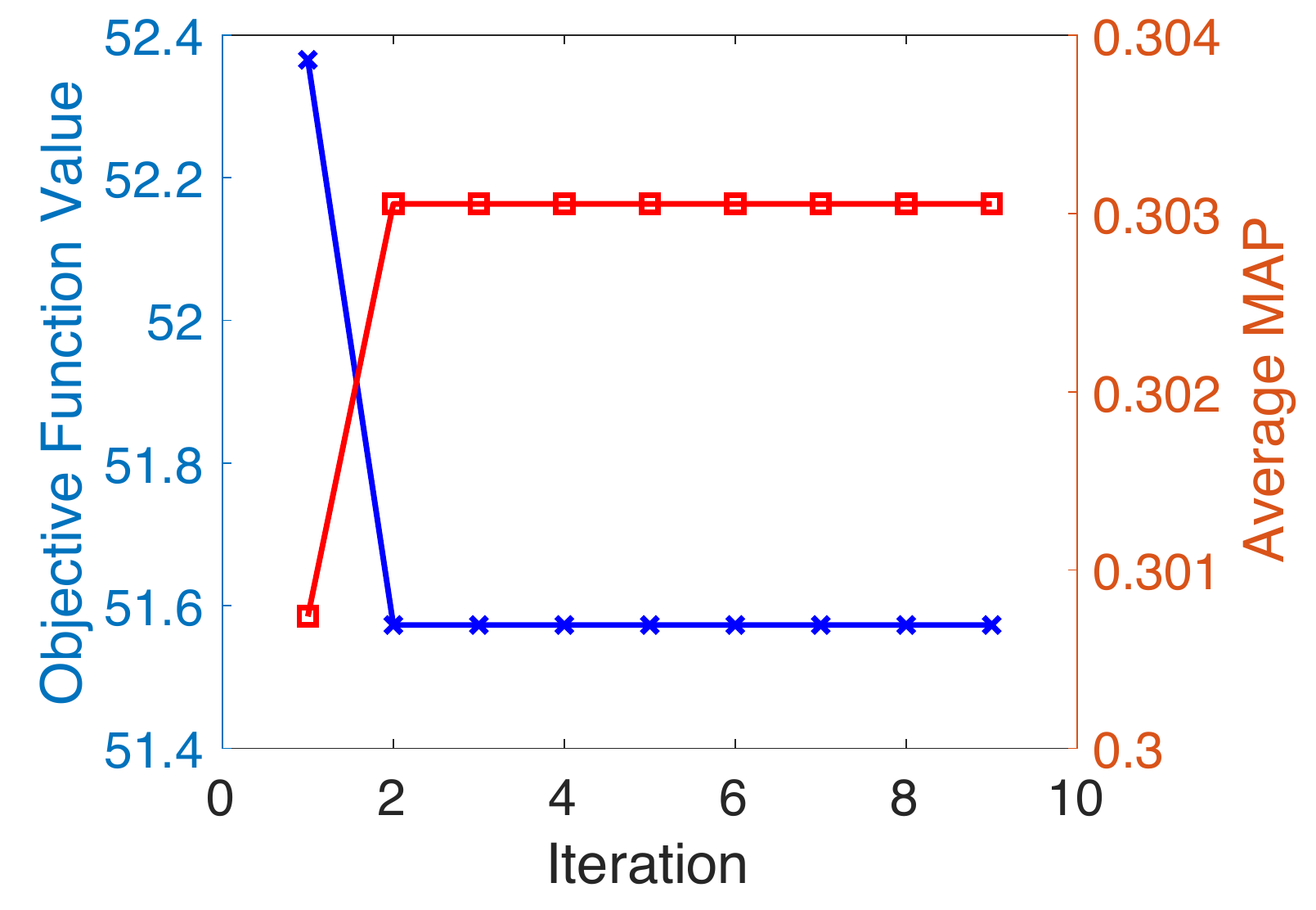}}
  \centerline{(d) Pascal VOC2007}\medskip
\end{minipage}
\caption{The evolution of the objective function and the algorithm performance during the iterations for all four data sets}
\label{figFaceObjective}
\end{figure}

We first study in Figure \ref{figFaceObjective} the evolution of the objective function \eqref{eqnOverallOptimization} along with the classification and the retrieval performance of MNSE throughout the optimization iterations on all four data sets.  The objective function is seen to steadily decrease throughout the iterations as expected. The updates on both the embeddings $\{\matP{Y} \}$ and the kernel scale parameters  $ \{ \matP{\sigma} \}$ ensure that the objective function is non-increasing. The improvement in the misclassification errors or MAP scores follows the decrease in the objective during the iterations. This suggests that the proposed objective function is indeed well-representative of the performance of the algorithm.

%shows that the misclassification MNSE algorithm rapidly converges to its optimum point for each modality. It results from the fact that the decrease in the overall cost function also leads to a decrease in the misclassification error. Thus, it can be inferred that the objective function is indeed well representative of the classification performance.

 %The variation of the optimization objective function throughout the iterations can be seen in Figure \ref{figFaceObjective}. The term "update" indicates update of the embeddings ($Y$) and update of the kernel scale parameter ($\sigma$) so that there exists two updates in each algorithm iteration. 

%

\vspace{0pt}
%%% MU-STUDY : MITCBCL %%%%
\begin{table}[t]
\footnotesize
\caption{The variation of the misclassification error with the weight parameters for the MIT-CBCL data set. Fixed parameters are chosen as $\mu_1=\mu_4=\mu_5=10^2$ in upper table and $\mu_2=10^{-3}$, $\mu_3=1$ in lower table. }
\begin{tabular}{lc|cccccccc}
%&& \multicolumn{8}{c}{$\mu_2$}\\
&  \hspace{-20pt}$\mu_3 \backslash \mu_2$ & 0   &  $10^{-3}$   &  $10^{-2}$    & $10^{-1}$   & $1$     &  $10^{1}$  &  $10^{2}$ & $10^{3}$  \\ \hline
%\rotatebox{90}{\multirow{10}*{\vspace{150pt}\hspace{-50pt} }}  
%
& \vspace{0pt} \hspace{-20pt}$0$  & \hspace{0pt}0.73 & \hspace{-10pt}0.35  &  \hspace{-10pt}0.35  &  \hspace{-10pt}0.35  &  \hspace{-5pt}0.35 &  \hspace{-5pt}0.35  &  \hspace{-5pt}1.5  &  \hspace{-5pt}10.7  \\
& \vspace{0pt} \hspace{-20pt}$10^{-3}$  & \hspace{0pt}0.16 & \hspace{-10pt}0.27  &  \hspace{-10pt}0.30  &  \hspace{-10pt}0.33  &  \hspace{-5pt}0.35 &  \hspace{-5pt}0.35  &  \hspace{-5pt}1.5  &  \hspace{-5pt}10.7  \\
& \vspace{0pt} \hspace{-20pt}$10^{-2}$  & \hspace{0pt}0.16 & \hspace{-10pt}0.23  &  \hspace{-10pt}0.27  &  \hspace{-10pt}0.30  &  \hspace{-5pt}0.33 &  \hspace{-5pt}0.35  &  \hspace{-5pt}1.51  &  \hspace{-5pt}10.6  \\
& \vspace{0pt} \hspace{-20pt}$10^{-1}$  & \hspace{0pt}0.16 & \hspace{-10pt}0.20  &  \hspace{-10pt}0.23  &  \hspace{-10pt}0.26  &  \hspace{-5pt}0.29 &  \hspace{-5pt}0.31  &  \hspace{-5pt}1.51  &  \hspace{-5pt}10.6  \\
& \vspace{0pt} \hspace{-20pt}$1$  & \hspace{0pt}0.16 & \hspace{-10pt}0.18  &  \hspace{-10pt}0.21  &  \hspace{-10pt}0.24  &  \hspace{-5pt}0.27 &  \hspace{-5pt}0.29  &  \hspace{-5pt}1.34  &  \hspace{-5pt}10.1  \\
& \vspace{0pt} \hspace{-20pt}$10^{1}$  & \hspace{0pt}0.16 & \hspace{-10pt}0.19  &  \hspace{-10pt}0.22  &  \hspace{-10pt}0.23  &  \hspace{-5pt}0.27 &  \hspace{-5pt}0.31  &  \hspace{-5pt}1.52  &  \hspace{-5pt}9.75  \\
& \vspace{0pt} \hspace{-20pt}$10^{2}$  & \hspace{0pt}0.16 & \hspace{-10pt}0.19  &  \hspace{-10pt}0.22  &  \hspace{-10pt}0.24  &  \hspace{-5pt}0.36 &  \hspace{-5pt}0.91  &  \hspace{-5pt}5.00  &  \hspace{-5pt}11.4  \\
& \vspace{0pt} \hspace{-20pt}$10^{3}$  & \hspace{0pt}0.16 & \hspace{-10pt}0.19  &  \hspace{-10pt}0.22  &  \hspace{-10pt}0.27  &  \hspace{-5pt}2.21 &  \hspace{-5pt}7.28  &  \hspace{-5pt}10.5  &  \hspace{-5pt}12.9  \\
\\
\end{tabular}
%
%\vspace{0.5cm}
%
\begin{tabular}{lc|cccccccc}
&  \hspace{-40pt}$\mu_4  \backslash \ \mu_1,\mu_5$ & \hspace{-0pt} 0   &  $10^{-3}$   &  $10^{-2}$    & $10^{-1}$   & $1$     &  $10^{1}$  &  $10^{2}$ & $10^{3}$  \\ \hline
& \vspace{0pt} \hspace{-20pt}$0$  & \hspace{0pt}7.94  & \hspace{-5pt}6.75  &  \hspace{-5pt}6.23  &  \hspace{-5pt}4.68  & \hspace{-10pt}1.42  &  \hspace{-10pt}0.24  &  \hspace{-10pt}0.21  &  \hspace{-5pt}0.18  \\
& \vspace{0pt} \hspace{-20pt}$10^{-3}$  & \hspace{0pt}7.61 & \hspace{-10pt}5.26  &  \hspace{-10pt}5.28  &  \hspace{-10pt}4.63  &  \hspace{-5pt}1.43 &  \hspace{-5pt}0.24  &  \hspace{-5pt}0.21  &  \hspace{-5pt}0.18  \\
& \vspace{0pt} \hspace{-20pt}$10^{-2}$  & \hspace{0pt}6.88 & \hspace{-10pt}5.83  &  \hspace{-10pt}5.88  &  \hspace{-10pt}3.38  &  \hspace{-5pt}1.34 &  \hspace{-5pt}0.24  &  \hspace{-5pt}0.21  &  \hspace{-5pt}0.18  \\
& \vspace{0pt} \hspace{-20pt}$10^{-1}$  & \hspace{0pt}2.92 & \hspace{-10pt}2.62  &  \hspace{-10pt}1.62  &  \hspace{-10pt}1.90  &  \hspace{-5pt}0.86 &  \hspace{-5pt}0.24  &  \hspace{-5pt}0.21  &  \hspace{-5pt}0.18  \\
& \vspace{0pt} \hspace{-20pt}$1$  & \hspace{0pt}1.35 &  \hspace{-5pt}1.10  &  \hspace{-5pt}0.71  &  \hspace{-5pt}0.65 & \hspace{-10pt}0.24  &  \hspace{-10pt}0.23  &  \hspace{-10pt}0.20  &  \hspace{-5pt}0.18   \\
& \vspace{0pt} \hspace{-20pt}$10^{1}$  & \hspace{0pt}0.94 & \hspace{-10pt}0.77  &  \hspace{-10pt}0.59  &  \hspace{-10pt}1.00  &  \hspace{-5pt}0.22 &  \hspace{-5pt}0.22  &  \hspace{-5pt}0.20  &  \hspace{-5pt}0.18  \\
& \vspace{0pt} \hspace{-20pt}$10^{2}$  & \hspace{0pt}0.86 & \hspace{-10pt}0.81  &  \hspace{-10pt}0.75  &  \hspace{-10pt}0.76  &  \hspace{-5pt}0.18 &  \hspace{-5pt}0.18  &  \hspace{-5pt}0.18  &  \hspace{-5pt}0.18  \\
& \vspace{0pt} \hspace{-20pt}$10^{3}$  & \hspace{0pt}0.78 & \hspace{-10pt}0.75  &  \hspace{-10pt}0.66  &  \hspace{-10pt}0.62  &  \hspace{-5pt}0.18 &  \hspace{-5pt}0.18  &  \hspace{-5pt}0.18  &  \hspace{-5pt}0.18  \\
\end{tabular}
\label{tab:mitcbcl_mu}
\end{table}

%%% MU-STUDY : MULTI-PIE %%%%
\begin{table}[]
\footnotesize
\caption{The variation of the misclassification error with the weight parameters for the Multi-PIE data set. Fixed parameters are chosen as $\mu_1=\mu_4=\mu_5=10^2$ in upper table and $\mu_2=10^{-3}$, $\mu_3=1$ in lower table. }
\begin{tabular}{lc|cccccccc}
%&& \multicolumn{8}{c}{$\mu_2$}\\
&  \hspace{-20pt}$\mu_3 \backslash \mu_2$ & 0   &  $10^{-3}$   &  $10^{-2}$    & $10^{-1}$   & $1$     &  $10^{1}$  &  $10^{2}$ & $10^{3}$  \\ \hline
%\rotatebox{90}{\multirow{10}*{\vspace{150pt}\hspace{-50pt} }}  
%
& \vspace{0pt} \hspace{-20pt}$0$  & \hspace{0pt}70.5 & \hspace{-10pt}44.0  &  \hspace{-10pt}49.8  &  \hspace{-10pt}45.1  &  \hspace{-5pt}59.9 &  \hspace{-5pt}88.8  &  \hspace{-5pt}86.9  &  \hspace{-5pt}76.0  \\
& \vspace{0pt} \hspace{-20pt}$10^{-3}$  & \hspace{0pt}17.0 & \hspace{-10pt}35.6  &  \hspace{-10pt}49.8  &  \hspace{-10pt}45.1  &  \hspace{-5pt}59.9 &  \hspace{-5pt}88.8  &  \hspace{-5pt}86.9  &  \hspace{-5pt}76.0  \\
& \vspace{0pt} \hspace{-20pt}$10^{-2}$  & \hspace{0pt}17.0 & \hspace{-10pt}18.5  &  \hspace{-10pt}42.4  &  \hspace{-10pt}48.0  &  \hspace{-5pt}58.9 &  \hspace{-5pt}88.8  &  \hspace{-5pt}86.9  &  \hspace{-5pt}76.0  \\
& \vspace{0pt} \hspace{-20pt}$10^{-1}$  & \hspace{0pt}17.0 & \hspace{-10pt}17.4  &  \hspace{-10pt}24.5  &  \hspace{-10pt}39.6  &  \hspace{-5pt}59.3 &  \hspace{-5pt}88.8  &  \hspace{-5pt}86.9  &  \hspace{-5pt}76.0  \\
& \vspace{0pt} \hspace{-20pt}$1$  & \hspace{0pt}17.0 & \hspace{-10pt}15.3  &  \hspace{-10pt}19.8  &  \hspace{-10pt}26.8  &  \hspace{-5pt}53.8 &  \hspace{-5pt}88.8  &  \hspace{-5pt}86.9  &  \hspace{-5pt}76.0  \\
& \vspace{0pt} \hspace{-20pt}$10^{1}$  & \hspace{0pt}17.0 & \hspace{-10pt}15.2  &  \hspace{-10pt}16.4  &  \hspace{-10pt}21.4  &  \hspace{-5pt}50.7 &  \hspace{-5pt}88.8  &  \hspace{-5pt}86.9  &  \hspace{-5pt}76.0  \\
& \vspace{0pt} \hspace{-20pt}$10^{2}$  & \hspace{0pt}17.0 & \hspace{-10pt}16.8  &  \hspace{-10pt}16.6  &  \hspace{-10pt}17.1  &  \hspace{-5pt}28.0 &  \hspace{-5pt}80.0  &  \hspace{-5pt}83.8  &  \hspace{-5pt}62.7  \\
& \vspace{0pt} \hspace{-20pt}$10^{3}$  & \hspace{0pt}17.0 & \hspace{-10pt}17.4  &  \hspace{-10pt}18.1  &  \hspace{-10pt}19.4  &  \hspace{-5pt}21.1 &  \hspace{-5pt}43.2  &  \hspace{-5pt}54.4  &  \hspace{-5pt}52.8  \\
\\
\end{tabular}
\begin{tabular}{lc|cccccccc}
&  \hspace{-40pt}$\mu_4  \backslash \ \mu_1,\mu_5$ & \hspace{-0pt} 0   &  $10^{-3}$   &  $10^{-2}$    & $10^{-1}$   & $1$     &  $10^{1}$  &  $10^{2}$ & $10^{3}$  \\ \hline
& \vspace{0pt} \hspace{-20pt}$0$  & \hspace{0pt}29.9 & \hspace{-10pt}29.1  &  \hspace{-10pt}41.9  &  \hspace{-10pt}44.9  &  \hspace{-5pt}77.3 &  \hspace{-5pt}89.4  &  \hspace{-5pt}91.8  &  \hspace{-5pt}95.5  \\
& \vspace{0pt} \hspace{-20pt}$10^{-3}$  & \hspace{0pt}28.2 & \hspace{-10pt}28.3  &  \hspace{-10pt}40.2  &  \hspace{-10pt}43.6  &  \hspace{-5pt}77.5 &  \hspace{-5pt}89.4  &  \hspace{-5pt}91.8  &  \hspace{-5pt}95.5  \\
& \vspace{0pt} \hspace{-20pt}$10^{-2}$  & \hspace{0pt}22.1 & \hspace{-10pt}22.2  &  \hspace{-10pt}20.7  &  \hspace{-10pt}31.5  &  \hspace{-5pt}75.9 &  \hspace{-5pt}89.4  &  \hspace{-5pt}91.8  &  \hspace{-5pt}95.5  \\
& \vspace{0pt} \hspace{-20pt}$10^{-1}$  & \hspace{0pt}17.3 & \hspace{-10pt}17.5  &  \hspace{-10pt}17.9  &  \hspace{-10pt}18.3  &  \hspace{-5pt}67.9 &  \hspace{-5pt}89.9  &  \hspace{-5pt}92.3  &  \hspace{-5pt}95.9  \\
& \vspace{0pt} \hspace{-20pt}$1$  & \hspace{0pt}16.44 & \hspace{-10pt}16.5  &  \hspace{-10pt}17.6  &  \hspace{-10pt}24.1  &  \hspace{-5pt}43.3 &  \hspace{-5pt}84.7  &  \hspace{-5pt}88.9  &  \hspace{-5pt}96.3  \\
& \vspace{0pt} \hspace{-20pt}$10^{1}$  & \hspace{0pt}16.0 & \hspace{-10pt}16.1  &  \hspace{-10pt}16.2  &  \hspace{-10pt}21.7  &  \hspace{-5pt}21.8 &  \hspace{-5pt}16.7  &  \hspace{-5pt}48.3  &  \hspace{-5pt}63.4  \\
& \vspace{0pt} \hspace{-20pt}$10^{2}$  & \hspace{0pt}16.0 & \hspace{-10pt}16.0  &  \hspace{-10pt}16.2  &  \hspace{-10pt}22.0  &  \hspace{-5pt}22.2 &  \hspace{-5pt}23.1  &  \hspace{-5pt}15.3  &  \hspace{-5pt}39.6  \\
& \vspace{0pt} \hspace{-20pt}$10^{3}$  & \hspace{0pt}16.0 & \hspace{-10pt}16.1  &  \hspace{-10pt}16.1  &  \hspace{-10pt}21.0  &  \hspace{-5pt}21.1 &  \hspace{-5pt}22.6  &  \hspace{-5pt}24.0  &  \hspace{-5pt}15.5  \\
\end{tabular}
\label{tab:multipie_mu}
\end{table}

Next, the effect of the weight parameters $\mu_1, \mu_2, \mu_3, \mu_4 , \mu_5$ on the algorithm performance is examined in Tables \ref{tab:mitcbcl_mu}-\ref{tab:pascal_mu} for all four data sets. The performance on the Pascal VOC2007 and Wikipedia data sets is measured on half of the training samples  assigned for validation. The upper half of each table shows the variation of the average performance with $\mu_{2}$ and $\mu_{3}$, and  the lower half of each table shows the variation with $\mu_{4}$ and $\mu_{1}=\mu_{5}$. The parameters $\mu_{1}$ and $\mu_{5}$ are set to be equal, motivated by the similarity in the construction of the between-class separation matrices associated with these parameters. Tables \ref{tab:mitcbcl_mu} and \ref{tab:multipie_mu} show that setting $\mu_{2}= 10^{-3}$ and choosing $\mu_3$ in the interval $[1,10^3]$ leads to reasonably small misclassification error in both data sets. A suitable choice for $\mu_1$, $\mu_4$, and  $\mu_5$ seems to be $\mu_4=\mu_1=\mu_5 \in [10^2, 10^3]$. On the other hand, the optimal parameter ranges are slightly different in the retrieval experiments. Choosing $\mu_2 = 1$ and $\mu_3 \in [1,10^3]$ maximizes the MAP score for both the Wikipedia and the Pascal VOC2007 data sets, while setting $\mu_4=100 \mu_1= 100 \mu_5$ and selecting $\mu_4 \in [1, 10]$ gives close to optimal performance. As an overall conclusion, the observation that $\mu_2$ and $\mu_3$ should have nonzero values in both classification and retrieval experiments confirms that the Lipschitz regularity terms in the objective function are necessary for high performance, which is a central idea in our study. Next, it may be more advantageous to select $\mu_4$ relatively higher than $\mu_1$ and $\mu_5$ in retrieval experiments, unlike in classification where they can be chosen equal. This suggests  that the alignment across different modalities may become even more critical in retrieval problems.

%%% MU-STUDY : WIKIPEDIA%%%%
\begin{table}[t]
\footnotesize
\caption{The variation of the MAP with the weight parameters for the Wikipedia data set. Fixed parameters are chosen as $\mu_1=\mu_4=\mu_5=10^{-3}$ in upper table and $\mu_2=\mu_3=1$ in lower table. }
\begin{tabular}{lc|cccccccc}
%&& \multicolumn{8}{c}{$\mu_2$}\\
&  \hspace{-20pt}$\mu_3 \backslash \mu_2$ & 0   &  $10^{-3}$   &  $10^{-2}$    & $10^{-1}$   & $1$     &  $10^{1}$  &  $10^{2}$ & $10^{3}$  \\ \hline
%\rotatebox{90}{\multirow{10}*{\vspace{150pt}\hspace{-50pt} }}  
%
& \vspace{0pt} \hspace{-20pt}$0$  & \hspace{0pt}0.10 & \hspace{-10pt}0.19  &  \hspace{-10pt}0.17  &  \hspace{-10pt}0.19  &  \hspace{-5pt}0.22 &  \hspace{-5pt}0.21  &  \hspace{-5pt}0.17  &  \hspace{-5pt}0.17    \\
& \vspace{0pt} \hspace{-20pt}$10^{-3}$  & \hspace{0pt}0.11 & \hspace{-10pt}0.18  &  \hspace{-10pt}0.17  &  \hspace{-10pt}0.19  &  \hspace{-5pt}0.22 &  \hspace{-5pt}0.21  &  \hspace{-5pt}0.17  &  \hspace{-5pt}0.17  \\
& \vspace{0pt} \hspace{-20pt}$10^{-2}$  & \hspace{0pt}0.11 & \hspace{-10pt}0.17  &  \hspace{-10pt}0.18  &  \hspace{-10pt}0.19  &  \hspace{-5pt}0.23 &  \hspace{-5pt}0.21  &  \hspace{-5pt}0.17  &  \hspace{-5pt}0.17  \\
& \vspace{0pt} \hspace{-20pt}$10^{-1}$  & \hspace{0pt}0.11 & \hspace{-10pt}0.17  &  \hspace{-10pt}0.20  &  \hspace{-10pt}0.22  &  \hspace{-5pt}0.23 &  \hspace{-5pt}0.21  &  \hspace{-5pt}0.17  &  \hspace{-5pt}0.17  \\
& \vspace{0pt} \hspace{-20pt}$1$  & \hspace{0pt}0.11 & \hspace{-10pt}0.20  &  \hspace{-10pt}0.21  &  \hspace{-10pt}0.22  &  \hspace{-5pt}0.23 &  \hspace{-5pt}0.22  &  \hspace{-5pt}0.18  &  \hspace{-5pt}0.17  \\
& \vspace{0pt} \hspace{-20pt}$10^{1}$  & \hspace{0pt}0.11 & \hspace{-10pt}0.20  &  \hspace{-10pt}0.21  &  \hspace{-10pt}0.22  &  \hspace{-5pt}0.23 &  \hspace{-5pt}0.22  &  \hspace{-5pt}0.17  &  \hspace{-5pt}0.17  \\
& \vspace{0pt} \hspace{-20pt}$10^{2}$  & \hspace{0pt}0.11 & \hspace{-10pt}0.20  &  \hspace{-10pt}0.21  &  \hspace{-10pt}0.22  &  \hspace{-5pt}0.23 &  \hspace{-5pt}0.22  &  \hspace{-5pt}0.19  &  \hspace{-5pt}0.18  \\
& \vspace{0pt} \hspace{-20pt}$10^{3}$  & \hspace{0pt}0.11 & \hspace{-10pt}0.20  &  \hspace{-10pt}0.21  &  \hspace{-10pt}0.22  &  \hspace{-5pt}0.23 &  \hspace{-5pt}0.22  &  \hspace{-5pt}0.20  &  \hspace{-5pt}0.18  \\
\\
\end{tabular}
\begin{tabular}{lc|cccccccc}
&  \hspace{-20pt}$\mu_4  \backslash \ \mu_1,\mu_5$ & \hspace{-10pt}0   &  $10^{-3}$   &  $10^{-2}$    & $10^{-1}$   & $1$     &  $10^{1}$  &  $10^{2}$ & $10^{3}$  \\ \hline
& \vspace{0pt} \hspace{-20pt}$0$ & \hspace{0pt}0.13 & \hspace{-10pt}0.19  &  \hspace{-10pt}0.19  &  \hspace{-10pt}0.18  &  \hspace{-5pt}0.11 &  \hspace{-5pt}0.13  &  \hspace{-5pt}0.13  &  \hspace{-5pt}0.13  \\
& \vspace{0pt} \hspace{-20pt}$10^{-3}$  & \hspace{0pt}0.22 & \hspace{-10pt}0.23  &  \hspace{-10pt}0.20  &  \hspace{-10pt}0.18  &  \hspace{-5pt}0.11 &  \hspace{-5pt}0.13  &  \hspace{-5pt}0.13  &  \hspace{-5pt}0.13  \\
& \vspace{0pt} \hspace{-20pt}$10^{-2}$  & \hspace{0pt}0.22 & \hspace{-10pt}0.22  &  \hspace{-10pt}0.23  &  \hspace{-10pt}0.20  &  \hspace{-5pt}0.11 &  \hspace{-5pt}0.13  &  \hspace{-5pt}0.13  &  \hspace{-5pt}0.13  \\
& \vspace{0pt} \hspace{-20pt}$10^{-1}$  & \hspace{0pt}0.22 & \hspace{-10pt}0.22  &  \hspace{-10pt}0.22  &  \hspace{-10pt}0.23  &  \hspace{-5pt}0.11 &  \hspace{-5pt}0.13  &  \hspace{-5pt}0.13  &  \hspace{-5pt}0.13  \\
& \vspace{0pt} \hspace{-20pt}$1$  & \hspace{0pt}0.21 & \hspace{-10pt}0.21  &  \hspace{-10pt}0.21  &  \hspace{-10pt}0.21  &  \hspace{-5pt}0.11 &  \hspace{-5pt}0.12  &  \hspace{-5pt}0.13  &  \hspace{-5pt}0.13  \\
& \vspace{0pt} \hspace{-20pt}$10^{1}$  & \hspace{0pt}0.20 & \hspace{-10pt}0.20  &  \hspace{-10pt}0.20  &  \hspace{-10pt}0.20  &  \hspace{-5pt}0.20 &  \hspace{-5pt}0.11  &  \hspace{-5pt}0.13  &  \hspace{-5pt}0.13  \\
& \vspace{0pt} \hspace{-20pt}$10^{2}$  & \hspace{0pt}0.19 & \hspace{-10pt}0.19  &  \hspace{-10pt}0.19  &  \hspace{-10pt}0.19  &  \hspace{-5pt}0.19 &  \hspace{-5pt}0.18  &  \hspace{-5pt}0.11  &  \hspace{-5pt}0.13  \\
& \vspace{0pt} \hspace{-20pt}$10^{3}$  & \hspace{0pt}0.19 & \hspace{-10pt}0.19  &  \hspace{-10pt}0.19  &  \hspace{-10pt}0.19  &  \hspace{-5pt}0.19 &  \hspace{-5pt}0.18  &  \hspace{-5pt}0.18  &  \hspace{-5pt}0.11  \\
\end{tabular}
\label{tab:wiki_mu}
\end{table}

%%% MU-STUDY : PASCAL VOC%%%%
\begin{table}[]
\footnotesize
\caption{The variation of the MAP with the weight parameters for the Pascal VOC2007 data set. Fixed parameters are chosen as $\mu_1=\mu_5=10^{-1}, \mu_4=10$ in upper table and $\mu_2=\mu_3=1$ in lower table.}
\begin{tabular}{lc|cccccccc}
%&& \multicolumn{8}{c}{$\mu_2$}\\
&  \hspace{-20pt}$\mu_3 \backslash \mu_2$ & 0   &  $10^{-3}$   &  $10^{-2}$    & $10^{-1}$   & $1$     &  $10^{1}$  &  $10^{2}$ & $10^{3}$  \\ \hline
%\rotatebox{90}{\multirow{10}*{\vspace{150pt}\hspace{-50pt} }}  
%
& \vspace{0pt} \hspace{-20pt}$0$  & \hspace{0pt}0.21 & \hspace{-10pt}0.24  &  \hspace{-10pt}0.23  &  \hspace{-10pt}0.22  &  \hspace{-5pt}0.29 &  \hspace{-5pt}0.28  &  \hspace{-5pt}0.22  &  \hspace{-5pt}0.22  \\
& \vspace{0pt} \hspace{-20pt}$10^{-3}$  & \hspace{0pt}0.25 & \hspace{-10pt}0.25  &  \hspace{-10pt}0.24  &  \hspace{-10pt}0.23  &  \hspace{-5pt}0.29 &  \hspace{-5pt}0.28  &  \hspace{-5pt}0.22  &  \hspace{-5pt}0.22  \\
& \vspace{0pt} \hspace{-20pt}$10^{-2}$  & \hspace{0pt}0.25 & \hspace{-10pt}0.25  &  \hspace{-10pt}0.24  &  \hspace{-10pt}0.23  &  \hspace{-5pt}0.29 &  \hspace{-5pt}0.28  &  \hspace{-5pt}0.22  &  \hspace{-5pt}0.22  \\
& \vspace{0pt} \hspace{-20pt}$10^{-1}$  & \hspace{0pt}0.25 & \hspace{-10pt}0.25  &  \hspace{-10pt}0.24  &  \hspace{-10pt}0.25  &  \hspace{-5pt}0.29 &  \hspace{-5pt}0.28  &  \hspace{-5pt}0.23  &  \hspace{-5pt}0.22  \\
& \vspace{0pt} \hspace{-20pt}$1$  & \hspace{0pt}0.25 & \hspace{-10pt}0.25  &  \hspace{-10pt}0.24  &  \hspace{-10pt}0.25  &  \hspace{-5pt}0.30 &  \hspace{-5pt}0.28  &  \hspace{-5pt}0.24  &  \hspace{-5pt}0.22  \\
& \vspace{0pt} \hspace{-20pt}$10^{1}$  & \hspace{0pt}0.25 & \hspace{-10pt}0.25  &  \hspace{-10pt}0.24  &  \hspace{-10pt}0.25  &  \hspace{-5pt}0.30 &  \hspace{-5pt}0.28  &  \hspace{-5pt}0.24  &  \hspace{-5pt}0.24  \\
& \vspace{0pt} \hspace{-20pt}$10^{2}$  & \hspace{0pt}0.25 & \hspace{-10pt}0.25  &  \hspace{-10pt}0.24  &  \hspace{-10pt}0.25  &  \hspace{-5pt}0.30 &  \hspace{-5pt}0.28  &  \hspace{-5pt}0.24  &  \hspace{-5pt}0.24  \\
& \vspace{0pt} \hspace{-20pt}$10^{3}$  & \hspace{0pt}0.25 & \hspace{-10pt}0.25  &  \hspace{-10pt}0.24  &  \hspace{-10pt}0.25  &  \hspace{-5pt}0.30 &  \hspace{-5pt}0.28  &  \hspace{-5pt}0.24  &  \hspace{-5pt}0.23  \\
\\
\end{tabular}
\begin{tabular}{lc|cccccccc}
&  \hspace{-20pt}$\mu_4  \backslash \ \mu_1,\mu_5$ &  \hspace{-10pt}0   &  $10^{-3}$   &  $10^{-2}$    & $10^{-1}$   & $1$     &  $10^{1}$  &  $10^{2}$ & $10^{3}$  \\ \hline
& \vspace{0pt} \hspace{-20pt}$0$  & \hspace{0pt}0.10 & \hspace{-10pt}0.16  &  \hspace{-10pt}0.16  &  \hspace{-10pt}0.16  &  \hspace{-5pt}0.12 &  \hspace{-5pt}0.12  &  \hspace{-5pt}0.12  &  \hspace{-5pt}0.12  \\
& \vspace{0pt} \hspace{-20pt}$10^{-3}$  & \hspace{0pt}0.19 & \hspace{-10pt}0.20  &  \hspace{-10pt}0.19  &  \hspace{-10pt}0.16  &  \hspace{-5pt}0.12 &  \hspace{-5pt}0.12  &  \hspace{-5pt}0.12  &  \hspace{-5pt}0.12  \\
& \vspace{0pt} \hspace{-20pt}$10^{-2}$  & \hspace{0pt}0.19 & \hspace{-10pt}0.19  &  \hspace{-10pt}0.20  &  \hspace{-10pt}0.20  &  \hspace{-5pt}0.12 &  \hspace{-5pt}0.12  &  \hspace{-5pt}0.12  &  \hspace{-5pt}0.12  \\
& \vspace{0pt} \hspace{-20pt}$10^{-1}$  & \hspace{0pt}0.21 & \hspace{-10pt}0.21  &  \hspace{-10pt}0.21  &  \hspace{-10pt}0.22  &  \hspace{-5pt}0.11 &  \hspace{-5pt}0.11  &  \hspace{-5pt}0.12  &  \hspace{-5pt}0.12  \\
& \vspace{0pt} \hspace{-20pt}$1$  & \hspace{0pt}0.29 & \hspace{-10pt}0.29  &  \hspace{-10pt}0.29  &  \hspace{-10pt}0.30  &  \hspace{-5pt}0.10 &  \hspace{-5pt}0.12  &  \hspace{-5pt}0.12  &  \hspace{-5pt}0.12  \\
& \vspace{0pt} \hspace{-20pt}$10^{1}$  & \hspace{0pt}0.30 & \hspace{-10pt}0.30  &  \hspace{-10pt}0.30  &  \hspace{-10pt}0.30  &  \hspace{-5pt}0.25 &  \hspace{-5pt}0.06  &  \hspace{-5pt}0.09  &  \hspace{-5pt}0.09  \\
& \vspace{0pt} \hspace{-20pt}$10^{2}$  & \hspace{0pt}0.30 & \hspace{-10pt}0.30  &  \hspace{-10pt}0.30  &  \hspace{-10pt}0.30  &  \hspace{-5pt}0.25 &  \hspace{-5pt}0.23  &  \hspace{-5pt}0.06  &  \hspace{-5pt}0.08  \\
& \vspace{0pt} \hspace{-20pt}$10^{3}$  & \hspace{0pt}0.30 & \hspace{-10pt}0.30  &  \hspace{-10pt}0.30  &  \hspace{-10pt}0.30  &  \hspace{-5pt}0.25 &  \hspace{-5pt}0.25  &  \hspace{-5pt}0.23  &  \hspace{-5pt}0.06  \\
\end{tabular}
\label{tab:pascal_mu}
\end{table}

%%
%\begin{figure}[t]
%%
%\begin{minipage}[b]{0.49\linewidth}
%  \centering
%  \centerline{\includegraphics[width=4.3cm]{figures/faceU2U3VsError.pdf}}
%  \centerline{(a) Error vs $\mu_{2}, \mu_{3}$}\medskip
%\end{minipage}
%\hfill
%\begin{minipage}[b]{0.49\linewidth}
%  \centering
%  \centerline{\includegraphics[width=4.3cm]{figures/faceU1U4U5VsError.pdf}}
%  \centerline{(b) Error vs $\mu_{1}, \mu_{4}, \mu_{5}$}\medskip
%\end{minipage}
%% \vspace{-0.5cm}
%\caption{The variation of the misclassification error with the weight parameters $\mu_{1}, \mu_{2}, \mu_{3}, \mu_{4}, \mu_{5}$ for the MIT-CBCL data set}
%\label{figWeightPar}
%%
%\end{figure}
%

\begin{figure}[t]
\begin{minipage}[b]{0.49\linewidth}
  \centering
  \centerline{\includegraphics[width=4.3cm]{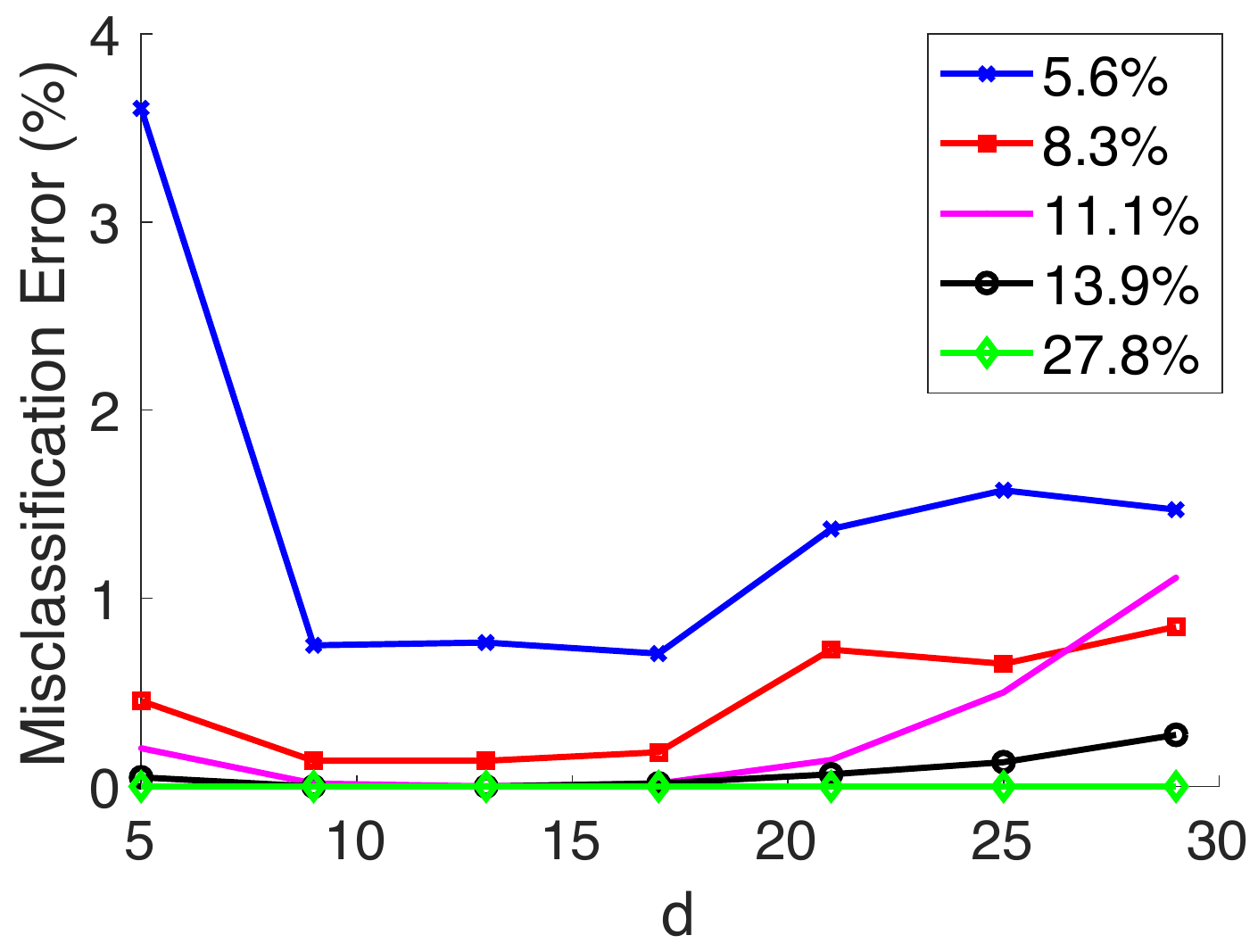}}
  \centerline{(a) MIT-CBCL}\medskip
\end{minipage}
\hfill
\begin{minipage}[b]{0.49\linewidth}
  \centering
  \centerline{\includegraphics[width=4.3cm]{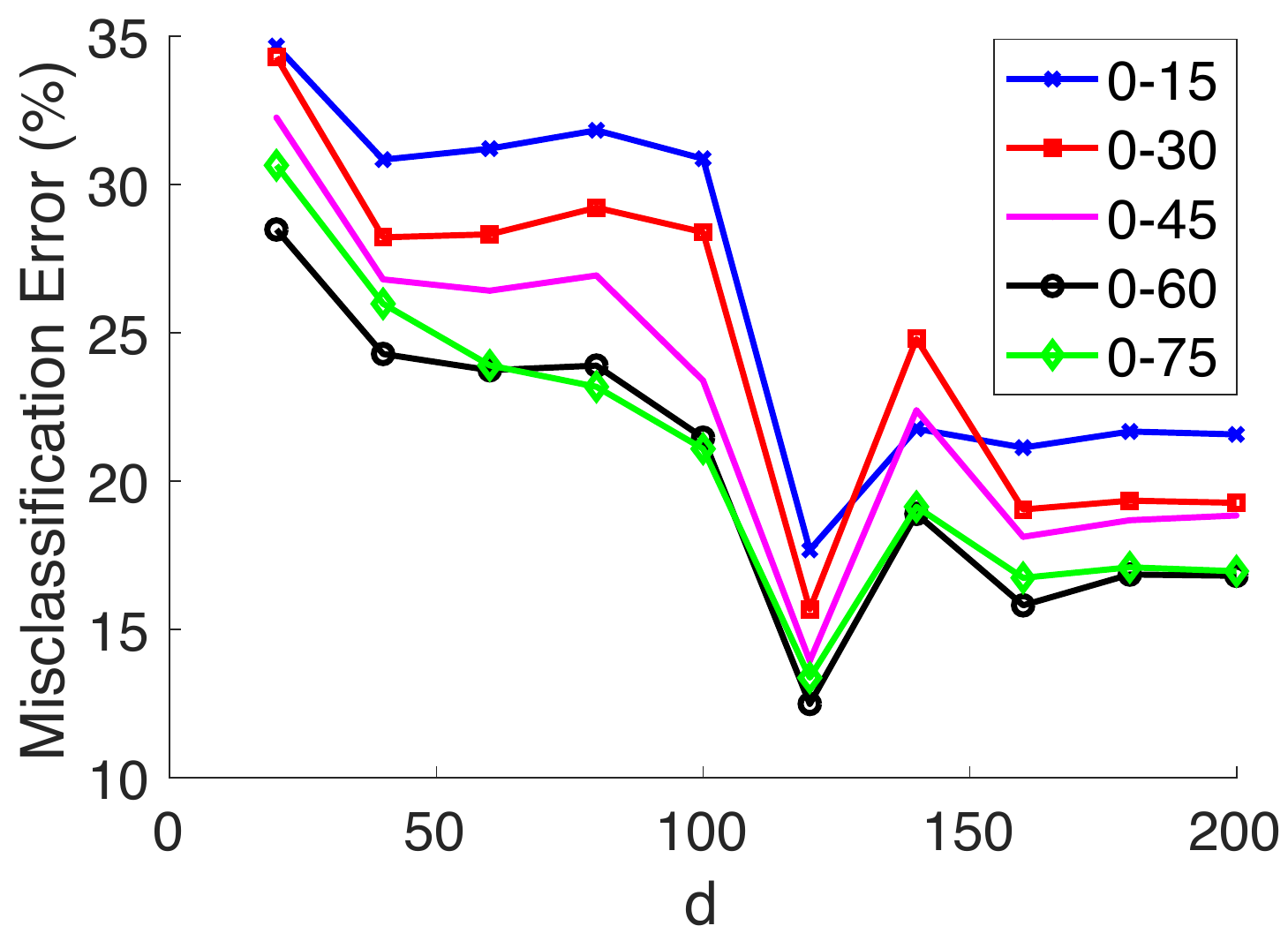}}
  \centerline{(b) MultiPIE}\medskip
\end{minipage}
\hfill
\begin{minipage}[b]{0.49\linewidth}
  \centering
  \centerline{\includegraphics[width=4.3cm]{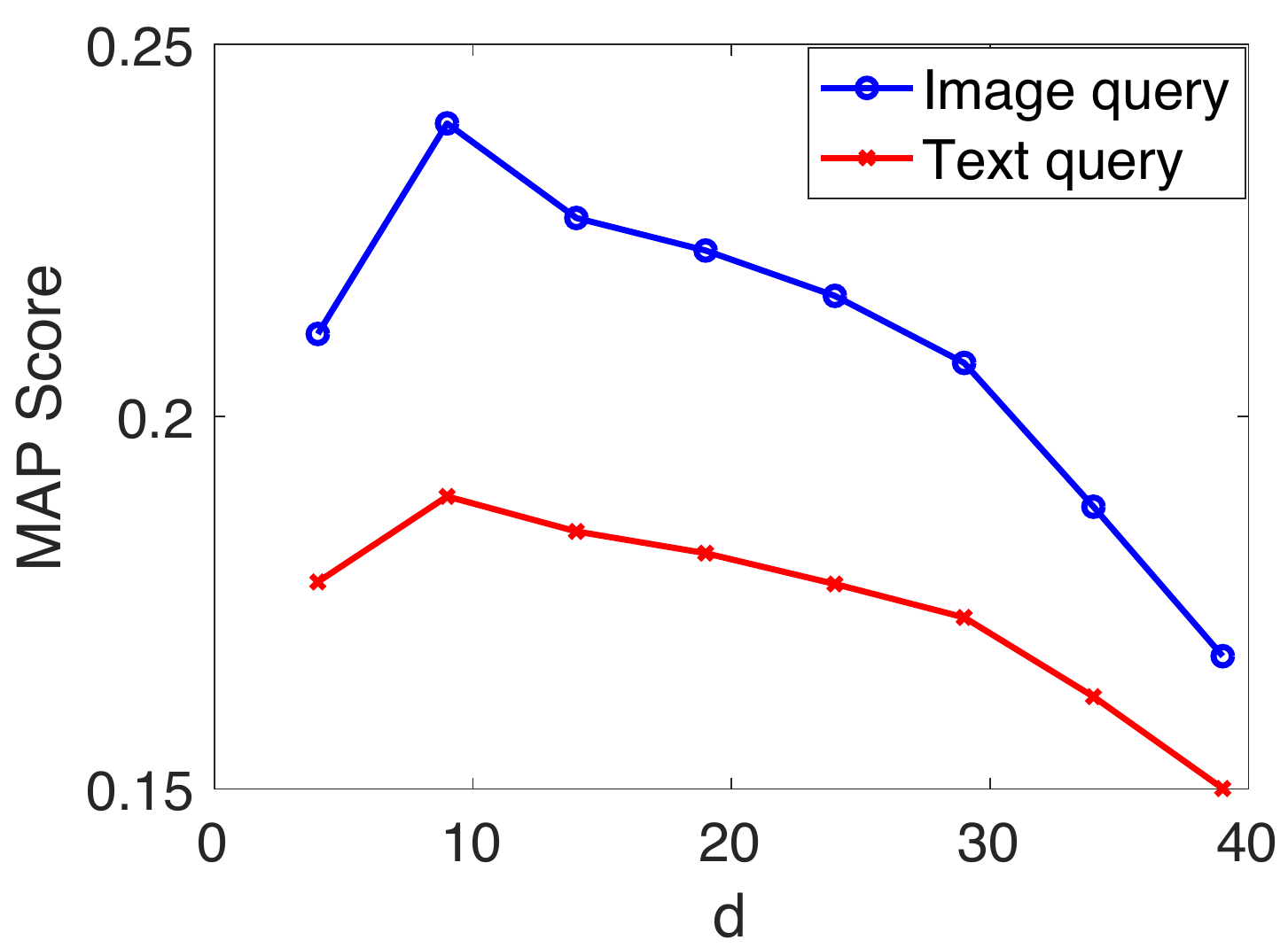}}
  \centerline{(c) Wikipedia}\medskip
\end{minipage}
\hfill
\begin{minipage}[b]{0.49\linewidth}
  \centering
  \centerline{\includegraphics[width=4.3cm]{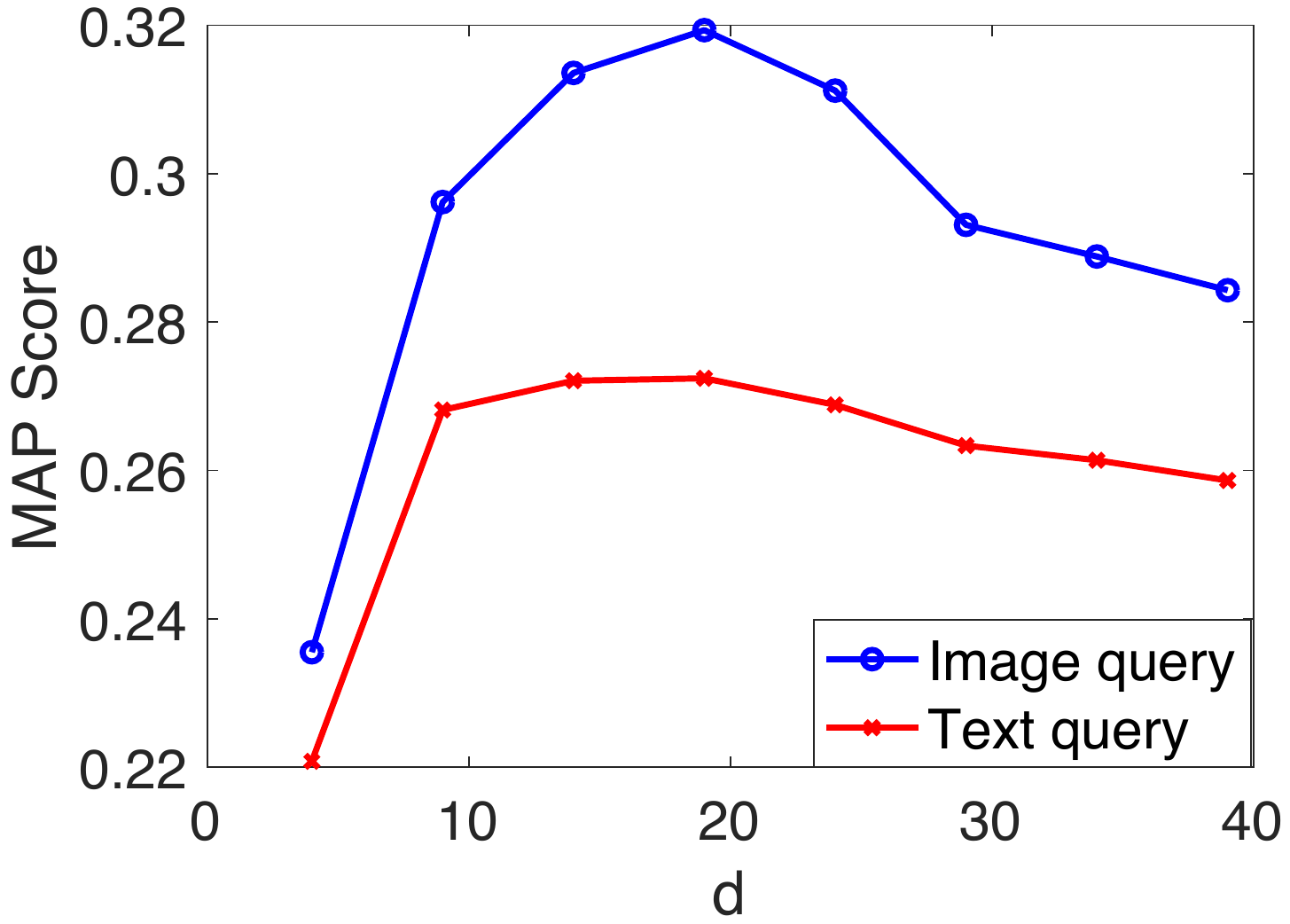}}
  \centerline{(d) Pascal VOC2007}\medskip
\end{minipage}
\caption{Variation of algorithm performance with embedding dimension $\dnew$}
\label{figEmbeddingDimensionAndKernelScale}
\end{figure}

Finally, we examine the effect of the embedding dimension $\dnew$ on the algorithm performance in Figure \ref{figEmbeddingDimensionAndKernelScale}. Different curves correspond to the misclassification errors obtained at different training sizes (ratio of training samples) for MIT-CBCL, and at different camera angles for Modality 2 for the Multi-PIE data set. The MAP scores of the image and text queries are reported individually for the Wikipedia and the Pascal VOC2007 data sets. The optimal embedding dimension is consistently seen to be close to the number of classes in all four data sets. In the rest of our experiments, the embedding dimension $\dnew$ is chosen as $M-1$ for each data set, where $M$ is the number of classes.

%
%
%From Figure \ref{figEmbeddingDimensionAndKernelScale}, it can be understood that the MNSE algorithm works well at lower dimensions for the MIT CBCL face data set. The smallest dimension value that yields a reasonable misclassification error in Figure \ref{figEmbeddingDimensionAndKernelScale} can be observed as $9$. Achieving high classification accuracy at low dimensions is also helpful as it decreases the computational load. For these reasons, the embedding dimension is chosen as $9$ during the experiments.

%
%For the MIT CBCL face images data set, there exist $10$ classes and the optimum embedding dimension is obtained as $9$. This suggests that the optimum embedding dimension may be expected to be close to the number of classes for the data set.

%
%It can be understood from Figure \ref{figEmbeddingDimensionAndKernelScale} that a suitable value for the kernel scale parameter $\sigma$ can be found through a basic search in an interval. This is the result from the fact that any increase in $\sigma$ decreases the cost of the kernel scale parameter in the overall objective, but it increases the cost of the kernel function norm and vice versa.

%
%
%

\subsection{Evaluation of the Algorithm Performance}
\label{ssec:eval_alg}

\par We now evaluate the performance of the proposed MNSE algorithm with comparative experiments in image classification and image-text retrieval applications. MNSE is compared to the multi-modal representation learning algorithms CCA, Kernel CCA \cite{surveyOnMl},  GMLDA \cite{gma}, Kernel GMLDA \cite{gma}, KMvMDA \cite{CaoICG18}, JFSSL \cite{jfssl}, and DeepMF \cite{ZhaoDF17}, as well as the baseline single-modal methods PCA, NN classification in the original domain, and NSSE \cite{nsse}. We use our own implementations for the GMLDA, Kernel GMLDA, KMvMDA, and JFSSL methods in the experiments. The CCA and GMLDA algorithms are applied after a preprocessing step of dimensionality reduction with PCA, which has been seen to improve their performance. The single-modal methods are applied independently in each modality. The parameters of the compared algorithms are optimized for the best performance on the test set. The parameter settings used for the algorithms that require parameter tuning are given in Table \ref{tab:alg_param} for all data sets (intervals mean that parameters may vary in different repetitions of the experiments.) %\textbf{In the classification experiments, the weight parameters of the proposed MNSE algorithm are selected within the regions suggested in Section \ref{ssec:sensit_analy}.} For the multi-modal methods, nonlinear embeddings of different modalities into a common domain are learnt with the training data, which are then used for embedding the test data. 

\begin{table}[t]
\scriptsize
\centering
\caption{Algorithm parameter values used in the experiments (denoted as in the papers cited in the table)}
\label{tab:alg_param}
\begin{tabular}{|c|c|c|c|c|c|c|c|}
\hline
Algorithm& Parameter & MIT-CBCL & Multi-PIE & Pascal VOC & Wikipedia \\ \hline
\textbf{PCA}  & $d$  (dim.) &   15 & 82  & - & -  \\ \hline
 &  $d$ & 8-20 & 119 & - & -  \\
 \textbf{NSSE} \cite{nsse}& $\mu_1$  & 42-400 & 400 & - & -  \\ 
 & $\mu_2$  & $10^{-4}-0.1$ & 0.1& - & -   \\
 & $\mu_3$  & 1-3 & 1.2 & - & -   \\ \hline
 \textbf{CCA} \cite{surveyOnMl}& $d$ (dim.)  & 15 & 80 & 7 & 8  \\ \hline
&  $d$ (dim.) & 15 & 220 & 19 & 8  \\
  \textbf{Kernel CCA} \cite{vanVaerenbergh2010} &  $c'$ & $10^{-5}$ & 0.1 & 100 & 1000  \\
 & Kernel type  & Gauss. & Gauss. & Gauss. & Gauss.  \\ \hline
 %& $\mu$  & X &X & X & X   \\ \hline
 &  $k$ (dim.)  & 15 & 124 & 9 & 9  \\
 \textbf{GMLDA} \cite{gma}& $\alpha$  & 10 & 10 & 500 & 500  \\ 
 & $\mu$  & 1 & 2 & 0.05 & 0.01   \\ \hline
  &  $k$ (dim.) & 15-18 & 124 & 9 & 9  \\
 \textbf{Kernel GMLDA} \cite{gma}& $\alpha$  & 10-100 & 100 & 500 & 500  \\ 
 & $\mu$  & 0.8-1.5 & 2 & 0.05 & 0.01   \\
 & Kernel type  & Chi sq. & Chi sq. & Chi sq. & Chi sq.  \\ \hline
  &  $k$ & 30 & 40 & 300 & 130  \\
& $\lambda_1$  & 0.01 & 0.1 & 0.01 & 0.01  \\ 
  \textbf{JFSSL} \cite{jfssl}& $\lambda_2$  & 0.01 &0.01 & 0.001 & 0.001   \\ 
  & $\beta$  & 1 & 1 & 1 & 1   \\ 
   & $\epsilon$  & $10^{-8}$ & $10^{-8}$ & $10^{-8}$ & $10^{-8}$   \\ \hline
  &  $\gamma$ & 0.95 & 0.95 & 0.5 & 0.5  \\
   \textbf{DeepMF} \cite{ZhaoDF17} &  $\beta$ & 0.01 & 0.01 & 0.01 & 0.01  \\
 & $[p_1, p_2]$  & [250, 45] & [250, 45]  & [100, 50]  & [100, 10]   \\ 
 & $k$  & 5 & 5 & 5 & 5   \\ \hline
   \textbf{KMvMDA} \cite{CaoICG18} &  $d$ & 38-120 & 119 & 19 & 9  \\
& Kernel type  & Gauss. & Gauss. & Gauss. & Gauss.  \\ \hline
  &  $d$ & 9 & 119 & 19 & 9  \\
 &  $\mu_1=\mu_5$ & 100 & 100 & 0.1 & 0.001-0.1  \\
 \textbf{MNSE} & $\mu_2$  & 0.001 & 0.001 & 1 & 0.1-1  \\ 
  & $\mu_3$  & 1 & 1 & 1 & 0.001-10   \\
 & $\mu_4$  & 100 &100 & 10 & 0.001-0.1   \\ \hline
\end{tabular}
\vspace{-0.5cm}
\end{table}

\subsubsection{Multi-modal image classification} The multi-modal classification experiments are done on the MIT-CBCL and the Multi-PIE face data sets. The weight parameters of the proposed MNSE method are chosen within the regions suggested in Section \ref{ssec:sensit_analy} as $\mu_1=\mu_4=\mu_5=10^2$, $\mu_2=10^{-3}$, $\mu_3=1$ for both data sets. Tables \ref{tab:errors_mitcbcl} and \ref{tab:errors_multipie} show the test misclassification errors in percentage for the MIT-CBCL and the Multi-PIE data sets. The upper and lower tables show the average errors and the error standand deviations over the 10 random trials for MIT-CBCL, and over the 4 experimentation settings considered for the Multi-PIE data set. The error is studied with respect to the training size (ratio of the training samples) for MIT-CBCL, and the camera angle of Modality 2 for Multi-PIE. The errors obtained for Modalities 1 and 2 are given in the top and the bottom rows for each method.

%
%%%%%% TABLE: ERRORS MITCBCL
\begin{table}[t]
	\caption{Misclassification errors (\%) for the MIT-CBCL data set: Average errors (Upper table) and error standard deviations (Lower table) over 10 random trials.}
\scriptsize
\centering
\begin{tabular}
{|c| l | l | l | l | l |} 
	\hline
    \multirow{2}{*}{\textbf{Avg. errors}} & \multicolumn{5}{c|}{\textbf{Training size}}  \\
    \cline{2-6}
     & \textbf{5.6\%} &  \textbf{8.3\%} &  \textbf{11.1\%} & \textbf{13.9\%} & \textbf{27.8\%}  \\ 
    \hline
    \multirow{2}{*}{NN}  	& 22.12 & 19 & 10.69 & 2.97 & 0.77  \\
        					& 19.68 & 17.64 & 6.94 & 1.71 &  \textbf{0}   \\
    \hline 
    \multirow{2}{*}{PCA}  	& 3.68 & 0.06 & 0.34 & 0.10 &  \textbf{0} \\
      						& 4.29  & 0.54 & 0.06  &  \textbf{0}  &  \textbf{0}   \\
    \hline  		
    \multirow{2}{*}{NSSE \cite{nsse}} 	& 1.94  & 0.03 & 0.03  &  \textbf{0}  &  \textbf{0} \\
      					  	            & 4.56 & 1.00 & 0.09 & 0.03 &  \textbf{0} \\ 
    \hline
    \multirow{2}{*}{CCA \cite{surveyOnMl}}  	& 3.67  & 0.06 & 0.34  & 0.10 &  \textbf{0}  \\
      						                    & 4.29  & 0.55 & 0.06  &  \textbf{0} &  \textbf{0}    \\ 
    
    \hline    
    \multirow{2}{*}{Kernel CCA \cite{surveyOnMl}} & 1.50  & 0.21 & 0.21  & 0.09 &  \textbf{0} \\
          				     			                       & 4.00    & 0.72 & 0.66  & 0.39 & 0.19   \\
	\hline    
    \multirow{2}{*}{GMLDA \cite{gma}} 	&  2.56 &  \textbf{0} &  \textbf{0} & \textbf{0} & \textbf{0} \\
    					            &  5.82 & 0.30  & 0.06  &  0.03 &  \textbf{0}   \\ 
         	
     \hline    
    \multirow{2}{*}{Kernel GMLDA \cite{gma}} 	&  7.68 &   0.70 &    1.81 &   0.39 &   \textbf{0} \\
         					            &    3.15 &   2.09  &   0.50 &    0.19 &    \textbf{0}   \\ 

	\hline    
    \multirow{2}{*}{JFSSL \cite{jfssl}} 	&   \textbf{0}&  \textbf{0}&  \textbf{0} & \textbf{0} & \textbf{0} \\
         					                & \textbf{0.12}  &  \textbf{0}  &  \textbf{0}   &  \textbf{0} &  \textbf{0}  \\
         					                	\hline    
    \multirow{2}{*}{DeepMF \cite{ZhaoDF17}} 	& 7.12 & 1.58 & 1.00 & 0.65 &  \textbf{0} \\
         					                & 5.97  & 1.36  & 0.63  & 0.45 &  \textbf{0}   \\

     \hline    
    \multirow{2}{*}{KMvMDA  \cite{CaoICG18}} 	&  11.91 &   0.48 &    1.72 &   0.84 &   \textbf{0} \\
         					            &    17.50 &   0.70  &   1.22 &    0.58 &    \textbf{0}   \\ 
					                
	\hline    
    \multirow{2}{*}{MNSE} 	& 0.15  &  \textbf{0} &  \textbf{0}  &  \textbf{0} &  \textbf{0} \\
          					& 1.35  & 0.27 & 0.03  &  \textbf{0} & \textbf{0}    \\ 
    \hline 
\end{tabular}
%\label{tab:errors_mitcbcl}
%\end{table}
%\normalsize
%
%

%
%
%%%%%% TABLE: STANDARD DEVIATIONS MITCBCL
%\begin{table}[t]
%	\caption{Error standard deviations for the MIT-CBCL data set.}
%\scriptsize
%\centering
\begin{tabular}
{|c| l | l | l | l | l |} 
	\hline
    \multirow{2}{*}{\textbf{Error st. dev.}} & \multicolumn{5}{c|}{\textbf{Training size}}  \\
    \cline{2-6}
     & \textbf{5.6\%} &  \textbf{8.3\%} &  \textbf{11.1\%} & \textbf{13.9\%} & \textbf{27.8\%}  \\ 
    \hline
    \multirow{2}{*}{NN}  	& 21.13 & 23.25 & 9.97& 2.32 & 0.65  \\
        					& 18.39 & 21.81 & 7.71& 2.73 & \textbf{0}   \\
    \hline 
    \multirow{2}{*}{PCA}  	& 4.46 & 0.13 & 0.88& 0.31 & \textbf{0} \\
      						& 9.55 & 1.07 & 0.20& \textbf{0} & \textbf{0}  \\
    \hline  		
    \multirow{2}{*}{NSSE \cite{nsse}} 	& 3.14 & 0.10 & 0.10& \textbf{0} & \textbf{0} \\
      					  	            & 3.98 & 2.38 & 0.30& 0.10 & \textbf{0} \\ 
    \hline
    \multirow{2}{*}{CCA \cite{surveyOnMl}}  	& 4.46 & 0.13 & 0.88& 0.31 & \textbf{0}  \\
      						                   & 9.55 & 1.08 & 0.20& \textbf{0} & \textbf{0}  \\ 
    
    \hline    
    \multirow{2}{*}{Kernel CCA \cite{surveyOnMl}} & 2.88 & 0.67 & 0.69& 0.31 & \textbf{0} \\
          				     			                       & 5.40 & 0.98 & 0.76& 0.60 & 0.49  \\
	\hline    
    \multirow{2}{*}{GMLDA \cite{gma}} 	& 2.16 & \textbf{0} & \textbf{0}& \textbf{0} & \textbf{0}\\
    					            & 3.76 & 0.96 & 0.20& 0.10 & \textbf{0}   \\ 
         	
     \hline    
    \multirow{2}{*}{Kernel GMLDA \cite{gma}} 	& 7.60 & 1.51 & 2.86& 1.02 & \textbf{0}\\
         					            & 4.71 & 4.32 & 1.47& 0.61 & \textbf{0}   \\ 

	\hline    
    \multirow{2}{*}{JFSSL \cite{jfssl}} 	& \textbf{0} & \textbf{0} & \textbf{0}& \textbf{0} & \textbf{0} \\
         					               & \textbf{0.28} & \textbf{0} & \textbf{0}& \textbf{0} & \textbf{0} \\
         					                	\hline    
    \multirow{2}{*}{DeepMF \cite{ZhaoDF17}} 	& 5.88 & 1.70 & 0.75& 1.05 & \textbf{0}\\
         					             & 4.93 & 1.53 & 0.61& 0.88 & \textbf{0}  \\

     \hline    
    \multirow{2}{*}{KMvMDA  \cite{CaoICG18}} 	& 10.52 & 1.13 & 2.97& 1.65 & \textbf{0}\\
         					          & 10.02 & 0.89 & 1.92& 1.02 & \textbf{0}   \\ 
					                
	\hline    
    \multirow{2}{*}{MNSE} & 0.25 & \textbf{0} & \textbf{0}& \textbf{0} & \textbf{0} \\
          					& 1.55 & 0.76 & 0.10& \textbf{0} & \textbf{0}  \\ 
    \hline 
\end{tabular}
\label{tab:errors_mitcbcl}
\end{table}
\normalsize

%%%

%%%%%% TABLE: MULTI-PIE
\begin{table}[t]
	\caption{Misclassification errors (\%) for the Multi-PIE data set: Average errors (Upper table) and error standard deviations (Lower table) over 4 experimentation settings}
\scriptsize
\centering
\begin{tabular}
{|c| l | l | l | l | l |} 
	\hline
    \multirow{2}{*}{\textbf{Avg. errors}} & \multicolumn{5}{c|}{\textbf{Camera angle for Modality 2}}  \\
    \cline{2-6}
     & \textbf{15\degree} &  \textbf{30\degree} &  \textbf{45\degree} & \textbf{60\degree} & \textbf{75\degree}  \\ 
    \hline
    \multirow{2}{*}{NN}  	& 20.28 & 20.28 & 20.28 & 20.28 & 20.28  \\
        					& 21.40 & 18.93 & 18.21 & 12.27 & 17.01   \\
    \hline 
    \multirow{2}{*}{PCA}  	& 22.51 & 22.51 & 22.51 & 22.51 & 22.51 \\
      						& 24.59 & 22.06 & 20.03 & 15.81 & 17.23   \\
    \hline  		
    \multirow{2}{*}{NSSE \cite{nsse}} 	& 17.46 & 17.46 & \textbf{17.46} &  \textbf{17.46} &  \textbf{17.46} \\
      					  	           & 19.85 & 18.12 & 16.12 & 10.78 & 15.51 \\ 
    \hline
    \multirow{2}{*}{CCA \cite{surveyOnMl}}  	& 22.80 & 22.80 & 22.80 & 22.80 & 22.80 \\
      						                    & 24.47 & 22.03 & 20.06 & 15.88 & 17.15  \\ 
    
    \hline    
    \multirow{2}{*}{Kernel CCA \cite{surveyOnMl}} & 20.88 & 21.22 & 21.08 & 21.67 & 21.08 \\
          				     			                       & 23.31 & 24.25 & 29.01 & 21.53 & 29.18  \\
	\hline    
    \multirow{2}{*}{GMLDA \cite{gma}} 	& 26.92 & 26.18 & 25.74 & 25.35 & 25.26 \\
    					           & 28.73 & 26.08 & 21.54 & 16.71 & 18.42  \\ 
         	
     \hline    
    \multirow{2}{*}{Kernel GMLDA \cite{gma}} 	& 45.42 & 45.43 & 49.87 & 53.42 & 56.36 \\
         					               & 42.99 & 37.63 & 29.72 & 34.74 & 36.73   \\

	\hline    
    \multirow{2}{*}{JFSSL \cite{jfssl}} 	& 24.60 & 24.61 & 24.62 & 24.61 & 24.62 \\
         					           & 30.77 & 25.30 & 18.93 & 16.12 & 19.99   \\ 
					           
\hline    
    \multirow{2}{*}{DeepMF \cite{ZhaoDF17}} 	& 29.28 & 26.67 & 26.31 & 26.01 & 26.06 \\
         					               & 33.15 & 32.24 & 32.73 & 25.48 & 29.48   \\

     \hline    
    \multirow{2}{*}{KMvMDA  \cite{CaoICG18}} 	&  46.88 &   48.67 &    44.58 &   39.40 &   46.20 \\
         					            &    50.97 &   52.74  &   47.16 &    37.68 &    50.92   \\ 
					                
	\hline    
    \multirow{2}{*}{MNSE} 	& \textbf{17.10} & \textbf{17.11} & 18.51 & 18.22 & 17.60 \\
          					&  \textbf{18.85} &  \textbf{14.24} & \textbf{11.26} & \textbf{ 9.85} &  \textbf{10.44}   \\ 
    \hline 
\end{tabular}
%\label{tab:errors_multipie}
%\end{table}
%\normalsize
%%%%

%%%%%% TABLE: MULTI-PIE
%\begin{table}[t]
%	\caption{Error standard deviations for the Multi-PIE data set.}
%\scriptsize
\centering
\begin{tabular}
{|c| l | l | l | l | l |} 
	\hline
    \multirow{2}{*}{\textbf{Error st. dev.}} & \multicolumn{5}{c|}{\textbf{Camera angle for Modality 2}}  \\
    \cline{2-6}
     & \textbf{15\degree} &  \textbf{30\degree} &  \textbf{45\degree} & \textbf{60\degree} & \textbf{75\degree}  \\ 
    \hline
    \multirow{2}{*}{NN}  	& 1.35 & 1.35 & 1.35 & 1.35 & 1.35  \\
        					& 1.41 & 1.80 & 2.16 &  \textbf{0.28} & 1.73   \\
    \hline 
    \multirow{2}{*}{PCA}  	& 1.03 & 1.03 & 1.03 & 1.03 & 1.03 \\
      						& 1.51 & 2.37 & 1.50 & 1.99 & 3.94   \\
    \hline  		
    \multirow{2}{*}{NSSE \cite{nsse}} 	&  \textbf{0.21} & 0.21 &  \textbf{0.21} &  \textbf{0.21} &  \textbf{0.21} \\
      					  	           & 2.09 & 1.89 & 1.30 & 0.72 & 1.84 \\ 
    \hline
    \multirow{2}{*}{CCA \cite{surveyOnMl}}  	& 1.15 & 1.15 & 1.15 & 1.15 & 1.15 \\
      						                    & 1.63 & 2.16 & 1.55 & 2.00 & 3.86  \\ 
    
    \hline    
    \multirow{2}{*}{Kernel CCA \cite{surveyOnMl}} & 1.86 & 1.22 & 1.67 & 1.24 & 1.66 \\
          				     			                       & 1.79 &  \textbf{0.75} & 3.79 & 0.98 & 3.39  \\
	\hline    
    \multirow{2}{*}{GMLDA \cite{gma}} 	& 2.11 & 2.14 & 1.62 & 2.19 & 1.66 \\
    					           & 3.74 & 4.44 & 3.47 & 3.96 & 5.34  \\ 
         	
     \hline    
    \multirow{2}{*}{Kernel GMLDA \cite{gma}} 	& 16.33 & 12.87 & 13.54 & 22.92 & 14.99 \\
         					           & 13.37 & 11.74 & 8.84 & 19.36 & 19.74   \\ 

	\hline    
    \multirow{2}{*}{JFSSL \cite{jfssl}} 	& 0.69 & 0.67 & 0.67 & 0.69 & 0.67 \\
         					               & 1.35 & 1.35 & 1.73 & 1.29 &  \textbf{1.23}  \\
       	\hline    
    \multirow{2}{*}{DeepMF \cite{ZhaoDF17}} 	& 1.93 & 1.80 & 1.81 & 1.82 & 1.77 \\
         					               & 0.64 & 1.58 &  \textbf{0.54} & 1.96 & 1.85   \\

     \hline    
    \multirow{2}{*}{KMvMDA  \cite{CaoICG18}} 	&  19.53 &   6.69 &    16.68 &   7.07 &   16.99 \\
         					            &    18.33 &   2.05  &   19.06 &    3.20 &    19.33   \\ 
					                
	\hline    
    \multirow{2}{*}{MNSE} 	& 0.25 &  \textbf{0.18} & 0.72 & 1.49 & 0.76 \\
          					&  \textbf{0.40} & 1.10 & 1.80 & 2.64 & 1.76   \\ 
    \hline 
\end{tabular}
%\label{tab:stdev_multipie}
\label{tab:errors_multipie}
\end{table}
\normalsize
%%%%

%
The results in Tables \ref{tab:errors_mitcbcl} and  \ref{tab:errors_multipie}  show that the proposed MNSE method outperforms all methods except for NSSE and JFSSL in all experiments. On the MIT-CBCL data set, the linear supervised JFSSL algorithm performs the best. The approach of aligning the modalities via linear projections in JFSSL is particularly suited to this synthetic and regularly structured face data set. The proposed MNSE method  closely follows JFSSL on MIT-CBCL with very small misclassification rates. On the other hand, MNSE outperforms JFSSL and the other multi-modal algorithms on the Multi-PIE data set.  The nonlinear MNSE method learns a relatively rich model while explicitly incorporating generalization performance in its objective. These features bring our method an advantage on the Multi-PIE data set, which has a more challenging structure than MIT-CBCL due to the large number of classes and facial expression and illumination variations. MNSE is seen to perform better than NSSE in most experiments, which also performs very well on this data set. NSSE computes nonlinear and smooth embeddings as MNSE;  however, in a single modality. The fact that MNSE often outperforms NSSE confirms that it can successfully exploit and combine the information from both modalities when optimizing the embedding parameters. The standard deviations of the errors suggest that the algorithms with small average errors maintain rather stable performance figures over different repetitions of the experiments for both data sets.

\subsubsection{Cross-modal image-text retrieval} 
\begin{figure}[t]
%
% \vspace{-0.5cm}
\begin{minipage}[b]{0.49\linewidth}
  \centering
  \centerline{\includegraphics[width=4.3cm]{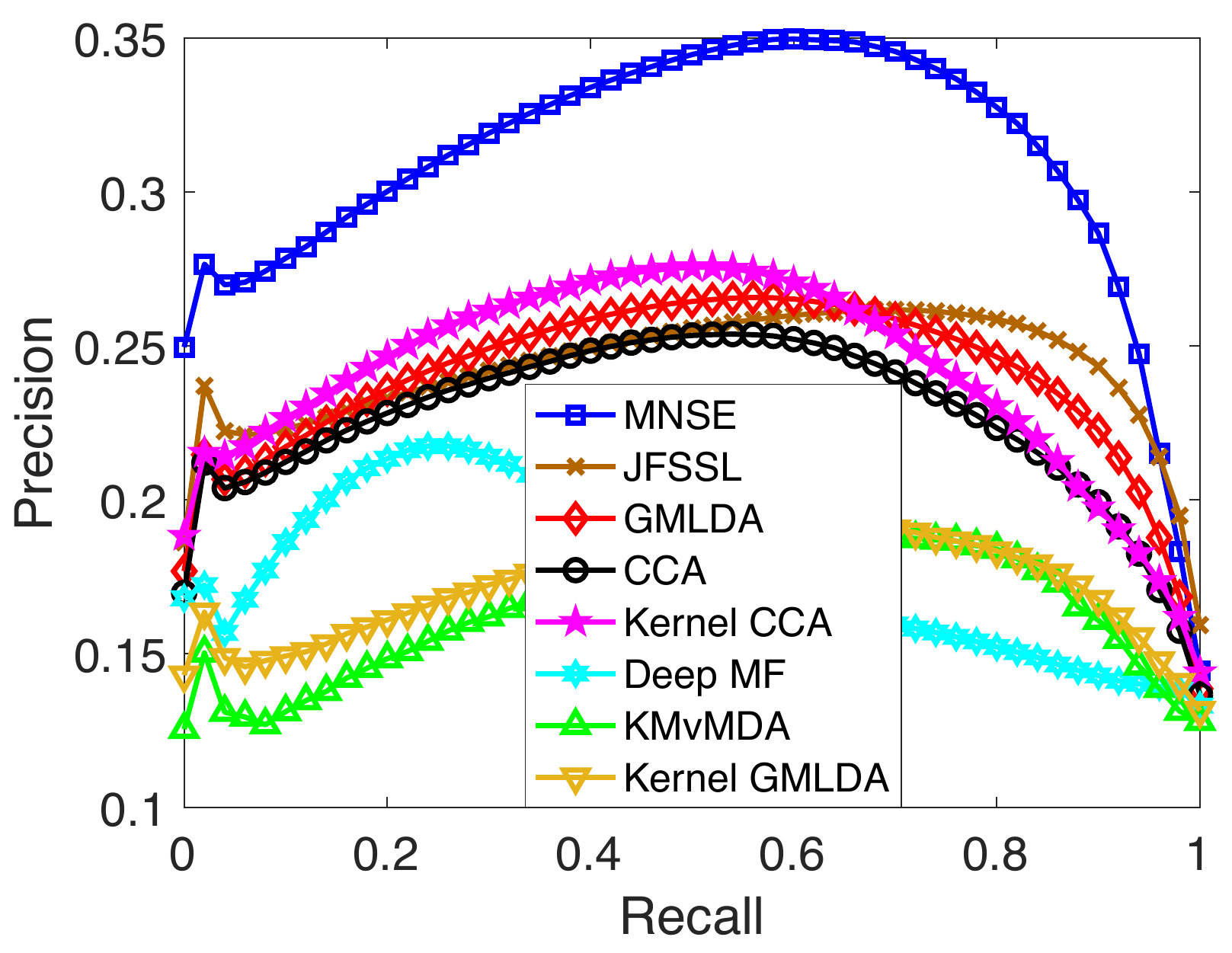}}
  \centerline{(a) Image query}\medskip
\end{minipage}
\hfill
\begin{minipage}[b]{0.49\linewidth}
  \centering
  \centerline{\includegraphics[width=4.3cm]{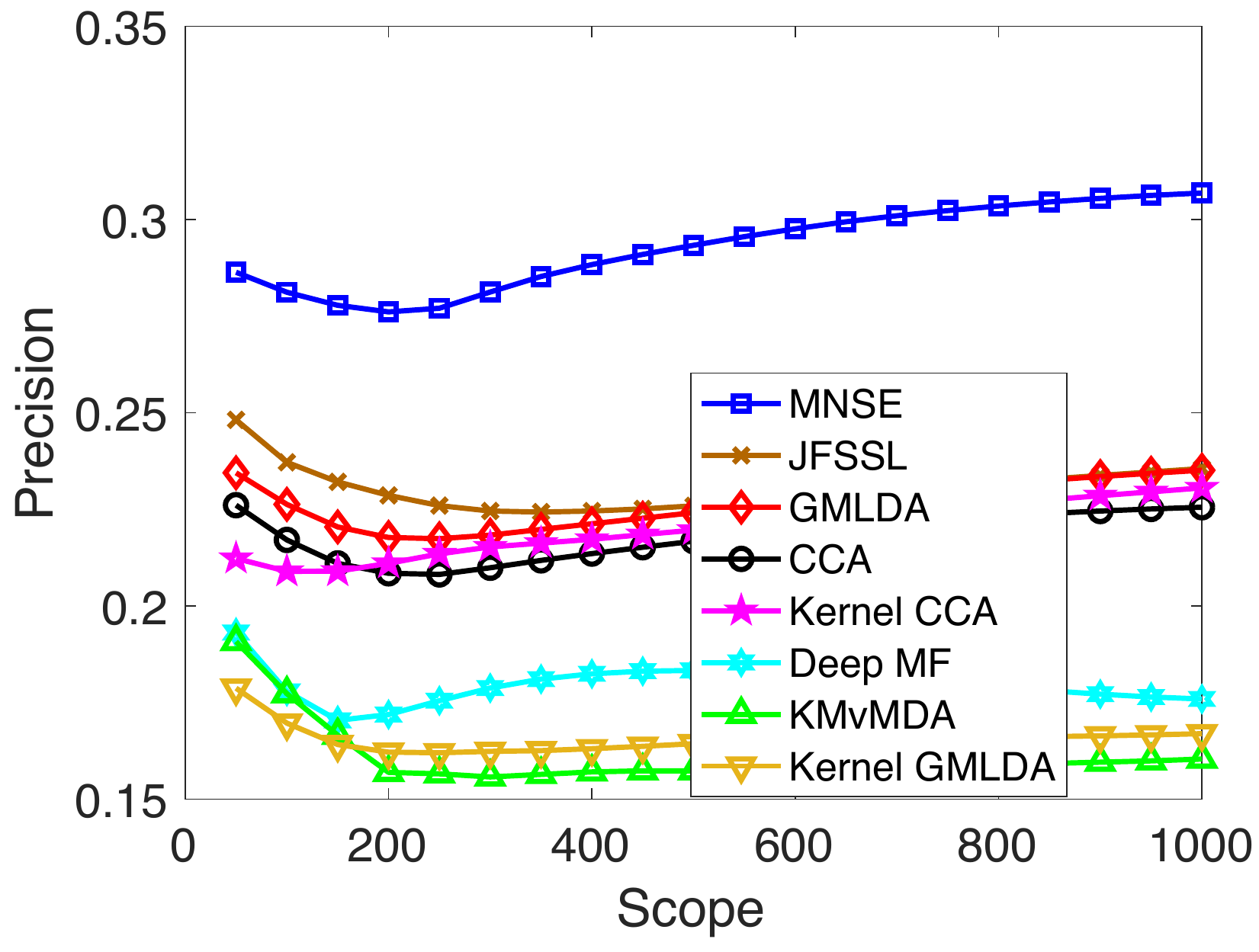}}
  \centerline{(b) Image query}\medskip
\end{minipage}
\begin{minipage}[b]{0.49\linewidth}
  \centering
  \centerline{\includegraphics[width=4.3cm]{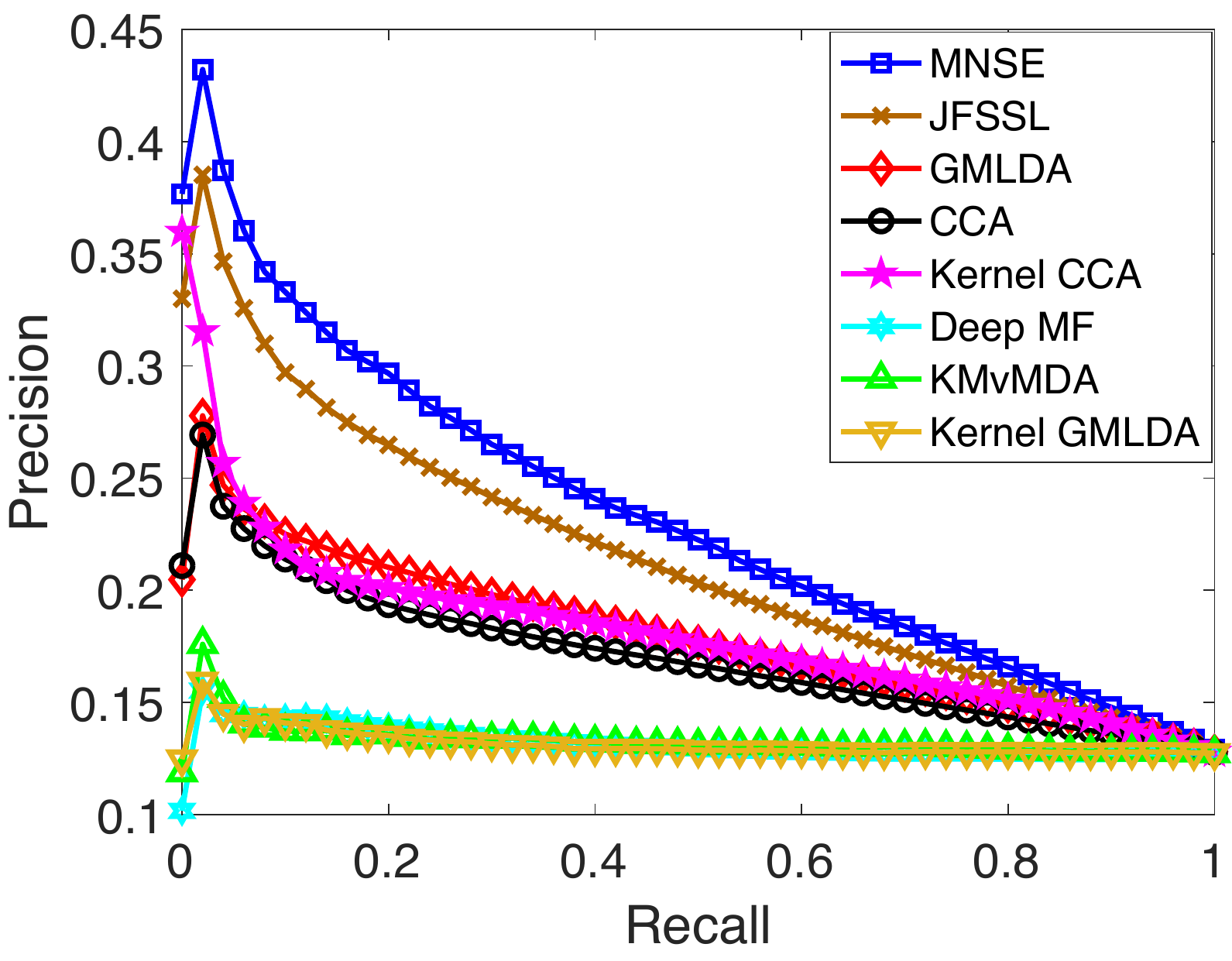}}
  \centerline{(c) Text query}\medskip
\end{minipage}
\hfill
\begin{minipage}[b]{0.49\linewidth}
  \centering
  \centerline{\includegraphics[width=4.3cm]{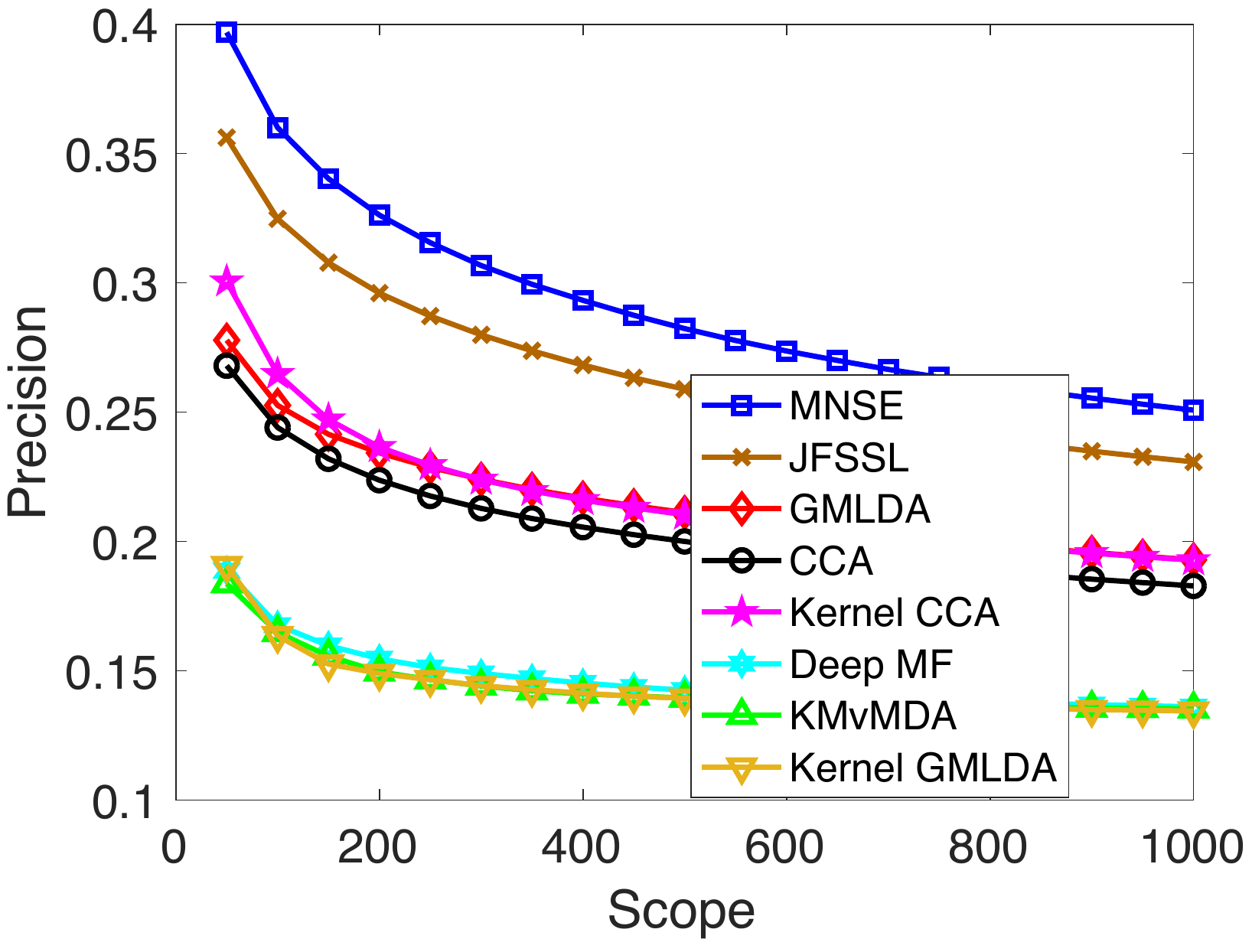}}
  \centerline{(d) Text query}\medskip
\end{minipage}
\vspace{-0.7cm}
\caption{Retrieval performance of the methods for Wikipedia}
\label{figWikiRetrieval}
%
% \vspace{-0.3cm}
\end{figure}

\begin{figure}[t]
%
% \vspace{-0.5cm}
\begin{minipage}[b]{0.49\linewidth}
  \centering
  \centerline{\includegraphics[width=4.3cm]{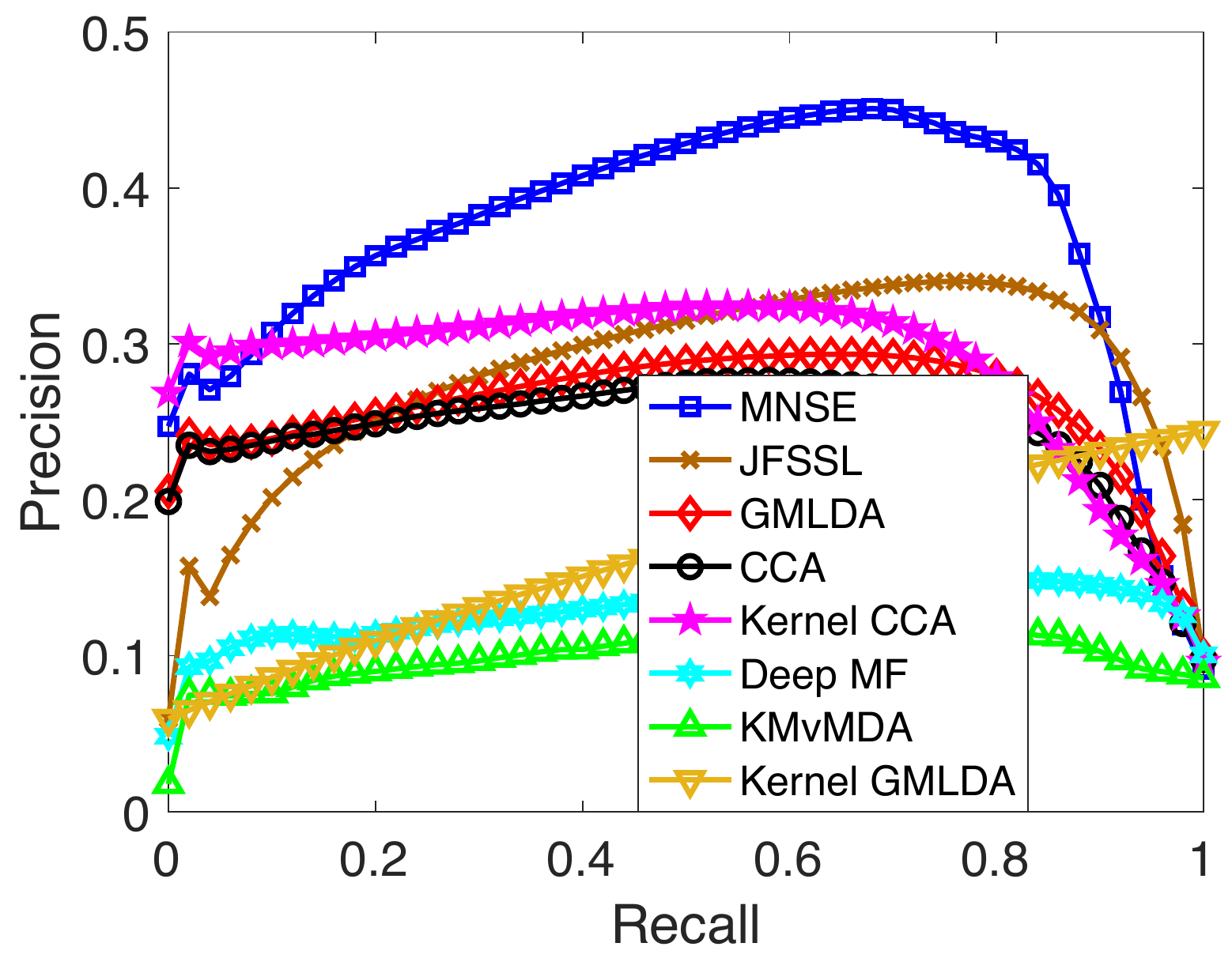}}
  \centerline{(a) Image query}\medskip
\end{minipage}
\hfill
\begin{minipage}[b]{0.49\linewidth}
  \centering
  \centerline{\includegraphics[width=4.3cm]{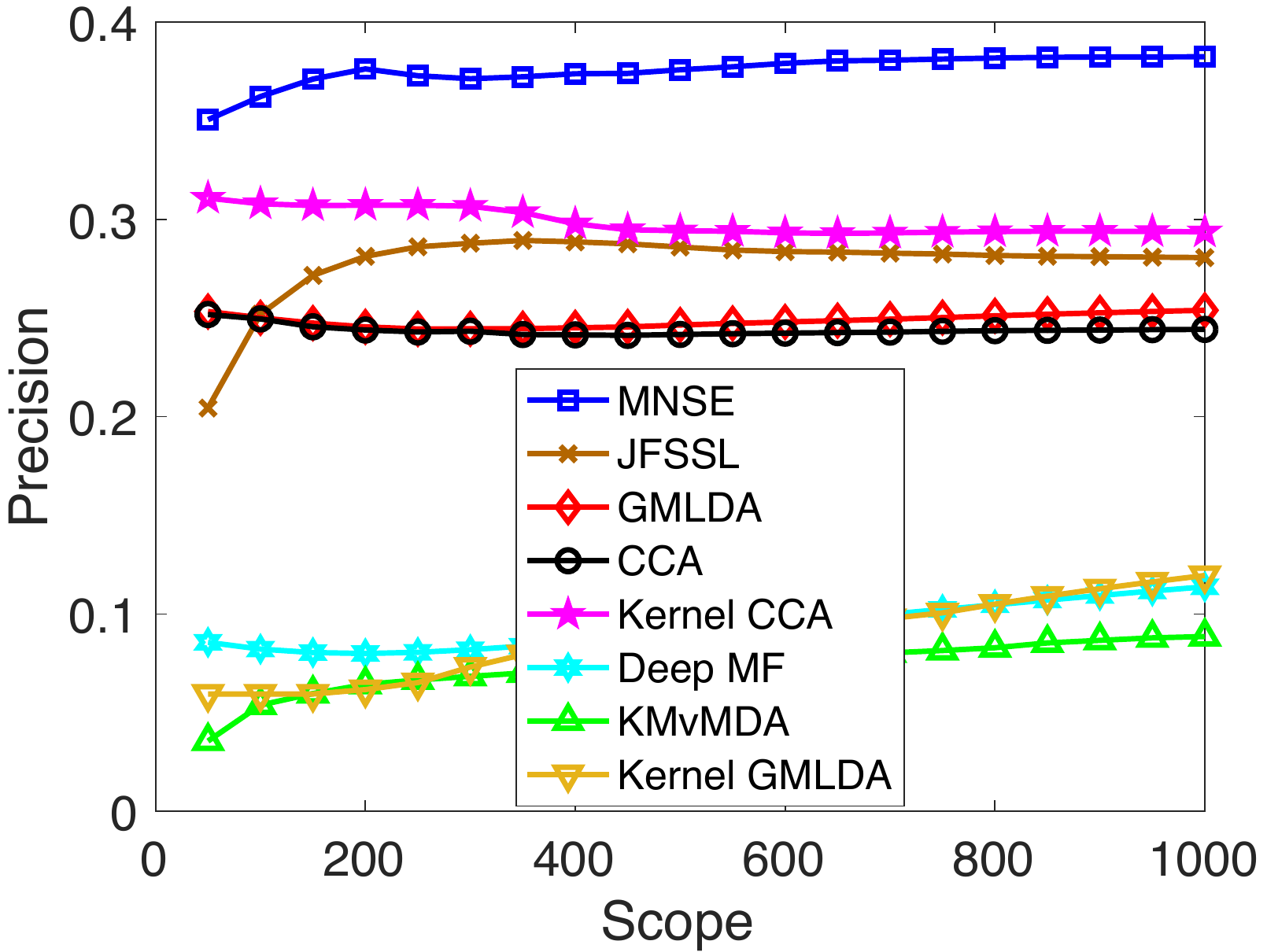}}
  \centerline{(b) Image query}\medskip
\end{minipage}
\begin{minipage}[b]{0.49\linewidth}
  \centering
  \centerline{\includegraphics[width=4.3cm]{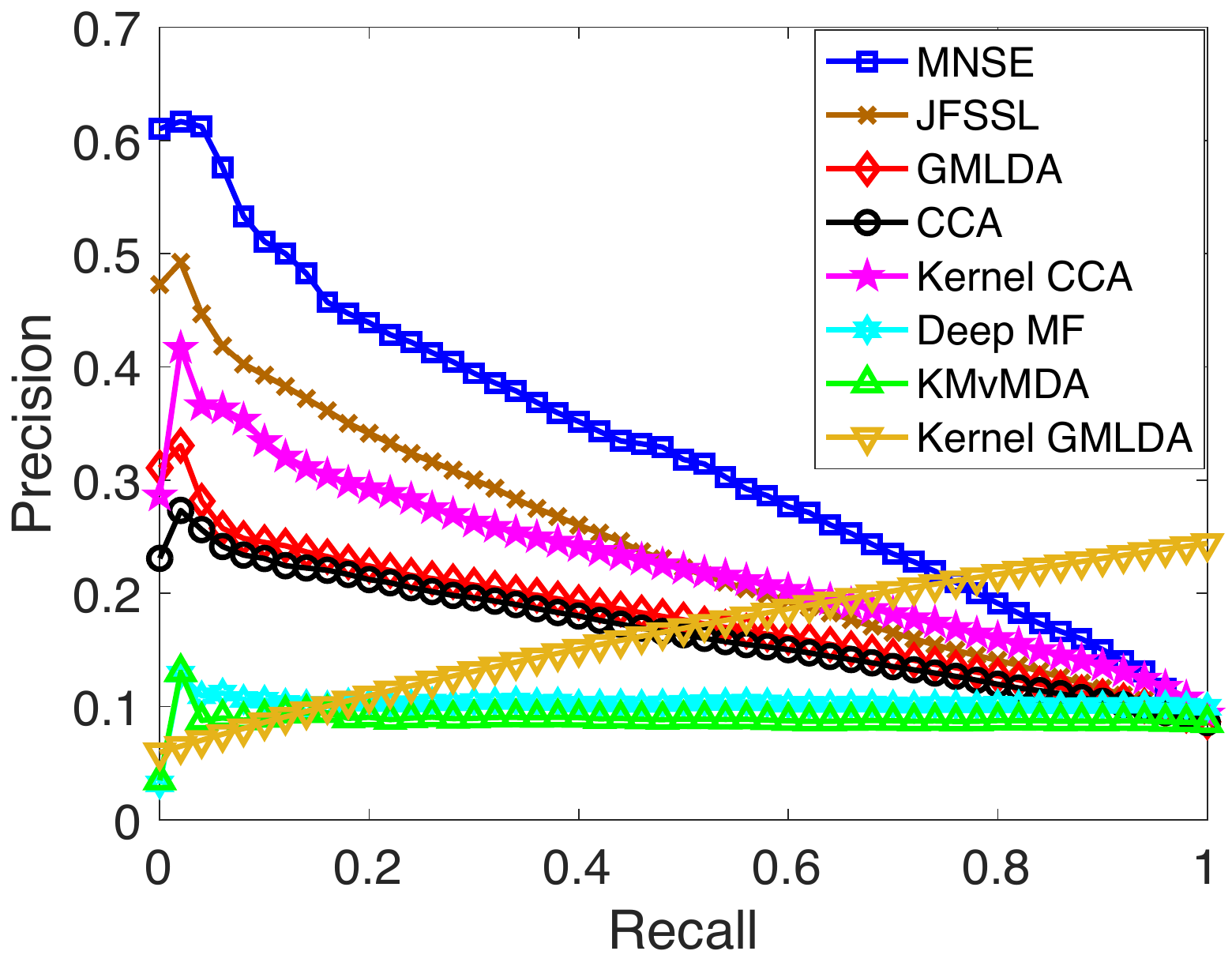}}
  \centerline{(c) Text query}\medskip
\end{minipage}
\hfill
\begin{minipage}[b]{0.49\linewidth}
  \centering
  \centerline{\includegraphics[width=4.3cm]{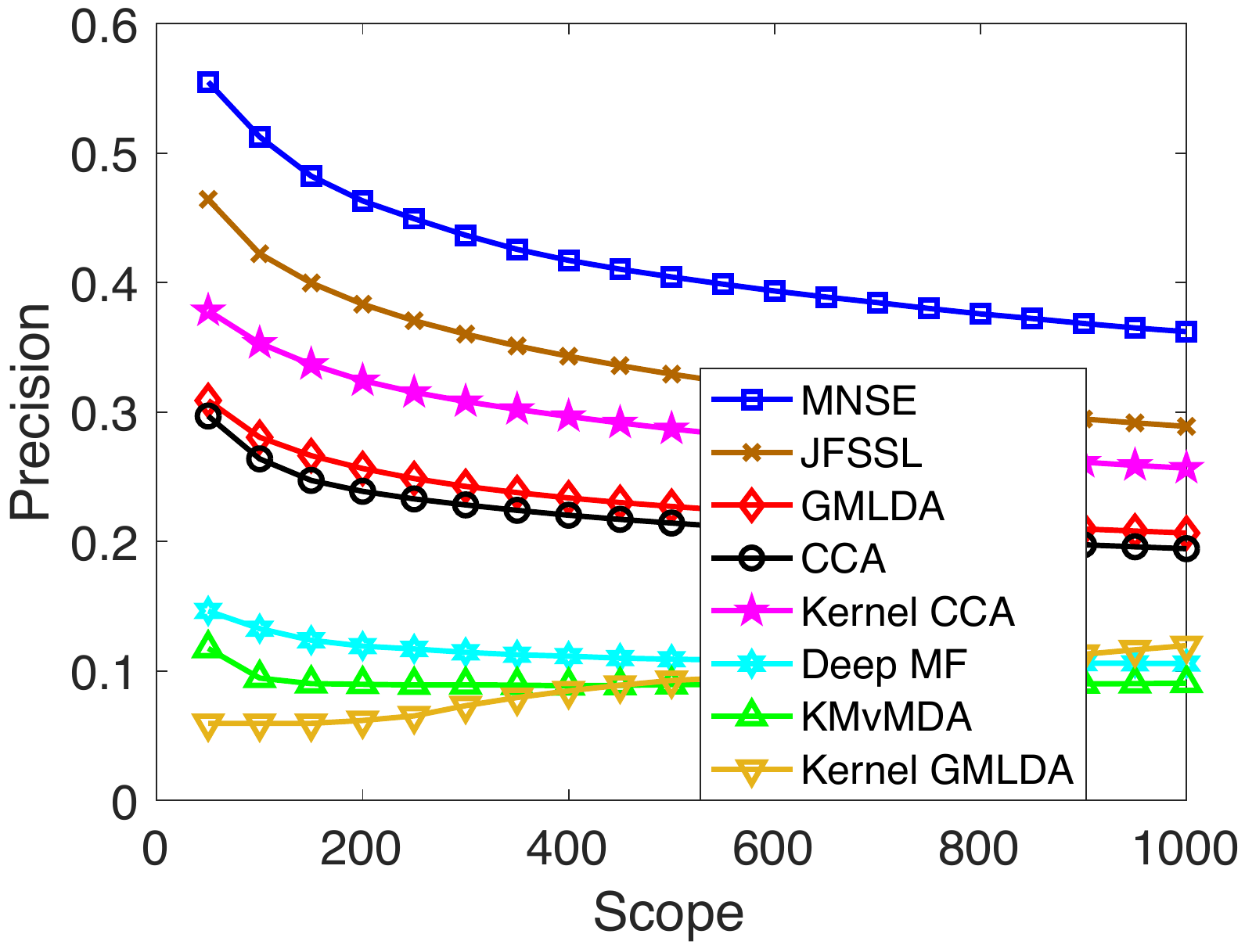}}
  \centerline{(d) Text query}\medskip
\end{minipage}
\vspace{-0.5cm}
\caption{Retrieval performance of the methods for Pascal VOC2007}
\label{figPascalRetrieval}
%
%\vspace{-0.3cm}
\end{figure}

The retrieval experiments are done on the Wikipedia and the Pascal VOC2007 image-text data sets. The weight parameters of MNSE are set to the values giving the highest average MAP score over the validation set of each experiment in the results of Section \ref{ssec:sensit_analy}. Figures \ref{figWikiRetrieval} and \ref{figPascalRetrieval} show the precision-recall and precision-scope curves for both types of queries, respectively on the Wikipedia and the Pascal VOC2007 data sets. Table \ref{tableWikiPascalRetrieval} reports the MAP scores of the methods on both data sets.

The proposed MNSE method outperforms all other multi-modal methods on both data sets. The Wikipedia and the Pascal VOC2007 data sets have diverse and irregular structures, with the two modalities bearing little resemblance. This makes the multi-modal representation learning task rather challenging, where the flexibility of the proposed nonlinear supervised embedding approach brings clear advantages over the other multi-modal methods in comparison. The performance gap between the proposed nonlinear MNSE method and the linear JFSSL and GMLDA algorithms can be explained in the way that nonlinear representations capture the intricate geometries of these two data sets better than linear representations. MNSE also performs significantly better than the supervised nonlinear methods Kernel GMLDA and KMvMDA, as well as the unsupervised Kernel CCA and DeepMF methods. These observations confirm the efficacy of the principle idea underlying MNSE: explicitly including a generalizability objective via the Lipschitz regularity of the interpolators improves the performance of nonlinear representation learning in data sets with complex geometries.

%% new table: WIKI & PASCAL
\begin{table}[t]
    \centering
	\caption{MAP scores for the Wikipedia and Pascal VOC2007 data sets}
	\scriptsize
\begin{tabular} {|*{5}{l|l|}} \hline
        \multirow{2}{*} \textbf{Algorithm} & \textbf{Wikipedia}  & \textbf{Wikipedia} & \textbf{Pascal VOC}  & \textbf{Pascal VOC} \\                                
     & \textbf{Image q.}  & \textbf{Text q.} & \textbf{Image q.}  & \textbf{Text q.} \\   \hline    
    {CCA \cite{surveyOnMl}} & 0.2280 & 0.1720 & 0.2470  &  0.1674  \\ \hline 
   {Ker. CCA \cite{surveyOnMl}}   & 0.2419  & 0.1815 & 0.2873 & 0.2282  \\ \hline  
 {GMLDA \cite{gma}}   & 0.2407  & 0.1815  & 0.2609  & 0.1791 \\ \hline 
  {Ker. GMLDA \cite{gma}}   & 0.1737  & 0.1326  & 0.1640 & 0.1640  \\ \hline 
{JFSSL \cite{jfssl}}   & 0.2440  & 0.2143   & 0.2814  & 0.2418  \\ \hline 
{DeepMF \cite{ZhaoDF17}}   & 0.1760  & 0.1335 & 0.1305  & 0.1038  \\ \hline 
 {KMvMDA \cite{CaoICG18}}   & 0.1661  & 0.1339   & 0.1023  & 0.0894 \\ \hline  
 {MNSE}   & \textbf{0.3109}  & \textbf{0.2332}   & \textbf{0.3710}  & \textbf{0.3221}  \\ \hline          
\end{tabular}
\label{tableWikiPascalRetrieval}
\vspace{-0.2cm}
\end{table}
\normalsize

\section{CONCLUSION}
\label{sec:concl}

We have first proposed a theoretical analysis of the performance of multi-modal supervised embedding methods in multi-modal classification and cross-modal retrieval applications. The main finding of our performance bounds is that achieving good between-class separation and cross-modal alignment is not sufficient, and the regularity of the multi-modal interpolation functions is also important for ensuring good generalization performance. Next, relying on these theoretical findings, we have proposed an algorithm for learning supervised multi-modal nonlinear embeddings, with  particular focus on the generalizability of the learnt representations to new test samples. The efficacy of the proposed method has been demonstrated in  multi-modal classification and cross-modal retrieval problems, where it has been shown to yield quite satisfactory performance in comparison with  recent multi-modal learning algorithms. We hope that our theoretical insights along with our methodological contributions will be useful towards improving the interpretability and the performance of nonlinear representation learning algorithms in multiple domains.

%In this study, a nonlinear subspace learning algorithm with an interpolator were obtained on different modalities. According to experimental results, supervised MNSE algorithm is more successful than unsupervised methods in terms of classification. Additionally, it was observed that MNSE algorithm had some advantegous than single modal algorithm NSSE. Neverthless, MNSSE produced similar results with JFSSL and PCA+GMLDA algorithms due to the linear structure of MITCBCL data set. MNSE gives better performance in not only classification but also retrieval. MNSE has the highest MAP scores on Wikipedia data set among compared algorithms. In the future, MNSE can be tested with a data set consists incomplete samples.

%
%\textcolor{red}{We may conduct several experiments on the data sets with more than two modalities or incomplete views in the future.}
%

\bibliographystyle{IEEEbib}
\bibliography{references}

\appendix

%\textbf{APPENDIX A}\\

%%%% Proof: Theorem 1 %%%%%%%%

\textbf{Proof of Theorem  \ref{thm:mm_class_bound}}

Before we prove Theorem \ref{thm:mm_class_bound}, we first present the following lemma, whose proof is given after that of Theorem \ref{thm:mm_class_bound}.

\begin{lemma}
\label{lem:bnds_single_q}
Let the training sample set $\mathcal{X}$ contain at least $\Nm$ training samples $\{\xii \}_{i=1}^{\Nm}$ from class $\m$, whose observations $\{ \xiq \}$ with $\xiq \sim  \matQ \nu_\m$ are available in all modalities $\q=1, \dots, \V$. Assume that the interpolation function $\fq: \matQ H \rightarrow \R^ \dy $ in each modality $\q$ is Lipschitz continuous with constant $L$.

Let $\x$ be a test sample from class $\m$ with an observation $\xp$ given in modality $\p$, drawn with respect to the probability measure $\nump$ independently of the training samples. Let $\xq$ be the observation of the same sample $\x$ in an arbitrary modality $\q$, which need not be available to the learning algorithm. For an arbitrary modality $\q \in \{1, \dots, \V \}$, define $\Aq$ as the set of the training samples from class $\m$ within a $\delta$-neighborhood of $\xq$ in $H^{(\q)}$
\[
\Aq = \{ \xiq : \xii \in \mathcal{X}, \ C(\xii) = \m, \ \xiq \in B_\delta(\xq) \}.
\] 

Assume that for some $Q \geq 1$ and $\delta >0$, the number of training samples from class $\m$ satisfies
\[
\Nm > \frac{Q}{\eta_{\m, \delta}}.
\]
Then for any $\epsilon >0$, with probability at least 
\begin{equation*}
\begin{split}
1 &- \exp \left(- \frac{2 (\Nm \eta_{\m, \delta} - Q)^2 }{\Nm} \right) 
 -  2\dy \exp \left(-\frac{ Q  \epsilon^2}{2 L^2 \delta^2} \right) \\
&-  (1-\eta_{\m, \delta} )^Q,
\end{split}
\end{equation*}
the set $\Aq$ contains at least $Q$ samples, the distance between $\fq(\xq)$ and the sample mean of the embeddings of its neighboring training samples is bounded as
\begin{equation}
\label{eq:event_lem3}
\left \|   
\fq (\xq) - \frac{1}{| \Aq |}  \sum_{\xiq \in \Aq}  \fq(\xiq)
\right \| 
\leq L\delta + \sqrt{\dy} \epsilon,
\end{equation} 
and also there is at least one $\xlq \in \Aq $ such that its observation $\xlp$ in modality $\p$ satisfies $\| \xlp - \xp \|\leq \delta$. 

\end{lemma}
% end of lemma

The purpose of Lemma \ref{lem:bnds_single_q} is to see how much the embedding of a test sample through a Lipschitz-continuous interpolator is expected to deviate from the average embedding of the training samples surrounding it. Lemma \ref{lem:bnds_single_q} provides a probabilistic upper bound on this deviation, which is used in Theorems \ref{thm:mm_class_bound} and \ref{thm:cm_retrieval_bnd} for bounding the classification and retrieval errors. Note that the classification algorithm knows the observation $\xp$ of the test sample $\x$ only in modality $\p$, and classifies it through its embedding $\fp(\xp)$ with respect to the rule in \eqref{eq:nn_class_rule}. The entity $\xq$ in the lemma denotes a hypothetical observation of $\x$ in an arbitrary modality $\q$. Although we conceptually refer to $\xq$ in the derivations, it is not known to the classification algorithm in practice (unless $\q = \p$). 

We can now prove Theorem \ref{thm:mm_class_bound}.

\begin{proof}
We first recall from Lemma \ref{lem:bnds_single_q} that for a particular modality $\q \in \{ 1, \dots, \V \}$, with probability  at least
\begin{equation*}
\begin{split}
1 &- \exp \left(- \frac{2 (\Nm \eta_{\m, \delta} - Q)^2 }{\Nm} \right) 
 -  2\dy \exp \left(-\frac{ Q  \epsilon^2}{2 L^2 \delta^2} \right) \\
&-  (1-\eta_{\m, \delta} )^Q
\end{split}
\end{equation*}
the set $\Aq$ has $| \Aq | \geq Q$ elements, the inequality in \eqref{eq:event_lem3} holds, and for at least one $\xlq \in \Aq$ we have $\| \xlp - \xp \|\leq \delta$. Since there are $\V$ modalities and the probability measures $\numq$ are independent, with probability at least 
\begin{equation*}
\begin{split}
1 &- \bigg(  \exp \left(- \frac{2 (\Nm \eta_{\m, \delta} - Q)^2 }{\Nm} \right) 
 +  2\dy \exp \left(-\frac{ Q  \epsilon^2}{2 L^2 \delta^2} \right) \\
&+  (1-\eta_{\m, \delta} )^Q \bigg)^V
\end{split}
\end{equation*}
there is at least one modality $\q \in \{1, \dots, V  \}$ such that all of these three events occur. Now, let $\xiq, \xjq \in B_\delta(\xq)$ denote two training samples from this modality $\q$ and class $\m$. As $\| \xiq - \xjq \| \leq 2 \delta $, from the assumption (P2) on the embedding, we have $ \| \yiq - \yjq  \| = \|  \fq (\xiq) - \fq (\xjq)  \| \leq \Adel$. This gives
\begin{equation}
\label{eq:bnd_fq_fqxjq}
\begin{split}
&\left \| \fq (\xiq) - \frac{1}{|\Aq|} \sum_{\xjq \in \Aq} \fq (\xjq)  \right \|  \\
&= \left \| \frac{1}{|\Aq|}  \sum_{\xjq \in \Aq} \left( \fq (\xiq)  - \fq (\xjq) \right) \right \| \\
& \leq  \frac{1}{|\Aq|}   \sum_{\xjq \in \Aq}  
\|  \fq (\xiq) - \fq (\xjq)   \| 
\leq \Adel.
\end{split}
\end{equation}
Then, for any $\xiq \in B_\delta(\xq)$, we have
\begin{equation}
\label{eq:bnd_fqxq_fqxiq}
\begin{split}
& \|  \fq(\xq) - \fq(\xiq)  \| \\
&= \bigg \|
 \fq(\xq)  - \frac{1}{|\Aq|} \sum_{\xjq \in \Aq} \fq (\xjq) \\
& \quad + \frac{1}{|\Aq|} \sum_{\xjq \in \Aq} \fq (\xjq) 
- \fq(\xiq)  
\bigg \| \\
& \leq
 \bigg \|
 \fq(\xq)  - \frac{1}{|\Aq|} \sum_{\xjq \in \Aq} \fq (\xjq) \bigg \| \\
 & +  \bigg \|
 \fq(\xiq) 
-
  \frac{1}{|\Aq|} \sum_{\xjq \in \Aq} \fq (\xjq)  
 \bigg \| \\
& \leq L\delta + \sqrt{\dy} \epsilon + \Adel
\end{split}
\end{equation}
where the last inequality follows from \eqref{eq:event_lem3} and \eqref{eq:bnd_fq_fqxjq}. 

Now, through the training sample $\xlq \in B_\delta(\xq)$ whose observation in modality $\p$ satisfies $\| \xlp - \xp \|\leq \delta$, and from property (P1) of the embedding, we observe that the deviation between the embedding $\fp(\xp)$ of the observation $\xp$ of the test sample used by the classification algorithm and the unknown embedding $\fq(\xq)$ of its unavailable observation $\xq$ is bounded as
\begin{equation*}
\begin{split}
& \|  \fp(\xp) - \fq(\xq) \| 
 \leq \|  \fp(\xp) - \fp(\xlp) \|  \\
& + \| \fp(\xlp) - \fq(\xlq) \|  
 + \|  \fq(\xlq) - \fq(\xq) \|  \\
 & \leq
 L\delta + \B + L \delta  =  2 L \delta + \B
\end{split}
\end{equation*}
where the second inequality follows from the Lipschitz continuity of the interpolators $\fp$ and $\fq$, and (P1). Combining this with \eqref{eq:bnd_fqxq_fqxiq}, we get that for the training samples $\xjq \in \Aq$ 
\begin{equation}
\label{eq:bnd_fpxp_fqxiq}
\begin{split}
& \|  \fp(\xp) - \fq(\xjq)  \| \\
& \leq  \|  \fp(\xp) - \fq(\xq)  \| 
+ \|  \fq(\xq)    - \fq(\xjq) \| \\
& \leq 
(2 L \delta + \B) + (L\delta + \sqrt{\dy} \epsilon + \Adel) \\
& = 3 L \delta + \sqrt{\dy} \epsilon + \Adel + \B.
\end{split}
\end{equation}
Next, let $\xkr$ be a training sample from another class than $m$, observed in any view $r=1, \dots, \V$.  The distance between the embeddings of $\xkr$ and the test sample $\xp$ is lower bounded as
\begin{equation}
\label{eq:fpxp_frxkr}
\begin{split}
& \|  \fp (\xp) - \fr (\xkr)  \| \geq  \\
& \|  \fq(\xjq) -\fr(\xkr)  \|  - \|  \fp(\xp) - \fq(\xjq) \| \\
& > \gamma - (3 L \delta + \sqrt{\dy} \epsilon + \Adel + \B)
\end{split}
\end{equation}
where the last inequality is obtained from the property (P3) of the embedding and the inequality in \eqref{eq:bnd_fpxp_fqxiq}. Using in \eqref{eq:fpxp_frxkr} the assumption \eqref{eq:cond_L_gamma} on the embedding, we get
\begin{equation}
\label{eq:fpxp_frxkr_final}
\begin{split}
\|  \fp (\xp) - \fr (\xkr)  \| > 3 L \delta + \sqrt{\dy} \epsilon + \Adel + \B.
\end{split}
\end{equation}
We finally observe from the inequalities \eqref{eq:bnd_fpxp_fqxiq} and \eqref{eq:fpxp_frxkr_final} that the embedding of any training sample $\xkr$ from another class than $m$ has distance larger than $3 L \delta + \sqrt{\dy} \epsilon + \Adel + \B$ to the embedding $\fp(\xp)$ of the test sample $\xp$, whereas there are at least $Q$ samples from the same class as $\xp$ within a distance of at most $3 L \delta + \sqrt{\dy} \epsilon + \Adel + \B$. We conclude that the test sample $\xp$ is then correctly classified via nearest neighbor classification through its embedding $\fp(\xp)$.
\end{proof}

%%%% Proof: Lemma 1 %%%%%%%%

\textbf{Proof of Lemma \ref{lem:bnds_single_q}}
\begin{proof}

For an arbitrary modality $\q \in \{ 1, \dots, \V \}$, the observation $\xiq$ of a training sample $\xii$ from class $\m$ drawn independently from the test sample $\x$ lies in a $\delta$-neighborhood of $\xq$ with probability
\[
P \left( \xiq \in B_\delta(\xq) \right) = \numq \left( B_\delta(\xq) \right)  
\geq \eta_{\m, \delta}.
\]
Then,
the probability that $B_\delta(\xq)$ contains at least $Q$ samples among the $\Nm$ training samples drawn from $\numq$ is given by 
\begin{equation*}
\begin{split}
& P(| \Aq| \geq Q) \\ 
&= \sum_{k =Q}^{\Nm} {{\Nm} \choose {k}}  \left ( \numq( B_\delta(\xq)) \right)^k \left(  1 - \numq( B_\delta(\xq)) \right)^{\Nm - k}\\
& \geq  \sum_{k =Q}^{\Nm} {{\Nm} \choose {k}} (\eta_{\m, \delta})^k (1-\eta_{\m, \delta})^{\Nm - k}.
\end{split}
\end{equation*}
This is obtained by evaluating the probability that at least $Q$ successes occur within $\Nm$ independent Bernoulli trials with success probability more than $\eta_{\m, \delta}$ in each trial. Following the approach in the proof of \cite[Theorem 5]{supervisedManifold}, from the assumption $\Nm > \frac{Q}{\eta_{\m, \delta}}$, we can lower bound this probability using a tail bound for distributions \cite{Herbrich99}. We thus get
\[
 P(| \Aq| \geq Q) \geq 1 - \exp \left(- \frac{2 (\Nm \eta_{\m, \delta} - Q)^2 }{\Nm} \right).
\]
Now assume that the event $| \Aq| \geq Q$ has occured for the modality $\q$, i.e., there are at least $Q$ training samples from class $m$ within a $\delta$-neighborhood of $\xq$. Then, from \cite[Lemma 3]{supervisedManifold}, with probability at least
\[
1 - 2\dy \exp \left(-\frac{|  \Aq | \epsilon^2}{2 L^2 \delta^2}\right) \geq 1 - 2\dy \exp \left(-\frac{ Q  \epsilon^2}{2 L^2 \delta^2} \right)
\]
the distance between $\fq(\xq)$ and the sample average of the embeddings of its neighboring training samples is bounded as
\begin{equation}
\label{eq:pf_fqxq_fqxiq}
\left \|   
\fq (\xq) - \frac{1}{| \Aq |}  \sum_{\xiq \in \Aq}  \fq(\xiq)
\right \| 
\leq L\delta + \sqrt{\dy} \epsilon.
\end{equation}
Next, still assuming that the event $| \Aq| \geq Q$ has occurred for the modality $\q$, for each sample  $ \xiq \in \Aq $, the probability that its observation $\xip$ in modality $\p$ is outside $B_\delta(\xp))$ is
\[
1 - \nump(B_\delta (\xp)) \leq 1 - \eta_{\m, \delta} .
\]
Therefore, with probability at least $1 - (1-\eta_{\m, \delta} )^Q $, there is at least one $\xlq \in B_\delta(\xq)$ whose observation in modality $\p$ satisfies $ \xlp \in B_\delta(\xp)$, or equivalently, $\| \xlp - \xp \|\leq \delta$. Combining the probability expressions we obtained so far, we conclude that for an arbitrary modality $q$, with probability at least 
\begin{equation*}
\begin{split}
1 &- \exp \left(- \frac{2 (\Nm \eta_{\m, \delta} - Q)^2 }{\Nm} \right) 
 -  2\dy \exp \left(-\frac{ Q  \epsilon^2}{2 L^2 \delta^2} \right) \\
&-  (1-\eta_{\m, \delta} )^Q
\end{split}
\end{equation*}
we have $| \Aq| \geq Q$, the event  in \eqref{eq:pf_fqxq_fqxiq} occurs, and there is at least one  $\xlq \in B_\delta(\xq)$ such that $\| \xlp - \xp \|\leq \delta$. 
\end{proof}

%%%% Proof of Theorem 2 %%%%%%
\textbf{Proof of Theorem  \ref{thm:cm_retrieval_bnd}}

\begin{proof}
Recall from Lemma \ref{lem:bnds_single_q} that with probability at least
\begin{equation*}
\begin{split}
1 &- \exp \left(- \frac{2 (\Nm \eta_{\m, \delta} - Q)^2 }{\Nm} \right) 
 -  2\dy \exp \left(-\frac{ Q  \epsilon^2}{2 L^2 \delta^2} \right) \\
&-  (1-\eta_{\m, \delta} )^Q
\end{split}
\end{equation*}
the embedding $\fq(\xq)$ of the query sample in modality $\q$ has $| \Aq | \geq Q$ neighboring training samples from the same class $\m$, the inequality in \eqref{eq:event_lem3} holds, and for at least one $\xlq \in \Aq$ we have $\| \xlp - \xp \|\leq \delta$. Assuming that all these three events have occurred and following the same steps as in the proof of Theorem \ref{thm:mm_class_bound}, we conclude that there are at least $Q$ samples $\xjq \in B_\delta(\xq)$ such that
\begin{equation}
\begin{split}
 \|  \fp(\xp) - \fq(\xjq)  \| 
 \leq 3 L \delta + \sqrt{\dy} \epsilon + \Adel + \B
\end{split}
\end{equation}
while the distance of the embedding of any training sample $\xkr$ from another class to $\fp(\xp)$ is lower bounded as
\begin{equation}
\begin{split}
 \|  \fp (\xp) - \fr (\xkr)  \| >
 3 L \delta + \sqrt{\dy} \epsilon + \Adel + \B.
\end{split}
\end{equation}

This implies that the $Q$ samples of smallest distance to the embedding $ \fp (\xp)$ of the query sample are all from class $\m$. Hence, for $\K \leq Q$, the precision rate over the $\K$ nearest neighbors is
\[
P=\K/\K=1.
\]
Similarly, when $\K>Q$, at least $Q$ of the $\K$ nearest neighbors of $ \fp (\xp)$  are from the same class $\m$, hence we get
\[ 
P\geq Q/K.
\]
Meanwhile, when $\K \leq Q$, the retrieval algorithm returns $\K$ samples out of the $\Nm$ training samples from the same class $\m$, hence
\[
R=\frac{K}{\Nm}.
\]
Finally, when $\K > Q$, since at least $Q$ of the retrieved training samples will be from the same class as the query sample, we have
\[
R \geq \frac{Q}{\Nm}.
\]
\end{proof}

\end{document}